\documentclass[9pt]{extarticle}
\pdfoutput=1

\usepackage[margin=1in]{geometry}
\usepackage{graphics}      
\usepackage[T1]{fontenc}   
\usepackage[pdflang={en-US},pdftex]{hyperref}
\usepackage{color}
\usepackage{booktabs}
\usepackage{todonotes}
\usepackage[sort&compress,numbers]{natbib}
\graphicspath{{./images/}{./images/learning/}{./images/performance/}{./images/user/}}

\usepackage{amsxtra}
\usepackage{amstext}
\usepackage{newpxmath}

\usepackage{array}
\usepackage{ctable} 
\usepackage{multirow}
\usepackage{arydshln}
\newcolumntype{L}[1]{>{\raggedright\let\newline\\\arraybackslash\hspace{0pt}}m{#1}}
\newcolumntype{C}[1]{>{\centering\let\newline\\\arraybackslash\hspace{0pt}}m{#1}}
\newcolumntype{R}[1]{>{\raggedleft\let\newline\\\arraybackslash\hspace{0pt}}m{#1}}

\usepackage{setspace}
\usepackage{rotating}

\usepackage{subcaption}
\usepackage{xcolor}
\usepackage{xr}
\usepackage{csquotes}

\newenvironment{itquote}
{\begin{quote}\itshape}
{\end{quote}}

\def\plaintitle{Interestingness Elements for Explainable Reinforcement Learning: Understanding Agents' Capabilities and Limitations}
\def\plainauthor{Pedro Sequeira and Melinda Gervasio}
\def\plainkeywords{Explainable AI; Reinforcement Learning; Interestingness Elements; Autonomy; Video Highlights; Visual Summaries}

\DeclareMathOperator*{\argmax}{\rm argmax}

\newcommand{\EE}[2][]{\mathbb{E}_{#1}\left[#2\right]}

\newcommand{\abs}[1]{\lvert#1\rvert}					
\newcommand{\set}[1]{\lbrace#1\rbrace}				
\newcommand{\vect}[1]{\left[#1\right]}					
\newcommand{\card}[1]{\lvert#1\rvert}					

\newcommand{\R}{\mathbb{R}}

\newcommand{\M}{\mathcal{M}}		
\newcommand{\MDP}[1][]{(\S_{#1},\A_{#1},\P_{#1},\Rwd_{#1},\gamma)}

\renewcommand{\P}{\mathsf{P}}		
\renewcommand{\S}{\mathcal{S}}		
\newcommand{\A}{\mathcal{A}}			
\newcommand{\Z}{\mathcal{Z}}			
\newcommand{\Rwd}{\mathsf{R}}		

\newcommand{\cs}{n}	
\newcommand{\csa}{n}	
\newcommand{\csas}{n}	
\newcommand{\ts}{\tau}	
\newcommand{\tsa}{\tau}	

\newcommand{\evenness}{\xi}				
\newcommand{\rwdavg}{\bar{r}}			
\newcommand{\z}{\mathsf{z}}				
\newcommand{\numfeats}{N}
\newcommand{\trans}{\mathcal{T}}			

\newcommand{\eps}{\varepsilon}

\newcommand{\eg}{\textit{e.g.},~}
\newcommand{\Eg}{\textit{E.g.},~}
\newcommand{\ie}{\textit{i.e.},~}

\newcommand{\extraction}[1]{\vspace{5pt}\noindent\textbf{Elements' extraction.} #1}

\definecolor{highcolor}{HTML}{D7191C}
\newcommand{\highdiv}[1]{\pmb{\textcolor{highcolor}{#1}}}
\definecolor{lowcolor}{HTML}{2B83BA}
\newcommand{\lowdiv}[1]{\pmb{\textcolor{lowcolor}{#1}}}

\hypersetup{%
    pdftitle={\plaintitle},
    pdfauthor={\plainauthor},
    pdfkeywords={\plainkeywords},
    pdfdisplaydoctitle=true, 
    bookmarksnumbered,
    pdfstartview={FitH},
    colorlinks,
    citecolor=blue,
    filecolor=black,
    linkcolor=blue,
    urlcolor=blue,
    breaklinks=true,
    hypertexnames=false
}

\begin{document}

\title{\plaintitle\thanks{Accepted version of manuscript. Article available at \url{https://doi.org/10.1016/j.artint.2020.103367}.}}

\author{Pedro Sequeira and Melinda Gervasio \\
	SRI International\\333 Ravenswood Avenue, Menlo Park, CA 94025, United States\\
	pedro.sequeira@sri.com, melinda.gervasio@sri.com}

\maketitle

\begin{abstract}

We propose an explainable reinforcement learning (XRL) framework that analyzes an agent's history of interaction with the environment to extract interestingness elements that help explain its behavior. The framework relies on data readily available from standard RL algorithms, augmented with data that can easily be collected by the agent while learning. We describe how to create visual summaries of an agent's behavior in the form of short video-clips highlighting key interaction moments, based on the proposed elements. We also report on a user study where we evaluated the ability of humans to correctly perceive the aptitude of agents with different characteristics, including their capabilities and limitations, given visual summaries automatically generated by our framework. The results show that the diversity of aspects captured by the different interestingness elements is crucial to help humans correctly understand an agent's strengths and limitations in performing a task, and determine when it might need adjustments to improve its performance.

\end{abstract}


\section{Introduction}%
\label{Sec:Intro}

Reinforcement learning (RL) is a popular computational approach for autonomous agents facing a sequential decision problem in dynamic and often uncertain environments \citep{Sutton1998}. The goal of any RL algorithm is to learn a \emph{policy}, \ie a mapping from states to actions, given trial-and-error interactions between the agent and an environment. Typical approaches to RL focus on memoryless (reactive) agents that select their actions based solely on their current observation \citep{Littman1994}. This means that by the end of learning, an RL agent can select the most appropriate action in each situation---the learned policy ensures that doing so will maximize the reward received by the agent during its lifespan, thereby performing according to the underlying task assigned by its designer. 

A trained RL agent does not explicitly reason about its future to select actions. All it knows is that it \emph{should} select actions as dictated by its learned policy, which makes it hard to explain its behavior. At most, agents know that choosing one action is preferable over others, or that some actions are associated with a higher value---but not why that is so or how it came to be. As a consequence, the \emph{why} behind decision-making is lost during the learning process as the policy converges to an optimal action-selection mechanism. RL further complicates explainability by enabling an agent to learn from \emph{delayed rewards}~\citep{Sutton1998}---the reward received after executing some action is ``propagated'' back to the states and actions that led to that situation, meaning that important actions may not be associated with any (positive) reward. 

Ultimately, RL agents lack the ability to know why some actions are preferred over others, to identify the goals that they are currently pursuing, to recognize what elements are more desirable, to identify situations that are ``hard to learn'', or to summarize the strategy learned to solve the task. 

This lack of self-explainability can be detrimental to establishing trust with human collaborators who may need to delegate critical tasks to agents. An eXplainable RL (XRL) system that enables humans to correctly understand the agent's \emph{aptitude} in a specific task, \ie both its capabilities and limitations, whether innate or learned, will enable human collaborators to delegate tasks more appropriately as well as identify situations where the agent's perceptual, actuating, or control mechanisms may need to be adjusted prior to deployment. 

\subsection{Approach Overview}

\begin{figure*}[!tb]
	\centering
	\includegraphics[width=1\textwidth]{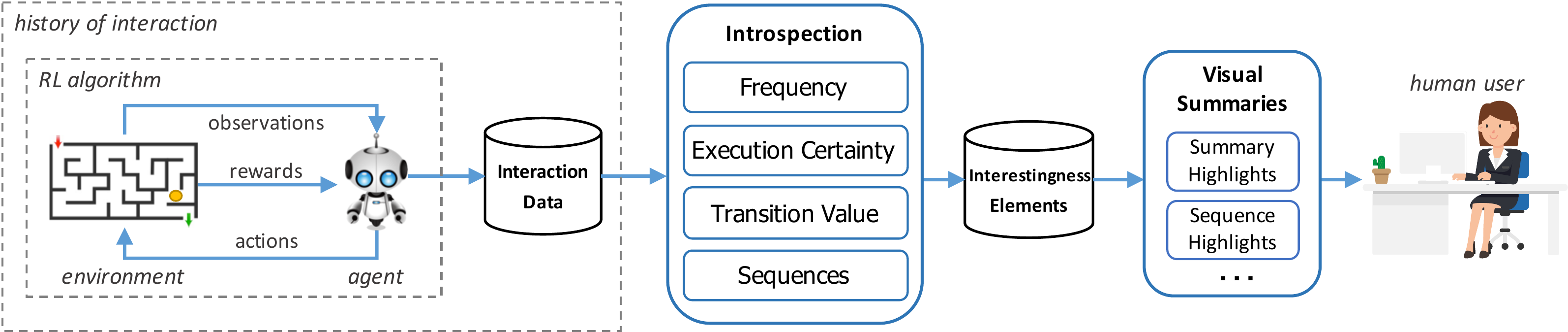}
	\caption{Our framework for explainable autonomous RL agents. During its interaction with the environment, the agent collects relevant statistical information. Our introspection framework analyzes the collected data and identifies interestingness elements of the history of interaction. These elements are then used to summarize an agent's learned behavior and its capabilities and limitations in the task to a human user.}%
	\label{Fig:Framework}
\end{figure*}

In this paper, we present a framework towards making autonomous RL agents more explainable through introspective analysis (Fig.~\ref{Fig:Framework}). The agent examines its history of interaction with the environment to extract \emph{interestingness elements}%
\footnote{We borrow the term from the association rule mining literature \cite[see \eg][]{Bayardo1999} to denote potentially relevant aspects of the agent's history of interaction with the environment.}
denoting meaningful situations, \ie potentially ``interesting'' properties of the interaction. The goal of the elements is to allow humans to understand an agent's underlying characteristics and aptitude in a task, thus making an RL system more \emph{explainable}.%
\footnote{Although the elements and visual summaries generated by the framework \emph{are not} explanations, they allow users to interpret an agent's behavior through the observation of relevant interactions.}
As depicted in Fig.~\ref{Fig:Framework}, four introspection dimensions are analyzed: frequency, execution certainty, transition-value and sequences.%
\footnote{In this paper we detail only the elements evaluated in our user study. We refer to Appendix~\ref{Sec:AppendixOtherElements} for additional elements of different types and to \cite{Sequeira2019} for a preliminary version.}
Each element is generated by applying statistical analysis and machine learning methods to data collected \emph{during} the interaction. The introspection framework is the basis of a summarization system following the approach in \citep{Amir2018}, that extracts visual summaries highlighting relevant moments of the agent's interaction with an environment according to the different interestingness elements.

We conducted a user study to evaluate the usefulness of the proposed interestingness elements in helping characterize the behavior of different RL agents. The goal was to understand \emph{why} and \emph{for what} each element might be useful. We designed agents with diverse goals and perceptual characteristics in a video game task. The goal was to simulate players with different capabilities and limitations, and needing improvement in different aspect of the games. We trained each agent using standard RL, resulting in distinct behaviors and levels of game performance. We then conducted an online study where we asked subjects to observe several videos comprising visual summaries, each depicting the behavior of a particular agent based on specific interestingness elements. 

Overall, our results show that different agent characteristics require different elements to be presented to a user to enable a correct understanding of proficiency in different parts of a task. Namely, presenting only ``good'' or only ``bad'' instances of agent behavior can lead users to form incorrect expectations about agent performance. The results also show that by capturing the diversity of an agent's interaction with the environment, frequently and infrequently visited states enable a correct understanding of the agent's general aptitude. Moreover, short summaries showing how an agent overcomes challenges and achieves its goals can provide a short but accurate depiction of its learned strategy.

The paper is organized as follows. Sec.~\ref{Sec:RelatedWork} discusses related works and compares them with our approach. Sec.~\ref{Sec:RL} introduces the necessary concepts behind the RL framework. Sec.~\ref{Sec:Framework} details the proposed interestingness elements while Sec.~\ref{Sec:Summaries} shows how visual summaries can be derived from them. Sec.~\ref{Sec:Study} presents the experimental study, including the design of the different RL agents and the survey. Sec.~\ref{Sec:Analysis} details the results of the user study and provides an in-depth discussion of the main insights stemming from the results. Finally, Sec.~\ref{Sec:Conclusions} summarizes the main findings and describes current and future developments.

\section{Related Work}%
\label{Sec:RelatedWork}

In this section we review recent works within XRL that have addressed some of the problems identified in this paper. We summarize the technical approaches, highlight the main contributions and compare them to our framework.

\subsection{Deriving Language Templates}

One approach is to use language templates to translate elements of the problem or the agent's policy into human-understandable explanations. For example, \citet{Khan2009} used the discounted state-visitation frequency according to a policy to form contrastive explanations about the expected return in a given state by executing the optimal action versus alternative ones. In particular, their minimal sufficient explanations (MSE) populate templates and convey to the user the probability or frequency of reaching certain states with high rewards by using the optimal action but not other actions. However, explanations can be long and ill-suited to non-technical users, for whom the concept of utility may not be easy to understand.

\citet{Elizalde2009} developed an intelligent assistant for operator training (IAOT) relying on factored state scenarios. The system is capable of providing explanations using templates tailored to novice, intermediate, and advanced users. Explanations are generated whenever a mistake is made and include the optimal action that should be performed, a textual justification according to the user level, and a (domain-dependent) visual diagram highlight the most relevant variable. The latter corresponds to the state feature providing the highest positive difference in the value function. A drawback of this method is that it requires creating a knowledge base for the justifications in the templates, \eg obtained from written documentation or domain experts.

\citet{Wang2016} generate explanations for partially-observable MDPs that expose users to the agent's policy, beliefs over states of the environment, transition probabilities, rewards, etc. in human-robot interaction (HRI) scenarios. The agent can refer to properties of the environment and its behavior but cannot identify the most pertinent situations or justify its actions. 

Unlike these approaches which require a large amount of domain knowledge to create explanations, our framework relies on standard RL data and is domain-agnostic. In addition, techniques like ours that use visual summaries of behavior make understanding the agent's behavior more accessible to non-expert users.

\subsection{Deep RL Over Image-Based Inputs}

Other approaches aim to provide insights about policies learned using deep RL techniques on image-based input tasks. For example, \citet{Zahavy2016} aggregate the high-dimensional state space of ATARI games into a hierarchical structure using t-Distributed Stochastic Neighbor Embedding (t-SNE) for dimensionality reduction. A graphical user interface then allows the manual selection and visualization of the aggregation of different states according to several types of features. However, the representation and provided information make it difficult for non-RL experts to extract meaningful information. 

\citet{Greydanus2018} use perturbation-based techniques to identify image regions that result in significant policy changes. The result is a video of agent performance where the relevant regions of the image are overlaid with different colors. This approach is computationally demanding as it requires sweeping whole input images at every step, and users of the system would be required to observe full-length videos of an agent's performance to understand its policy. 

An advantage of these approaches is that they allow scaling to large input space domains. A drawback is that they cannot generalize to tasks with non-image-based inputs. Our approach is agnostic to the type of observations to derive the different elements. In addition, unlike these approaches that expose the entirety of an agent's behavior, we follow others who select only the moments of the agent's behavior that are relevant according to different criteria. 

\subsection{Abstracting Explainable Representations}

\citet{Koul2019} learn finite state representations of policies learned using deep recurrent neural networks (RNNs) that are agnostic to the input type. The idea is to abstract the learned policy into a simpler, more interpretable structure. First, quantized bottleneck networks (QBNs) are learned from a trained RNN policy that can reconstruct the observations and hidden states using a lower-dimensional representation space. From this, a finite state machine is extracted and discretized. Although the resulting structures provide insights on the nature of the learned policy, \eg that it contains certain cycles, it is hard to interpret the semantics behind the discretized observations and hidden states. 

Other works focus on abstracting state representations of the task and creating graph structures denoting the agent's behavior. For example, \citet{Hayes2017} and \citet{Waa2018} frame explanations as summaries assembled from outputs of binary classifiers that characterize the agent's trajectories and decisions. The system in \citep{Hayes2017} was designed for HRI scenarios where users pose questions that are identified from a set of question templates, resolved to a relevant set of world states, and then summarized and composed into a natural language form to respond to the inquiring user. 

The approach by \citet{Waa2018} allows users to ask contrastive questions (why $x$ instead of $y$?) about the agent's policy. Similar to our element of sequences, a path is constructed from the most likely state-action pairs. An explanation is then constructed from a language template. A similar approach was proposed by \citet{Madumal2019}, where action influence graphs for RL are extracted via structural causal models. This allows generating explanations for \emph{why} and \emph{why not} questions regarding the actions are selected over others. A user study was performed by applying the method to an agent policy learned on a simplified StarCraft II combat game, and results show that the method allowed for a better understanding of the agent's strategy.

In contrast to these approaches, our framework identifies relevant aspects of the agent's behavior according to different criteria and provides the semantics for the visual summaries.

\subsection{Extracting Behavior Examples}

Some approaches generate examples of agent behavior in the form of sequences of state-action pairs that maximize the understanding of the agent's policy by a human observer. Such approaches typically rely on inverse reinforcement learning (IRL) to generate optimal examples, \ie trajectories that, when observed by humans, allow them to recover the reward function used by the agent, which subsequently allows them to reason about the agent's planning process.

For example, \citet{Huang2019} model how humans make inferences about an agent's (robot) reward function from examples of its behavior. To select maximally informative examples, authors rely on algorithmic teaching techniques and different distance functions to select trajectories that best describe the reward function when compared to other reward functions, \ie that maximize the probability of humans inferring the correct agent goals. 

\citet{Lage2019} use a similar machine teaching approach by producing summaries of policies by extracting trajectories that maximize the quality of their reconstruction according to the principle of maximum entropy. The approach also modeled the human reconstruction of the agent's policy using imitation learning instead of IRL. Human-subject studies on different RL domains showed the importance of selecting appropriate user models for summarization to domain context.

This approach is related to how we select examples of behavior by using the interestingness elements, but it's targeted at having people understand agent goals---or goal-directed behavior---rather than their competency or aptitude in a task. Therefore, the selection criteria of trajectories are based on how informative they are of the agent's typical behavior. In contrast, within our framework, behavior summaries are only \emph{a means} of explaining the agent's behavior; the focus is on identifying situations that are relevant to how the agent performs the task.

\subsection{Identifying Key Interaction Moments}

Approaches more similar to ours try to identify key moments of the agent's interaction. As proposed by \citet{Amir2019}, the idea is that behavior summaries or examples should cover states that are likely to be of interest to users. In that paper, the authors provide a conceptual model of the summarization process of agents and propose several strategies for selecting important states, many of which are realized by our proposed interestingness elements.

\citet{Huang2018} present a system capturing \emph{critical states}, in which the value of one action is comparatively much higher than others. They performed a user study where subjects were exposed to critical states extracted for two different policies in the game of Pong, where one policy had a much higher training time. Subjects were then asked whether they trusted each policy and whether they would take control over them in a series of query states. Results show that exposing users to critical states enables an appropriate level of trust with regard to the agents' performance. 

Our approach for visual summaries is based on the highlights system proposed by \citet{Amir2018}. Their system captured video clips highlighting an agent's behavior in the game of Pacman using the concept of importance, dictated by the largest difference in the $Q$-values of a given state. This allows the identification of trajectories that are representative of the agent's best performance. Subjects in a user study were shown highlights of three agents with increasing amounts of training (and hence, proficiency) and the results show that the proposed importance metric enabled correct assessment of the agents' relative performance compared to other baseline summarization methods. 

A first observation is that we go beyond the concept of importance for explaining the agent's behavior proposed in these papers as dictated solely by the highest value-difference states. In particular, we propose mathematical interpretations within RL for concepts such as frequency, uncertainty, predictability and contradiction. Our framework also includes an element that allows visualizing sequences of relevant states rather than single moments. Further, the prior user studies focused mainly on the perception of agent proficiency and whether users should trust them based on their performance. In this paper we are primarily concerned with establishing connections between the different elements and the correct understanding of both the agent's capabilities and limitations. Importantly, our results provided insights on which elements are better at indicating when and how agents might need intervention to improve their performance.

\section{Reinforcement Learning}%
\label{Sec:RL}

We now introduce the necessary notation and terminology used throughout the paper related to the computational framework of RL. RL agents can be modeled using the \emph{Markov decision process} (MDP) framework \citep{Puterman1994}. Formally, we denote an MDP as a tuple $\M=\MDP$, where: $\S$ is the set of all possible environment states; $\A$ is the action repertoire of the agent; $\P(s'\mid s,a)$ is the probability of the agent visiting state $s'$ after having executed action $a$ in state $s$; $\Rwd(s,a)$ represents the reward that the agent receives for performing action $a$ in state $s$; $\gamma\in[0,1]$ is some discount factor denoting the importance of future rewards.

An MDP evolves as follows. At each time step $t=0,1,2,\ldots$, the environment is in some state $S_t=s$. The agent selects some action $A_t=a\in\A$ and the environment transitions to state $S_{t+1}=s'$ with probability $\P(s'\mid s,a)$. The agent receives a reward $\Rwd(s,a)=r\in\R$, and the process repeats.

The goal of the agent can be formalized as that of gathering as much reward as possible throughout its lifespan discounted by $\gamma$. This corresponds to maximizing the value $v=\EE{\sum_t\gamma^t r}$, where the reward $r$ dictates the immediate utility of taking action $a$ in state $s$ at time-step $t$, in light of the underlying task that the agent must solve. In order to maximize $v$, the agent must learn a policy, denoted by $\pi:\S\to\A$, that maps each state $s\in\S$ directly to an action $\pi(s)\in\A$. In the case of MDPs, this corresponds to learning a policy $\pi^*:\S\to\A$ referred to as the \emph{optimal policy} maximizing the value $v$. 

Typical approaches to RL learn a function $Q^*:\S\times\A\to\R$ associated with $\pi^*$ that verifies the recursive relation:

\begin{equation}\label{Eq:Bellman}
	Q^*(s,a)=r+\gamma\sum_{s'\in\S}\P(s'\mid s,a)\max_{b\in\A}Q^*(s',b).
\end{equation}
$Q^*(s,a)$ represents the \emph{value} of executing action $a$ in state $s$ and henceforth following the optimal policy. RL algorithms like $Q$-learning \citep{Watkins1992} assume that the agent has no knowledge of either $\P$ or $\Rwd$. Hence, they typically start by \emph{exploring} the environment---selecting actions in some exploratory manner---collecting samples in the form $(s,a,r,s')$ which are then used to successively approximate $Q^*$ using Eq.~\ref{Eq:Bellman}. After exploring its environment, the agent can \emph{exploit} its knowledge and select the actions that maximize (its estimate of) $Q^*$.

\section{Introspection Framework}%
\label{Sec:Framework}

In this section we describe our introspection framework, including the generated interestingness elements. 

\subsection{Interaction Data}%
\label{Subsec:InteractionData}

As shown in Fig.~\ref{Fig:Framework}, the introspection framework relies on the following data collected by the agent during its interaction with the environment:
\begin{itemize}
	\item $\cs(s)$: the number of times the agent visited $s$; $\csa(s,a)$: the number of times it executed action $a$ after observing $s$; and $\csas(s,a,s')$: the number of times it visited $s'$ after executing action $a$ when observing $s$;
	\item $\hat{\P}(s'\mid s,a)$: the \emph{estimated probability} of observing $s'$ when executing action $a$ after observing $s$. This can be modeled from the interaction according to $\hat{\P}(s'\mid s,a)=\csas(s,a,s') / \csa(s,a)$;
	\item $Q(s,a)$: the agent's \emph{estimate of the $Q$ function}, corresponding to the expected value of executing $a$ having visited $s$ and henceforth following the current policy. This can be estimated using any value-based RL algorithm;
	\item $V(s)$: the agent's \emph{estimate of the $V$ function} that indicates the value of observing $s$ and henceforth following the current policy. This corresponds to $V(s)=\max_{a\in\A}Q(s,a)$;
\end{itemize}
%
Some of this data is already collected by standard RL methods, namely the $Q$ and $V$ functions, and, in the case of model-based approaches, the $\hat{\P}$ and $\hat{\Rwd}$ functions. The remainder can easily be collected by the agent during its interaction with the environment by updating counters and running averages.

As shown in Fig.~\ref{Fig:Framework}, this data is used in various introspective analyses designed to extract statistical information from the interaction that might help explain the agent's behavior. We note that it is up to the end-user of the system to determine what constitutes the \emph{history of interaction} during which the data is collected. Namely, if the interaction data is collected \emph{during training}, it can retain information about the interaction beyond the learned policy, thereby potentially capturing interesting challenges encountered by the agent while learning. In contrast, capturing data only \emph{after training} might miss some subtleties of the training phase but will be more representative of the agent's learned policy. Another possibility is to capture data during both training and testing to capture a set of interesting situations without losing aspects of the agent's optimal behavior.

Table~\ref{Table:Elements} lists the analyses considered in this paper and a short description of the elements that each generates. Next, we present the purpose behind each dimension of analysis and discuss how each can be used to help explain the agent's interaction with an environment, providing a mathematical interpretation for the extraction of the interestingness elements given available interaction data.

\begin{table*}[!tb]
	\centering
	\small
	\renewcommand{\arraystretch}{1.5}
	\setlength\tabcolsep{0pt}
	\caption{Overview of the dimensions of analysis evaluated in the user study and the generated interestingness elements.}%
	\label{Table:Elements}
	\begin{tabular}{L{60pt}@{\hspace{6pt}} L{280pt} }
	\toprule
	\textbf{Dimension} & 
	\textbf{Generated Interestingness Elements} \\ 
	\hline
	\textbf{Frequency} & 
	{(in)frequent situations:} \emph{situations the agent finds very (un)common}\\
	\textbf{Execution Certainty} &
	{(un)certain executions:} \emph{hard/easy to predict situations with regards to action execution}\\
	\textbf{Transition-Value} & 
	{minima and maxima:} \emph{favorable and adverse situations}\\
	\textbf{Sequence} & 
	{most likely sequences to maxima:} \emph{capture the agent's learned strategy}\\
	\bottomrule
	\end{tabular}
\end{table*}

\subsection{Frequency Analysis}

This dimension of analysis can be used to expose the agent's (in)experience with its environment, denoting both commonly and rarely encountered situations. It extracts two interestingness elements: \textbf{frequent situations} and \textbf{infrequent situations}.

Infrequent elements may indicate states not sufficiently explored by the agent. \Eg  imagine an agent learning to run through a maze. If during learning the agent finds locations that are hard to reach but are not relevant to solving the maze, after training these would likely be identified as infrequent. Another example is a situation that had such a negative impact on the agent's performance, \eg a death situation in a game, that its action-selection mechanisms made sure they were rarely visited in subsequent visits to nearby states. This element can also denote novel or rare situations, which could indicate a highly dynamic environment. In an interactive setting, these elements provide opportunities for an agent to ask a user for guidance, \eg in choosing the appropriate actions in infrequently encountered situations.

In contrast, frequent elements can be used to provide examples of the agent's typical behavior, which can help to understand its learned strategy, or how the agent faces situations that are recurrent, as occurs in many video-game scenarios where the player has available multiple ``lives''.
	
\extraction{Frequent and infrequent situations correspond to states that were visited less or more frequently than others during interaction, respectively. We can extract these elements according to whether the frequency in the tables $\cs(s)$, $\cs(s,a)$ and $\cs(s,a,s')$ is below/above a given threshold, or by selecting the top/bottom $k$ states.}

\subsection{Execution Certainty Analysis}
\label{Sec:ExecCertainty}

This dimension can be used to reveal the agent's confidence in its decisions. Two complementary elements are extracted: \textbf{certain executions} and \textbf{unertain executions}.

Situations where the agent is uncertain of what to do indicate opportunities to ask a user for help, or indicate parts of the state space in which the agent requires more training. For example, in a combat scenario where the agent controls a group of units, the distribution over actions might be more even if the agent encounters a larger opposing force than encountered during training, meaning that the agent is not sure of what action to select. Uncertain situations are also significant because people tend to require explanations mostly for unusual behavior \citep{Miller2019}. 

In contrast, certain situations denote parts of the task that the agent has learned well. These can also be used to reveal to a user the agent's behavior in situations where it is more predictable.

\extraction{This dimension estimates the (un)certainty of each state with respect to action execution. Observations where many different actions are executed often (high dispersion) are considered uncertain, while those in which only a few actions are selected are considered certain. Given a state $s$, the execution certainty associated with $s$ is measured by the concentration of the executions of the actions $a\in\A$. We use the \emph{evenness} of the distribution over next states as given by its normalized true diversity (information entropy) \citep{Mulder2004}. Formally, let $p(X)$ be a probability distribution over $x_{i}\in X, i=1,\ldots,N$ of set $X$. The evenness of $p$ over $X$ is provided by: 

\begin{equation}\label{Eq:Evenness}
	\evenness(X)=-\sum_{x_{i}\in X^{+}}{p(x_{i})\ln{p(x_{i})}} / \ln{N},
\end{equation}
where $X^{+}\doteq \forall_{x_{i}\in X} \colon p(x_{i})>0$. We then use this evenness measure to calculate the dispersion of the distribution over actions according to $\evenness_{s}=\evenness(\pi(s))$, where $\pi$ is any policy of interest. 

In our study we approximate the agent's interaction policy using $\hat{\pi}(s)=\csa(s,\A)/\cs(s)$. We argue that this formulation retains information about the agent's history of interaction beyond the learned ``optimal'' policy by, \eg capturing situations that were harder to learn. However, one could use the policy directly derived by the algorithm, in the case of policy-based methods. Similarly, we could compute the evenness based on the learned (normalized) $Q$-values instead.}

\subsection{Transition-Value Analysis}

This dimension aims at enabling users to understand the desirability attributed by the agent to a given situation. The goal is to analyze how the value attributed to some state changes with respect to possible states visited at the next time-step. The elements extracted by this analysis are \textbf{minima} and \textbf{maxima} situations. 

Maxima may denote acquired preferences or \emph{favorable} situations for the agent. An example of such a situation can occur when the agent ``wins'' a game or moves to a higher level, in which case it has achieved the maximal value situation. Another situation can occur in more dynamic settings. \Eg in a game like Asteroids, an agent destroying an asteroid might represent a (local) maximum but, inevitably, the agent will encounter new asteroids to be destroyed, and will no longer be in a maximum state. Still, in such situations it achieves a locally-maximal value and its policy will likely lead the agent to visit similar states in the future. Therefore, presenting maxima elements might allow a user to observe the agent performing well in the task according to what it has learned.%
\footnote{Although our experimental study informed us that maxima and minima often occur when the agent performs the task well/poorly, additional experiments will be required to further characterize \emph{all} situations in which they might occur.}
This lets users verify whether performance is in line with expectations, or that adjustment of the agent's training, reward function, perceptual characteristics, etc. is needed. 

In contrast, minima denote highly \emph{adverse} situations that the agent wants to avoid and where taking any action leading to a different state is preferable. They can also denote undesirable situations visited by the agent not because of its behavior but due to the stochastic nature of the environment. By observing the agent's behavior in such situations, a user can understand how well the agent handles difficult situations, and whether further adjustments are required.

\extraction{This analysis combines information from the agent's estimated $V(s)$ function and the transition function $\hat{\P}(s'\mid s,a)$. These refer to states whose values are lower/greater than or equal to the values of all possible next states, as informed by the agent's learned predictive model. Formally, let $\trans_{s} \doteq \set{\forall_{s'\in\S} \colon \exists_{a\in\A}{\hat{\P}(s'\mid s,a)>0}}$ be the set of observed transitions starting from state $s$. The \emph{local minima} are defined by $\S_{min} \doteq \forall_{s\in\S} \colon \forall_{s'\in\trans_{s}}{V(s)\leq V(s')}$. Similarly, the \emph{local maxima} are defined by $\S_{max} \doteq \forall_{s\in\S} \colon \forall_{s'\in\trans_{s}}{V(s)\geq V(s')}$.}

\subsection{Sequence Analysis}

The goal of this analysis is to calculate interesting sequences of actions learned by the agent. In particular, these sequences of agent behavior start and end with important states as identified by the other analyses. Each sequence is associated with a certain likelihood of occurring. In the experiments reported in this paper, we specifically selected sequences starting from a (local) minimum, and then performing actions until a (local) maximum is observed, as they they provide good indicators of an agent's aptitude in the task. We thus refer to this element as \textbf{most likely sequences to maxima}. Alternatively, the initial and final state sets can include the (in)frequent, (un)certain, or other states of interest discovered by the different analyses. Each sequence would then have a different semantic meaning attached to it, revealing the agent's behavior when transitioning to distinct situations.

This element can be used to reveal more purposeful agent behavior to a human user, \eg performing actions from a low-valued situation to a learned maximum. This avoids requiring the user to view complete episodes while providing potentially more information for understanding the agent's behavior compared to viewing performance in the context of a single element. 
The extracted sequences can also be used by the user to query the agent about its future goals and behavior in \emph{any} possible situation. Namely, given the flexibility of the sequence-finding procedure, the initial and final sets can be user-defined. This would allow providing the user with contrastive explanations about why the alternatives to some actions---the foils---are not as desirable as those chosen by the agent \citep{Miller2019}. The starting points may denote the \emph{origins} of behavior while the sequence likelihood and the target states denote the agent's \emph{beliefs} and \emph{reasons}, respectively---two crucial elements people use to explain intentional events \citep{DeGraaf2017}.

\extraction{We first create a state-transition graph where nodes are states $s\in\S$ and edges are actions $a\in\A$ denoting the observed transitions, weighted according to the probability $\hat{\P}(s'\mid s,a)$. The resulting graph is directed with nonnegative weights, making it amenable to best-first search algorithms to find the most likely sequence of state-action pairs between two given states. 

We then use a variant of \citeauthor{Dijkstra1959}'s algorithm \citep{Dijkstra1959}, whose input is an initial state $s_{i}\in \S_{i}$, where $\S_{i}$ is a state set of interest, and a set of possible final states, $\S_{f}$. First, we determine the most likely paths between $s_{i}$ and each target state $s_{f}\in\S_{f}$. Let $P_{if}=\vect{s_{0}=s_{i},a_{1},s_{1},\ldots,a_{n}, s_{n}=s_{f}}$ denote a path between $s_{i}$ and $s_{f}$. The probability of the agent visiting $s_{f}$ after $s_{i}$ and following path $P_{if}$ is thus given by $p(s_{i},s_{f})=\sum_{t=1}^{\card{P_{if}}}{\hat{\pi}(s_{t-1},a_{t})\hat{\P}(s_{t}\mid s_{t-1},a_{t})}$, where the probability of selecting actions, given by $\hat{\pi}$, can be estimated as described in Sec.~\ref{Sec:ExecCertainty} for the extraction of execution certainty. After calculating the paths' probabilities, we then choose the \emph{most likely path weighted by value}, connecting the initial state and an optimal final state, according to: $s_{i}^{*}=\argmax_{s_{f}\in\S_{f}}{p(s_{i},s_{f}) V(s_{f}})$.}

\section{Visual Summaries}%
\label{Sec:Summaries}

As depicted in Fig.~\ref{Fig:Framework}, provided that the behavior of the agent can be readily visualized, the interestingness elements extracted by our analyses can be used to generate \emph{visual summaries}. In particular, we follow an approach similar to \citep{Amir2018}, where short video-clips highlighting the behavior of the agent during key moments of interaction are recorded. This type of behavior summarization is therefore suitable for agents relying on a visual input such as robots or game-playing agents. Nonetheless, as noted in \citep{Amir2018} the extraction of summaries will depend on the characteristics of the specific task domain.

The rationale behind this approach is that simply watching the behavior of the agent performing the task, \eg during whole episodes, might be burdensome for a user and mostly uninformative. Therefore, the system is given a \emph{budget} for a given summary. The budget will typically be application-dependent and can be defined in terms of, \eg the maximum length of the video clip or the maximum number of time-steps or highlights per summary. We generate two types of video highlights in our framework: summaries and sequences.

\subsection{Summary Highlights}

These highlights combine different examples of the agent's behavior. Like the approach in \citep{Amir2018}, given some history of interaction with an environment, \eg a sequence of simulated episodes, we summarize agent behavior by selecting $k$ highlights (agent trajectories) of length $l$ time-steps, where $(l-1)/2$ time-steps occur before and after the situation of interest. The highlights are then combined into a single video clip for visualization.

Departing from \citep{Amir2018}, we introduce novel ways in which the key moments are selected and presented to a user. First, as mentioned earlier, each interestingness element is defined according to some analysis dimension. For example, frequency elements can be sorted from the most to the least frequent states. During highlight selection, we try to sample situations closer to the extrema---\eg when highlighting frequent states, a trajectory passing through the most frequent state is the most preferred; in the case of the sequences dimension, we select the largest sequence. The idea is to select the most representative and informative examples regarding each source of interestingness. 

For summaries where $k>1$, we use a secondary selection criterion of \emph{diversity}, which is similar to the HIGHLIGHTS-DIV approach in \citep{Amir2018}, where here we use a different example selection procedure. Namely, whenever the maximum number of $k$ highlights for a type of element has been reached, the framework decides whether to replace any of the existing highlights with the new state. Formally, let $d(s_1, s_2)\in[0,1]$ be an state diversity/distance metric relevant to the task, \eg how visually distinct the states are, or how distant in time the states were visited. Then, given the $k+1$ highlights, we keep the set of $k$ highlights maximizing:

\begin{equation}%
\label{Eq:HighlightDiversity}
	\max_{s_{i},s{j}}d(s_i, s_j)\times \min_{s_{i},s{j}}d(s_i, s_j), \quad i\neq j, \quad i,j=0,\ldots,k-1
\end{equation}
The idea is to maximize the total diversity of the final set of highlights such that the user can observe, \eg different nuances of the agent's learned behavior, or contextually distinct situations. Therefore, Eq.~\ref{Eq:HighlightDiversity} aims at including maximally-separated examples by maximizing both the maximal and the minimal distances between pairs of examples to be included in a summary.

Finally, we introduce the step of deciding how to visualize and compile the highlights. When experimenting with visualization, we noticed that, depending on the task, the user sometimes fails to understand which part of the agent's behavior is being highlighted. This can be critical in highly dynamic environments, where the behavior of the agent can vary dramatically at each time-step. To mitigate this problem, we introduce fade-in/fade-out effects around each highlight so that the user can focus on the important moment being highlighted while also being presented with the context in which it occurred.

\subsection{Sequence Highlights}

Our framework provides an additional type of highlighting capability based on the element of likely sequences to favorable situations, which enables highlighting relevant sequences of behavior rather than single moments. In particular, in our user study we focus on sequences starting in a \emph{minimum} and ending in a \emph{maximum}, as informed by the transition-value analysis. This reveals examples of the agent's learned strategy, \ie a trajectory showcasing its capabilities in overcoming perceived difficulties and in reaching its learned preferences.

\section{Experimental Study}%
\label{Sec:Study}

\begin{figure}[!tb]
	\centering
	\centering
        \begin{subfigure}[b]{0.4\columnwidth}
            \includegraphics[width=\textwidth]{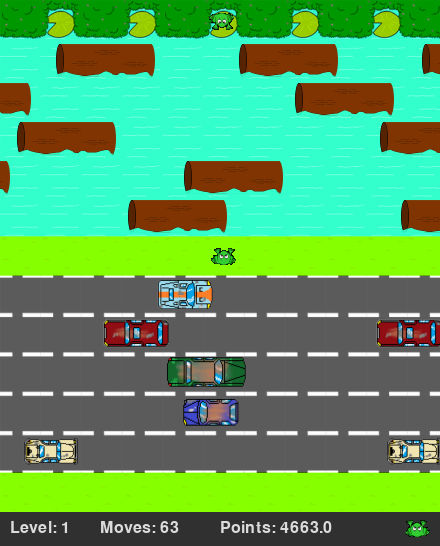}
            \caption{Screenshot}%
            \label{Fig:Frogger}
        \end{subfigure}%
        \hspace{10pt}
        \begin{subfigure}[b]{0.45\columnwidth}
        	    \centering
            \begin{subfigure}[b]{0.5\textwidth}
                \includegraphics[width=\textwidth]{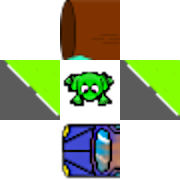}
                \caption{Observation features}%
                \label{Fig:Features}
            \end{subfigure}
            \\\vspace{4pt}
            \begin{subfigure}[b]{0.8\textwidth}
                \includegraphics[width=\textwidth]{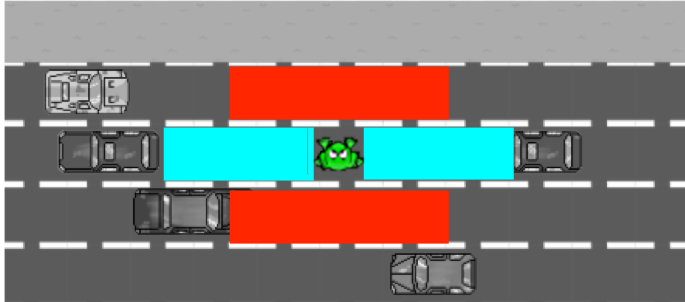}
                \caption{Vision range}%
                \label{Fig:Vision}
            \end{subfigure}
        \end{subfigure}
    	\caption{The Frogger game task (left) and agent design aspects (right).}%
    	\label{Fig:Task}
\end{figure}

To evaluate the usefulness of our XRL framework in helping humans correctly understand the aptitude of different RL agents,
we conducted a user study using a simple video-game scenario based on Frogger.%
\footnote{We used the implementation in \url{https://github.com/pedrodbs/frogger}.}
Fig.~\ref{Fig:Frogger} shows a screenshot of the game. The player is responsible for controlling frogs, one at a time, to go from the grassy strip at the bottom to the lilypads at the top, with only one frog at a time allowed on a lilypad. The player has to first cross the road without getting hit by cars and then cross the river by jumping on logs. The control set is $\A=\set{N,S,E,W}$, where each action deterministically moves the frog by $40$ pixels in the corresponding direction. 

When a frog reaches a lilypad, a new frog appears at the bottom of the screen. When two frogs reach the lilypads, the agent completes the current level and starts the next level. This clears the lilypads and makes cars and logs go faster. A player has a limit of $100$ moves to pass a level and $3$ lives in a game. The player loses a life when the maximum number of moves is reached, a frog is hit by a car, a frog jumps into the river, or a frog is on a log that goes off-screen. At the beginning of each game, the time interval at which logs are introduced in the environment is randomized.

\subsection{Agent Experiments}

To make the Frogger game an interesting task from the perspective of behavior explanation, we designed agents with \emph{partial observability}. Specifically, each state $s\in\S$ provides a local view of the environment according to the frog's location. States are represented by four discrete features, denoted by $s=\vect{\phi_{N},\phi_{S},\phi_{E},\phi_{W}}$, each $\phi_{i}\in\Phi$ indicating the element that is visible in the corresponding direction, where $\Phi=\set{empty, water, car, log, lilypad, bounds}$ is the feature set, $bounds$ means that the agent is near the environment's borders, and $empty$ that no element is visible in that direction.%
\footnote{The agent does not distinguish between the road and grass for feature $empty$, nor the type of car for the $car$ feature.}
 
This partial observability introduces uncertainty in the environment's dynamics \emph{from the agent's perspective}. Fig.~\ref{Fig:Features} shows what an agent observes given the true game state in Fig.~\ref{Fig:Frogger}. As we can see, $\phi_{S}=car$ although the car is not directly beneath the frog. This has to do with the way we designed the perception of cars, where two parameters control for an agent's vision range, as illustrated in Fig.~\ref{Fig:Vision}. The blue squares indicate the agent's \emph{horizontal vision range}, $vis_{h}$, defined as the horizontal distance (in pixels) within which an agent can see cars in its $E$ and $W$ directions, \ie a car in the same lane. Red squares indicate the \emph{vertical vision range} $vis_{v}$, defined by the vertical distance (in pixels) within which an agent can see cars to its $N$ and $S$, \ie in the lanes \emph{directly} above and below the agent, respectively. %
%
%
As for the logs, we designed the $log$ feature such that an agent anticipates the movement of adjacent logs.%
\footnote{The agent's observations are extracted without access to the game engine's internal dynamics, including the logs' velocities, hence they are not perfect. This means there is still a chance ($p>0$) that performing action $N$ when $\phi_{N}=log$ will result in a death in the river. }

The reward function $\Rwd$ was defined as follows. The agent receives a reward $r(s,a)=-200$ in situations where executing $a$ when observing $s$ results in one of the aforementioned death conditions. A punishment of $-300$ occurs when the agent depletes its available lives and a reward of $5{,}000$ is received whenever a frog arrives on a lilypad. A reward of $-1$ is also received at each time-step.

\subsubsection{Agent Types}

To study the impact of the different interestingness elements in helping understand the aptitude of RL agents, we designed three different agents:
\begin{description}
	\item[Optimized:] this agent can observe cars at a moderate distance, \ie when they are neither too close nor too far away. Its parameters were empirically fine-tuned to achieve a very high performance in the task. The agent uses the default reward function.
	\item[High-vision:] this agent has very high vision capability, \ie it can anticipate the presence of cars from afar. It was designed to simulate an agent with inadequate perceptual capabilities, \ie whose sensing mechanisms were poorly calibrated to the task. It also uses the default rewards.
	\item[Fear-water:] this agent has the same visual capabilities as the optimized agent. It also uses the default reward function with one exception: when it dies on the river, it receives a punishment of $r=-10{,}000$. This simulates a motivational impairment---the rewards idealized by its designer are not appropriate for the agent to learn the intended task.
\end{description}
A detailed parameterization of each agent is listed in Table~\ref{Tab:Agents}.

\begin{table*}[!tb]
    \centering
    \caption{The parameterization of each agent in our experiments. Legend: $r_{river}$: reward received after a death in the river; $q_{init}$: $Q$-values initialization. See text for more details.}%
    \label{Tab:Agents}
    \begin{tabular}{l | r r r r }
    \toprule
    \textbf{Agent}
    & $\pmb{vis_{h}}$	
    & $\pmb{vis_{v}}$	
    & $\pmb{r_{river}}$	
    & $\pmb{q_{init}}$ \\
    \midrule
    \textit{Optimized} 	
    & $60$		& $40$	& $-200$		& $5{,}000$	\\
    \textit{High-vision} 
    & $140$	& $120$	& $-200$		& $5{,}000$	\\
    \textit{Fear-water}	
    & $60$		& $40$ 	& $-10{,}000$	& $0$       \\
    \bottomrule
    \end{tabular}
\end{table*}

\subsubsection{Agent RL Training}

Each agent was first trained using standard $Q$-learning \citep{Watkins1992} for $2{,}000$ episodes,%
\footnote{Source code of the interestingness elements framework, the agents and algorithms used in the experimental study can be found at \url{https://github.com/SRI-AIC/InterestingnessXRL}.}
with each episode corresponding to one game, \ie ending when the agent lost all of its $3$ lives or when a limit of $300$ time-steps was reached. We used a Softmax action-selection mechanism with exponentially-decaying temperature $\beta$, \ie given training episode $e$, the temperature is given by $\beta=\beta_\text{min} + \beta_\text{max}0.995^{e}$. In our experiments, we set $\beta_\text{min}=0.05$ and $\beta_\text{max}=20$. To promote exploration, we used optimistic initialization by setting $Q(s,a)=q_{init}$ for all states $s\in\S$ and actions $a\in\A$, where $q_{init}$ controls the agent's optimism. As listed in Table~\ref{Tab:Agents}, unlike other agents, the Fear-water agent was initialized with $q_{init}=0$ to further prevent it from reaching lilypads as a result of pure exploration. After training, the learned policy was tested for another $2{,}000$ episodes by setting $\beta=\beta_\text{min}$, resulting in greedy action-selection.%
\footnote{The performance of each agent \emph{during learning} is detailed and discussed in Appendix~\ref{Sec:AppendixAgentExperiments}.}

\subsubsection{Agent Performance Results}

\begin{table*}[!tb]
    \caption{The performance of each agent in $2{,}000$ test episodes. 
    Legend: River: agent jumped into river; Car: agent was hit by a car; Time: maximum moves to pass a level ($100$) reached.}
    \label{Tab:ResultsPerf}
    \centering
    \begin{tabular}{l | 
    r@{ $\pm$ }r | 
    R{30pt} R{30pt} R{30pt} : R{30pt} | 
    r@{ $\pm$ }r}
    \toprule
    \multirow{2}{*}{\textbf{Agent}} 
    & \multicolumn{2}{c|}{\textbf{Mean}}
    & \multicolumn{4}{c|}{\textbf{Number of Deaths}}
    & \multicolumn{2}{c}{\textbf{Mean}} \\
    & \multicolumn{2}{c|}{\textbf{Level}}
    & \textbf{River} & \textbf{Car} & \textbf{Time} & \textbf{Total} 
    & \multicolumn{2}{c}{\textbf{Steps}}\\
    \midrule
    \textit{Optimized} 	
    & $4.6$ 	& $0.7$
    & $2{,}395$ 	& $1{,}051$	& $14$      & $3{,}460$
    & $288$ 	& $26$ \\
    \textit{High-vision} 
    & $2.0$ 	& $0.4$	
    & $411$ 	& $888$	    & $3{,}027$   & $4{,}326$
    & $291$ 	& $22$ \\
    \textit{Fear-water}	
    & $1.0$ 	& $0.0$	
    & $0$ 		& $3{,}604$	& $2{,}396$   & $6{,}000$
    & $222$ 	& $46$ \\
    \bottomrule
    \end{tabular}
\end{table*}

\begin{figure*}[!tb]
	\centering
        \begin{subfigure}[b]{.47\textwidth}
            \includegraphics[width=\textwidth]{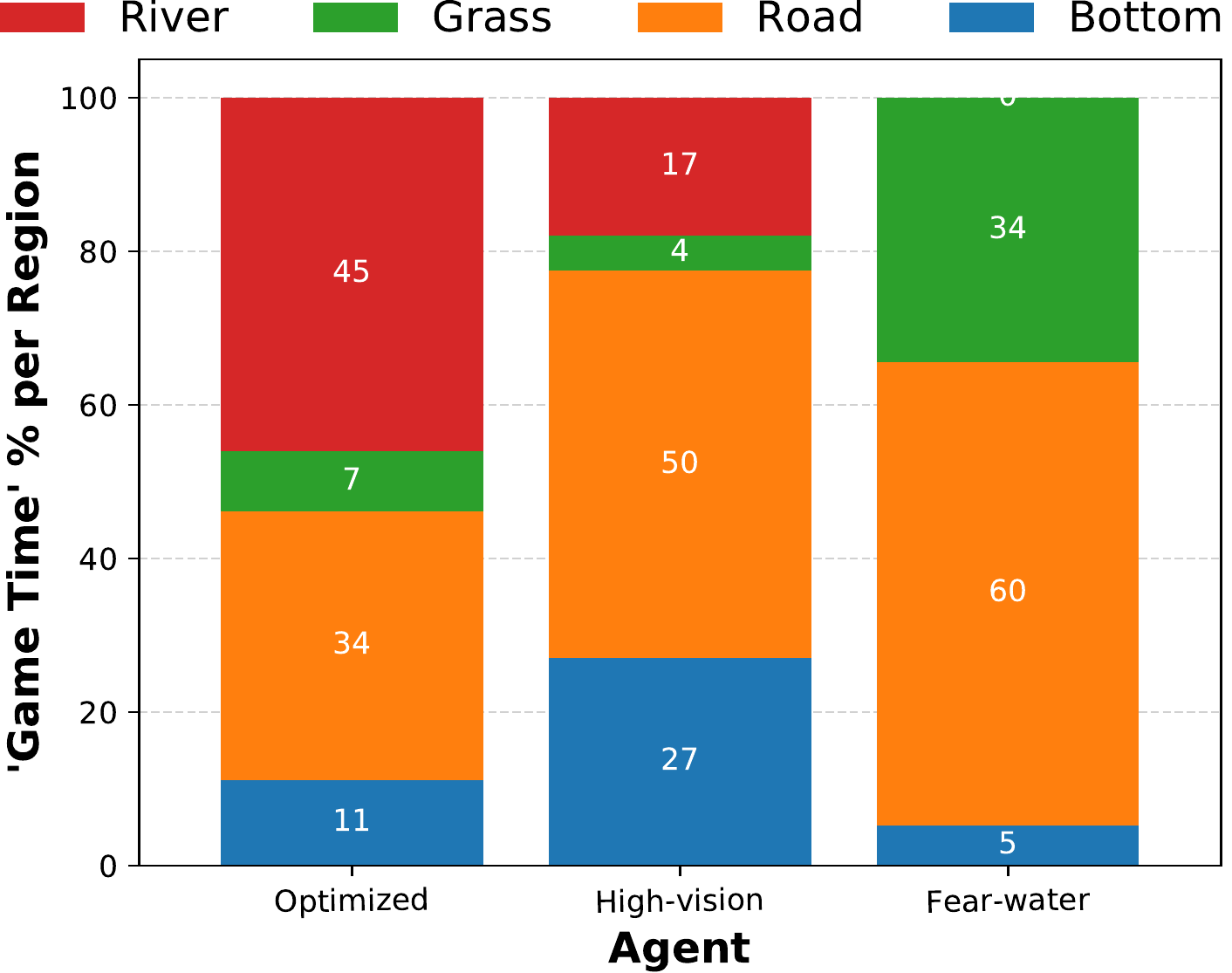}
            \caption{\% of game time}%
            \label{Fig:GameTime}
        \end{subfigure}%
        \hfill
	\begin{subfigure}[b]{.47\textwidth}
            \includegraphics[width=\textwidth]{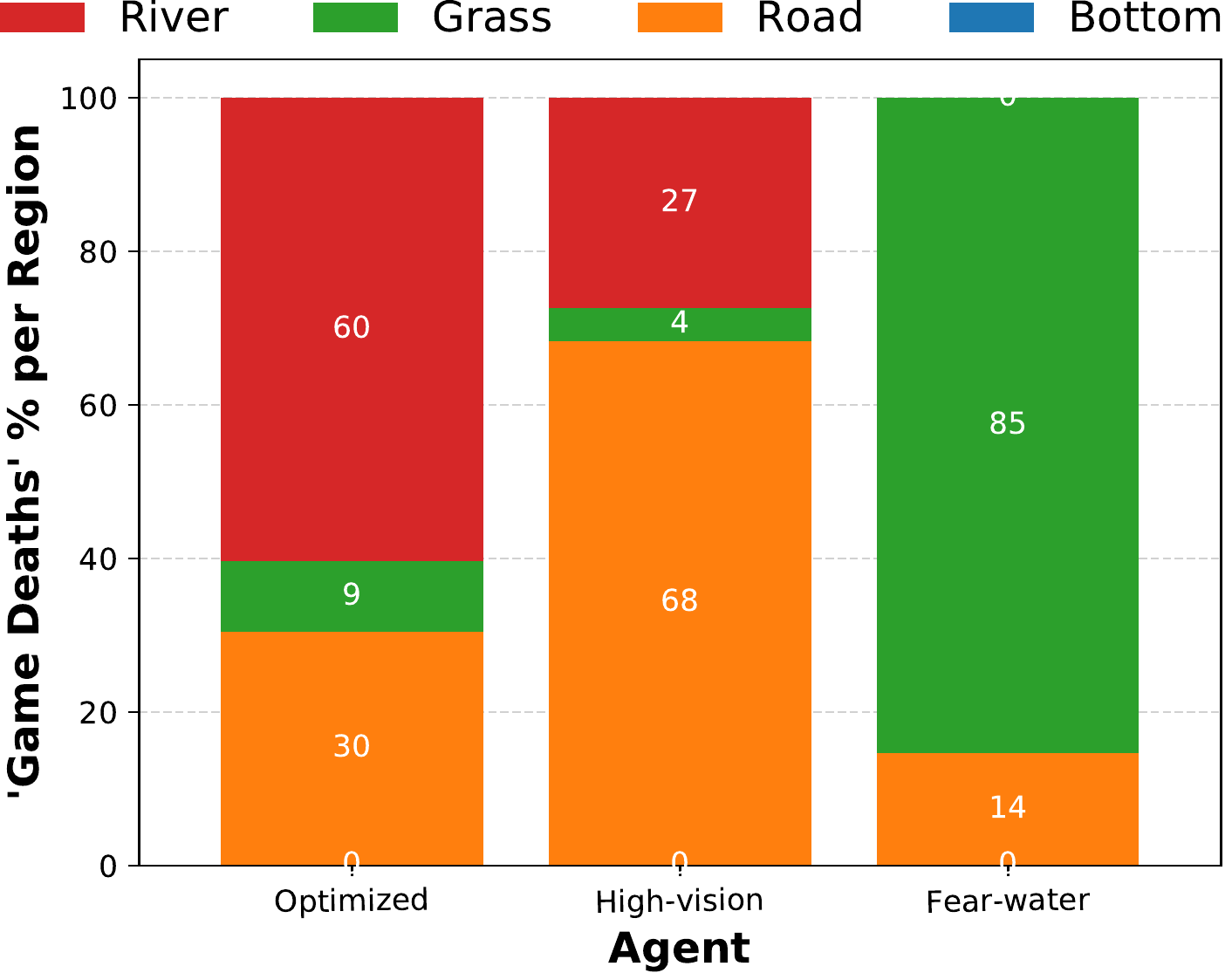}
            \caption{\% of game deaths}%
            \label{Fig:GameDeaths}
        \end{subfigure}%
    	\caption{The percentage of time spent and deaths occurring in each environment region over $2{,}000$ test episodes (\emph{Grass:} middle grassy strip; \emph{Bottom:} grassy strip on the bottom).}%
    	\label{Fig:RegionPerfResults}
\end{figure*}

Summary statistics of the agents' performance during the $2{,}000$ test episodes after learning are listed in Table~\ref{Tab:ResultsPerf}. Fig.~\ref{Fig:RegionPerfResults} depicts the agents' performance relative to each region of the environment. Our goal was to design the agents such that, after training, their underlying characteristics and limitations would lead to contrasting behaviors. As we can see, the different characteristics indeed resulted in the agents attaining different performances in the task. 

The optimized agent achieves the best performance among the three agents, reaching significantly higher levels. Compared to the high-vision agent, we see that this is due to the differences in the vision range parameters. The high-vision agent can see cars at a higher distance, resulting in more cautious behavior while crossing the road. While this leads the agent to spend more time on the road, it gets hit by cars significantly less than other agents. 

For the fear-water agent, receiving a highly negative reward from dying in the river (two times the magnitude of the lilypad reward) prevents it from reaching any lilypad. The agent was also not optimistically initialized, making it less motivated to explore and inadvertently discover the lilypads. We see in Fig.~\ref{Fig:RegionPerfResults} that the agent usually dies by getting hit by a car from the middle grass row. This is because the value of jumping into the river (action $N$) is significantly lower than that of any other action, \ie the agent ``fears'' falling into the water more than anything else.%
\footnote{We note that this agent is still able to cross the road and avoid cars but could not jump to logs when reaching the river. This agent serves as a baseline for assessing the usefulness of the interestingness elements in identifying behavioral impairments.}

\subsection{Introspection and Visual Summaries}

Referring back to Fig.~\ref{Fig:Framework}, an agent's history of interaction in our experiments comprises a total of $4{,}000$ episodes ($2{,}000$ training and $2{,}000$ testing). During that time, all the interaction data described in Sec.~\ref{Subsec:InteractionData} was collected. As mentioned earlier, the goal was to capture both the challenges encountered by the agents during learning and aspects of their learned policy. We then applied our introspection framework to extract the interestingness elements in Table~\ref{Table:Elements} for each agent.%
\footnote{Appendix~\ref{Sec:AppendixAgentExperiments} provides statistics on the interestingness elements captured for each agent.}

After introspective analysis, visual summaries were produced by generating video clips highlighting the agents' performance according to the procedures described in Sec.~\ref{Sec:Summaries}. The summaries were captured during the $2{,}000$ test episodes.%
\footnote{While introspection was performed over all episodes, summaries were extracted only from the test episodes. This was both to avoid capturing mistakes due to the training process itself (exploration) and not because of the agents' inherent limitations, and to show how the agent learned to deal with potentially rare or uncertain situations that occurred during training.}
For the diversity metric $d(s_1, s_2)$ over highlights, we used the absolute difference of game score between the moments in which $s_1$ and $s_2$ were observed, the goal being to capture the agents' behavior at various stages of the game.

\subsection{User Study}

To assess whether the visual examples derived from the different interestingness elements would help human users correctly understand the different agents' underlying capabilities and limitations, we conducted a user study.
\footnote{The study was approved by SRI International's Institutional Review Board (IRB).}
We recruited participants using the psiTurk tool \citep{McDonnell2012} for Amazon Mechanical Turk (MTurk). A total of $82$ participants ($40\%$ female, age group mode $[25-34]$ years) participated in the study, each receiving $\$7.50$ for their completion of the Human Intelligence Task (HIT).

\subsubsection{Summarization Techniques}

\begin{table}[!tb]
    \caption{Composition and budget of the summarization techniques used in the user study.}%
    \label{Tab:Highlights}
    \centering
    \begin{tabular}{l l r}
    \toprule
    \textbf{Name}	& \textbf{Summary Composition}		& $\pmb{k}$ \\
    \midrule
    \textsf{Max}	    		& all maxima						& 4          \\
    \textsf{Min}	 		& all minima						& 4          \\
    \textsf{Cert}			& all certain execution states    			& 4          \\
    \textsf{Uncert}		& all uncertain execution states  		& 4          \\
    \textsf{Freq}			& all frequent states    		    		& 4          \\
    \textsf{Infreq}			& all infrequent states    	    			& 4          \\
    \hdashline
    \textsf{Max-Min} 		& half minima, half maxima      			& 4          \\
    \textsf{Cert-Uncert}	& half certain-execution, half uncertain	& 4          \\
    \textsf{Freq-Infreq}	& half frequent, half infrequent	    		& 4          \\
    \hdashline
    \textsf{All}			& one of each type				        & 6          \\
    \hdashline
    \textsf{Seq}		& sequence from minimum to maximum		& 1          \\
    \bottomrule
    \end{tabular}
\end{table}

Our objective was to understand the differential contribution of each interestingness element to understanding an agent's behavior. We wanted to know whether single elements are sufficient for correctly perceiving an agent's aptitude or if a combination works better to convey certain behavioral characteristics. 

Table~\ref{Tab:Highlights} lists the \emph{summarization techniques} used in our study. The first six techniques generated summaries composed of highlights for a single type of interestingness element. The goal was to assess how each type exposed different characteristics of the agents' behavior. The second group of techniques generated videos composed of trajectories highlighting the extrema of each dimension of analysis, enabling us to evaluate the explanatory power of each dimension. The \textsf{All} technique, given a higher budget, generated highlights including $6$ examples with the goal of assessing the effect of including one highlight per element. Finally, the \textsf{Seq} technique produces a single sequence highlight to determine whether a single trajectory is sufficient to reveal subtleties underlying the agents' behavior.

We used a trajectory length of $l=21$ time-steps ($\approx5$ secs.) for each highlight in a summary and a maximum length of $80$ time-steps for sequence highlights. To avoid biases during the observation of highlights by human users, we removed all game information at the bottom of the screen from the videos, \ie the level, remaining moves, points and available lives.%
\footnote{The videos used in the study are at \url{https://github.com/pedrodbs/frogger-study}.}

\subsubsection{Conditions}

Each condition of our study corresponds to a \emph{scenario} where we assess the subjects' perception of all three agents' characteristics given one summarization technique. Since we have $11$ techniques and three agents, we have two independent variables: \emph{agent}, with $3$ levels, and \emph{scenario}, with $11$. Informed by a pilot study, we determined that requiring subjects to watch videos for all players and all summarization techniques (a full within-subjects design), would be too onerous. On the other hand, getting accustomed to the task requires time and effort, which made exposing each subject to only one condition (between-subjects design) equally undesirable. Therefore, we opted for a partially-balanced, incomplete block experimental design \citep{Bose1953}, where each subject was exposed to $6$ randomly-selected scenarios, and in each scenario subjects were exposed to the behavior of all three agents. 

\subsubsection{Experimental Procedure}

After accepting the task, subjects started by agreeing to a consent form explaining the goals and protocol of the study. They were then redirected to a page describing the dynamics and rules of Frogger, after which they were presented with a $5$-question quiz about the game.%
\footnote{Screenshots of the several pages and transcriptions of all the instructions, questions and options are included in Appendix~\ref{Sec:AppendixSurveyMaterials}.}
Failed questions were highlighted in red to reinforce the main concepts, but all participants were allowed to continue. The primary goal was to ensure that subjects understood the dynamics underlying the task (\eg that reaching a new level speeds up cars and logs) rather than to prevent them from participating in the study. 

\begin{figure}[!tb]
	\centering
	\includegraphics[width=0.8\columnwidth]{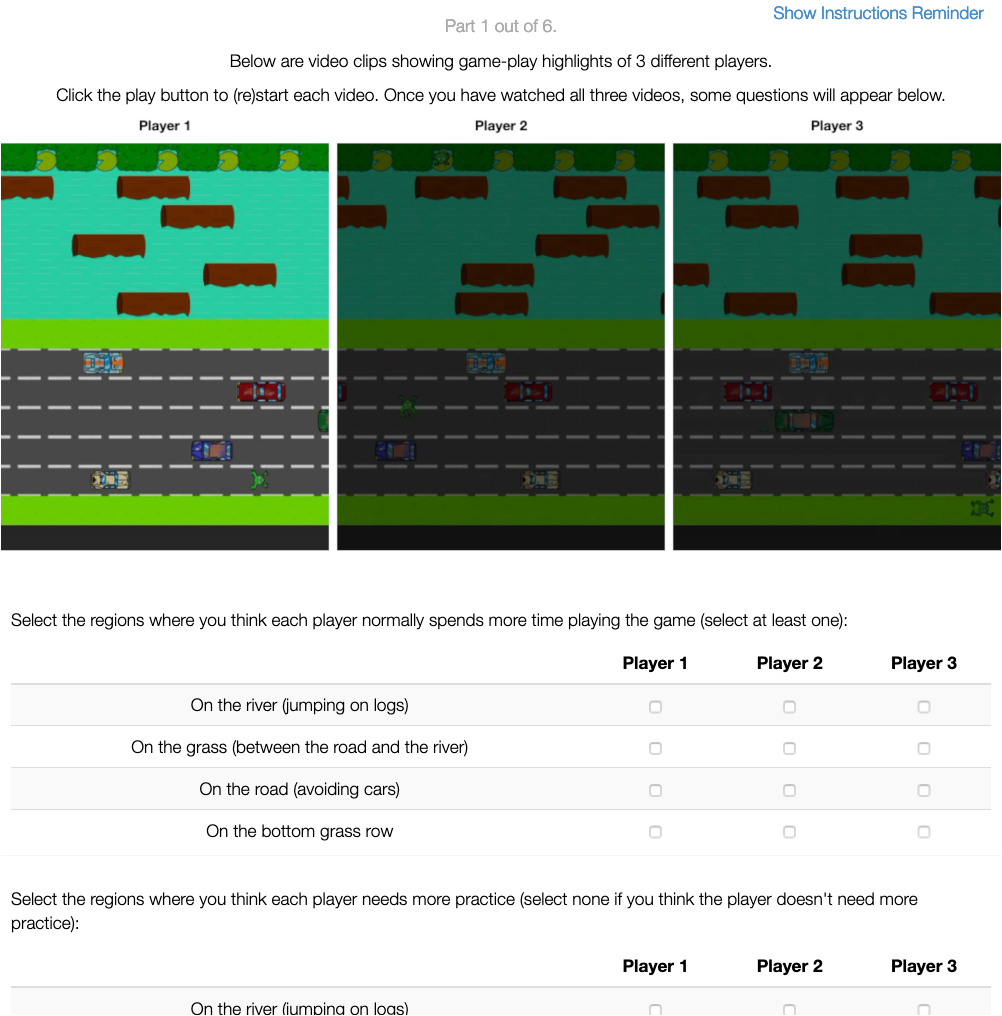}
	\caption{Questionnaire used in the user study.} 
	\label{Fig:Scenario}
\end{figure}

An instruction page was then presented that explained what the participants had to do in each of the $6$ scenarios. As seen in Fig.~\ref{Fig:Scenario}, each scenario showed videos of the three agents in parallel. The relative location of the video for each agent in the page (left, middle, right) was randomly assigned. As seen on the top of Fig.~\ref{Fig:Scenario}, participants were told that they were watching videos of the gameplay of $3$ different \emph{players}. Further, the instructions stated that the players were different for each scenario and that the order of appearance was not relevant. The reason for this is two-fold: first, we wanted subjects to believe they were watching the behavior of human players, and second, we wanted to avoid learning effects between scenarios.%
%

Subjects could start each video when they pleased and after playing all videos (enforced programmatically), a questionnaire with the survey questions appeared below the videos, as depicted in Fig.~\ref{Fig:Scenario}. We designed a set of questions for each scenario to assess subjects' perception of the agents' underlying characteristics and also properties of their learned behavior, given the scenario's summarization technique. 

We knew from the previous performance results (see Fig.~\ref{Fig:RegionPerfResults}) that the agents' capabilities in each region are quite different. Each region in Frogger provides different challenges and requires distinct learned skills to successfully overcome them. As such, we created two \emph{region-related} questions:
\begin{description}
	\item[Time:] \textit{``Select the regions where you think each player normally spends more time playing the game (select at least one)''}. 
	\item[Practice:] \textit{``Select the regions where you think each player needs more practice (select none if you think the player doesn't need more practice)''}.
\end{description}
These questions assess the subjects' ability to perceive the difficulty agents had in different regions of the game. \emph{Time} assesses relative presence while \emph{practice} evaluates difficulties experienced in each region. For each question subjects could select from \{\emph{river}, \emph{middle grass row}, \emph{road}, \emph{bottom grass row}\}. The outcome of each question is thus a Boolean variable for each region. 

We also designed two questions about the overall \emph{aptitude} of the agents:
\begin{description}
	\item[Level:] \textit{``The player is capable of reaching advanced levels''}. 
	\item[Help:] \textit{``The player needs help to be better at the game''}.
\end{description}
Subjects were asked to rate their agreement to each statement on a 5-point Likert scale ($1$ strongly disagree -- $5$ strongly agree). The goal was to evaluate the subjects' perception of the overall capabilities of each agent. In particular, \emph{level} estimates the perception of how well an agent plays the game, while \emph{help} evaluates whether subjects could correctly determine if an agent needed intervention to improve its performance. Subjects were instructed to answer questions with respect to what they inferred about the players given what they had seen in the video clips. The goal was for subjects to evaluate the players' aptitude beyond their performance in the videos.%
\footnote{Subjects were allowed to play, stop and rewind videos as many times as they wanted.}
A final question assessed the subjects' \emph{confidence} in their responses on a 5-point Likert scale ($1$ not confident -- $5$ very confident).
The outcomes of these questions are thus ordinal variables with values in $[1,5]$. 

After answering the questions for all $6$ scenarios, a final page collected subjects' demographics and general opinions (open question) about the study. Subjects then returned to MTurk to collect their compensation.

\section{Analysis and Results}%
\label{Sec:Analysis}

We are mainly interested in answering these two research questions:
\begin{description}
	\item[RQ1:] \emph{Does the information conveyed by the various summarization techniques induce a different perception of each agent's aptitude?}
	\item[RQ2:] \emph{What summarization techniques enable a correct perception of the agents' aptitude in the task?}
\end{description}

To begin, we removed the responses of subjects who did not seem to take the task seriously. On average, participants correctly answered $4.2\pm1.0$ out of $5$ quiz questions and took an average of $19.5\pm11.9$ min.~to complete the survey. We eliminated $5$ subjects who had $\leq2$ correct answers or spent less than $6$ min.~on the survey. The data for the remaining $77$ subjects resulted in each scenario being sampled $42\pm4$ times (min: $36$, max: $50$).

\subsection{Analysis of Perceived Agent Performance in Different Regions}

\subsubsection{Research Question 1}

To assess whether the different summarization techniques had an effect on subjects' perception of agent behavior in the different regions, we performed a $\chi^{2}$ (chi-squared) test over the sum of positive responses and calculated the effect size using Cram\'{e}r's $V$ with the bias correction in \citep{Bergsma2013}. A Bonferroni correction was then applied to test for significant pairwise differences.

\begin{table}[!tb]
    \caption{Significant differences in the responses of the region-related questions ($V$: Cram\'{e}r's $V$ statistic; Scenarios: scenarios with the most significant pairwise differences. Empty cells denote non-significant differences where $p\geq0.01$).}%
    \label{Tab:ResultsRegionDiff}
    \centering
    \begin{tabular}{l l r l r l}
    \toprule
    \multirow{2}{*}{\textbf{Region}} & \multirow{2}{*}{\textbf{Agent}} & \multicolumn{2}{c}{\textbf{\textit{Time}}} & \multicolumn{2}{c}{\textbf{\textit{Practice}}} \\
    & & \multicolumn{1}{c}{$\pmb{V}$} & \textbf{Scenarios} & \multicolumn{1}{c}{$\pmb{V}$} & \textbf{Scenarios}\\
    \midrule
    \multirow{3}[2]{*}{\textbf{River}} 
    & Optimized & $0.52$ & \textsf{Min} & $0.30$  & \textsf{Infreq}\\
    & High-vision & $0.65$ & \textsf{Min}, \textsf{Uncert}, \textsf{Freq} & & \\
    & Fear-water & & & & \\
    \hdashline
    \multirow{3}[2]{*}{\textbf{Grass}} 
    & Optimized & & & $0.26$ & \\
    & High-vision & & & $0.28$ & \textsf{Infreq}\\
    & Fear-water & $0.41$ & \textsf{Min}, \textsf{Uncert} & & \\
    \hdashline
    \multirow{3}[2]{*}{\textbf{Road}} 
    & Optimized & $0.50$ & \textsf{Min} & $0.46$ & \textsf{Seq}, \textsf{Cert} \\
    & High-vision & $0.42$ & \textsf{Uncert} & $0.49$  & \textsf{Infreq} \\
    & Fear-water & $0.36$ & \textsf{Max} & $0.35$  & \textsf{Min} \\
    \hdashline
    \multirow{3}[2]{*}{\textbf{Bottom}} 
    & Optimized & $0.19$ & & & \\
    & High-vision & $0.69$ & \textsf{Min} & $0.27 $ & \textsf{Infreq} \\
    & Fear-water & $0.42$ & \textsf{Uncert} & & \\
    \bottomrule
    \end{tabular}
\end{table}%
 
Table~\ref{Tab:ResultsRegionDiff} shows the significant effects ($p<0.01$) of the summarization technique on the subjects' responses found for each agent.%
\footnote{Appendix~\ref{Sec:AppendixUserResponses} contains plots depicting subjects' responses for the region-related questions.}
We also show which techniques had the most pairwise significant differences, as informed by the Bonferroni comparison---this indicates what interestingness elements and introspection dimensions exposed the most distinct characteristics of the agents. 
%
For the \emph{time} response, significant effects were found for all agents in the road and bottom grass regions, whereas for the \emph{practice} variable significant effects were found for all agents only in the road region.

\subsubsection{Research Question 2}

\begin{table}[!tb]
    \caption{Pairing of response variables and agent performance data (ground-truth).}%
    \label{Tab:RegionRespPerfPair}
    \renewcommand{\arraystretch}{1.2}
    \centering
    \begin{tabular}{L{60pt} L{260pt}}
    \toprule
    \textbf{Response variable} 	& \textbf{Agent performance variable} \\
    \midrule
    \textit{Time}	    			& Mean percentage of game time spent in the region. \\[2pt]
    \textit{Practice}	    		& Mean number of deaths occurred in the region (excludes lives lost due to maximum number of moves achieved). \\
    \bottomrule
    \end{tabular}
\end{table}

To  identify which summarization techniques allowed for a correct assessment of the agents' aptitude in each region, we first paired each response variable with a game performance variable, as listed in Table~\ref{Tab:RegionRespPerfPair}. The performance data serves as ground truth for the agents' behavior and characteristics against which we compare the subjects' interpretations.
Then, for each response variable, we transformed the vector containing the percentage of subjects' responses for each agent and summarization technique across all regions into a Boltzmann distribution.%
\footnote{\Eg if all subjects selected all regions in the \emph{practice} question for an agent in a given scenario, this means they believed that the agent needed to practice equally in each region, resulting in a probability of $1/4$ for each element of the response distribution.}
Similarly, we derived a Boltzmann distribution from the mean values of the corresponding performance variable, resulting in the expected relative performance of a given agent in each region.%
\footnote{\Eg for the \emph{practice} response variable, this corresponds to the probability of an agent losing a life in each region.}
To determine whether the subjects' responses diverged from the agents' actual performance (ground-truth) we used the Jensen-Shannon divergence (JSD) \citep{Fuglede2004}, which is essentially a distance metric with scores ranging from $0$ (identical distributions) to $1$ (maximally different). 

\begin{table}[!tb]
    \caption{JSD between response and performance variables for the region-related questions, for each scenario and agent (\lowdiv{blue:} low divergence, \highdiv{red:} high divergence).}%
    \label{Tab:ResultsRegionCorrect}
    \centering
    \begin{tabular}{l | rrr | rrr} 
    \toprule
    \multirow{2}{*}{\textbf{Scenario}} &
    \multicolumn{3}{c|}{\textbf{\emph{Time}}} &
    \multicolumn{3}{c}{\textbf{\emph{Practice}}} \\
    & \textbf{Opt.} & \textbf{High} & \textbf{Fear}
    & \textbf{Opt.} & \textbf{High} & \textbf{Fear}\\    
    \midrule
    \textsf{Max} & $0.09$  & $\highdiv{0.70}$ & $0.60$  & $\lowdiv{0.01}$ & $0.47$  & $0.54$ \\
    \textsf{Min} & $0.45$  & $0.60$ & $\lowdiv{0.03}$ & $0.35$  & $0.08$  & $\highdiv{0.80}$ \\
    \textsf{Cert} & $0.18$  & $\highdiv{0.70}$ & $0.15$  & $\lowdiv{0.02}$ & $0.63$ & $0.51$ \\
    \textsf{Uncert} & $0.37$  & $\lowdiv{0.04}$ & $0.18$  & $0.29$  & $0.18$ & $0.61$ \\
    \textsf{Freq} & $\lowdiv{0.03}$ & $0.55$  & $0.07$  & $\lowdiv{0.02}$ & $0.16$ & $0.60$ \\
    \textsf{Infreq} & $\lowdiv{0.05}$  & $\highdiv{0.77}$ & $0.57$  & $\lowdiv{0.00}$ & $\highdiv{0.75}$ & $\highdiv{0.73}$ \\
    \hdashline
    \textsf{Max-Min} & $0.11$  & $0.33$  & $0.35$  & $0.47$  & $\lowdiv{0.03}$ & $\highdiv{0.66}$ \\
    \textsf{Cert-Uncert} & $0.18$  & $0.14$  & $0.13$  & $0.35$  & $0.23$  & $\highdiv{0.74}$ \\
    \textsf{Freq-Infreq} & $\lowdiv{0.03}$ & $0.30$  & $0.50$  & $\lowdiv{0.01}$ & $\lowdiv{0.04}$ & $0.44$ \\
    \hdashline
    \textsf{All} & $0.22$ & $0.11$  & $0.18$  & $0.36$  & $0.15$  & $0.61$ \\
    \hdashline
    \textsf{Seq} & $\lowdiv{0.01}$ & $0.37$  & $0.27$  & $\lowdiv{0.01}$ & $0.32$  & $0.62$ \\
    \bottomrule
    \end{tabular}
\end{table}

Table~\ref{Tab:ResultsRegionCorrect} presents the results of this analysis. We highlight the summarization techniques that resulted in either a very correct (low JSD) or very incorrect (high JSD) interpretation of the agents' performance relative to each region.%
\footnote{We note that the JSD thresholds used in our analysis, marked with the red and blue colors, are used only for the purpose of isolating the relevant results of our study.}

\subsection{Analysis of Perceived Overall Agent Aptitude}

\subsubsection{Research Question 1}

To assess whether the summarization techniques resulted in different perceptions of the agents' aptitudes, we modeled the response for each agent as an ordinary least squares (OLS) regression model and performed a Kruskal-Wallis H-test. Effect sizes were calculated using the $\epsilon^{2}$ (epsilon-squared) statistic \citep{Tomczak2014}, followed by a Bonferroni correction post-hoc pairwise comparison.

\begin{table}[!tb]
    \caption{Significant differences ($p<0.01$) in the responses of the aptitude-related questions ($\epsilon^{2}$: epsilon-squared effect; Scenarios: scenarios with most significant pairwise differences).}%
    \label{Tab:ResultsAptitudeDiff}
    \centering
    \begin{tabular}{l l r l}
    \textbf{Response var.} & \textbf{Agent} & $\pmb{\epsilon^{2}}$ & \textbf{Scenarios} \\
    \midrule
    \multirow{3}[2]{*}{\textit{\textbf{Level}}} 
	& Optimized & 0.32  & \textsf{Min} \\
	& High-vision & 0.39  & \textsf{Min}, \textsf{Uncert} \\
	& Fear-water & 0.07  & \textsf{Uncert} \\
    \midrule
    \multirow{3}[2]{*}{\textit{\textbf{Help}}} 
    	& Optimized & 0.33  & \textsf{Min}, \textsf{Uncert} \\
	& High-vision & 0.40  & \textsf{Min}, \textsf{Freq} \\
	& Fear-water & 0.07  & \textsf{Uncert} \\
    \bottomrule
    \end{tabular}
\end{table}%

Table~\ref{Tab:ResultsAptitudeDiff} shows a significant effect of the summarization technique (scenario) on the \emph{level} and \emph{help} responses for all agent types ($p<0.01$). The results for effect size also show that the videos depicting the behavior of optimized and high-vision agents had a greater impact on the differences induced by the summarization techniques. The Bonferroni comparison shows that the \textsf{Min} and \textsf{Uncert} elements had the greatest impact in inducing different responses from the subjects.

\subsubsection{Research Question 2}

\begin{table}[!tb]
    \caption{Pairing of response variables and agent performance data (ground-truth).}%
    \label{Tab:AptitudeRespPerfPair}
    \renewcommand{\arraystretch}{1.2}
    \centering
    \begin{tabular}{l l}
    \toprule
    \textbf{Response variable} 	& \textbf{Agent performance variable} \\
    \midrule
    \textit{Level}	    			& Mean game level achieved. \\
    \textit{Help}	    			& Mean number of lives lost per game. \\
    \bottomrule
    \end{tabular}
\end{table}

To investigate the correctness of the subjects' assessment of the agents' capabilities and limitations we again paired each response variable with a corresponding agent performance variable, as listed in Table~\ref{Tab:AptitudeRespPerfPair}. We linearly rescaled (min-max scale) the response variables for all scenarios and the corresponding performance variables so that they fell in the $[0,1]$ interval, and then modeled them as Gaussian distributions.%
\footnote{For example, for the \emph{level} performance variable, this models the probability of an agent reaching high levels.}

\begin{table}[!tb]
    \caption{$H$ distance between response and performance variables for the aptitude-related questions, for each scenario and agent (\lowdiv{blue:} low divergence, \highdiv{red:} high divergence).}%
    \label{Tab:ResultsAptitudeCorrect}
    \centering
    \begin{tabular}{l | rrr | rrr} 
    \toprule
    \multirow{2}{*}{\textbf{Scenario}} &
    \multicolumn{3}{c|}{\textbf{\emph{Level}}} &
    \multicolumn{3}{c}{\textbf{\emph{Help}}} \\
    & \textbf{Opt.} & \textbf{High} & \textbf{Fear}
    & \textbf{Opt.} & \textbf{High} & \textbf{Fear}\\    
    \midrule
    \textsf{Max} & $0.27$ & $\highdiv{0.95}$ & $\highdiv{0.99}$ & $0.31$ & $0.57$ & $\highdiv{0.99}$ \\
    \textsf{Min} & $0.56$ & $0.48$ & $\highdiv{0.99}$ & $0.40$ & $0.30$ & $\highdiv{0.99}$ \\
    \textsf{Cert} & $0.45$ & $\highdiv{0.91}$ & $\highdiv{0.99}$ & $0.41$ & $0.50$ & $\highdiv{0.99}$ \\
    \textsf{Uncert} & $0.47$ & $0.52$ & $\highdiv{0.99}$ & $0.37$ & $0.20$ & $\highdiv{0.98}$ \\
    \textsf{Freq} & $0.32$ & $0.53$ & $\highdiv{0.99}$ & $0.26$ & $0.28$ & $\highdiv{0.99}$ \\
    \textsf{Infreq} & $0.39$ & $\highdiv{0.96}$ & $\highdiv{0.99}$ & $0.11$ & $0.63$ & $\highdiv{0.99}$ \\
    \hdashline
    \textsf{Max-Min} & $0.31$ & $0.73$ & $\highdiv{0.99}$ & $0.14$ & $0.27$ & $\highdiv{0.99}$ \\
    \textsf{Cert-Uncert} & $0.35$ & $0.71$ & $\highdiv{0.99}$ & $\lowdiv{0.04}$ & $0.20$ & $\highdiv{0.99}$ \\
    \textsf{Freq-Infreq} & $0.31$ & $0.71$ & $\highdiv{0.99}$ & $\lowdiv{0.04}$ & $0.26$ & $\highdiv{0.99}$ \\
    \hdashline
    \textsf{All} & $0.38$ & $0.75$ & $\highdiv{0.99}$ & $\lowdiv{0.09}$ & $0.32$ & $\highdiv{0.99}$ \\
    \hdashline
    \textsf{Seq} & $0.40$ & $\highdiv{0.91}$ & $\highdiv{0.99}$ & $0.36$ & $0.55$ & $\highdiv{0.99}$ \\
    \bottomrule
    \end{tabular}
\end{table}

%
Then, for each agent, we calculated the Hellinger distance ($H$) \citep{Hellinger1909} between the distributions of the response and performance variables for each scenario to obtain a divergence measure with scores ranging from $0$ (matching distributions), to $1$ (diverging distributions). Table~\ref{Tab:ResultsAptitudeCorrect} highlights the largest and smallest Hellinger distances between the response and game performance variables for each agent and summarization technique.%
\footnote{The thresholds for the $H$ distance were chosen empirically for purposes of analysis, \ie to highlight some results and compare the influence of the different summarization techniques.}

\subsubsection{Assessing Relative Performance}

\begin{figure*}[!tb]
	\centering
        \begin{subfigure}[b]{0.8\textwidth}
            \includegraphics[height=81pt]{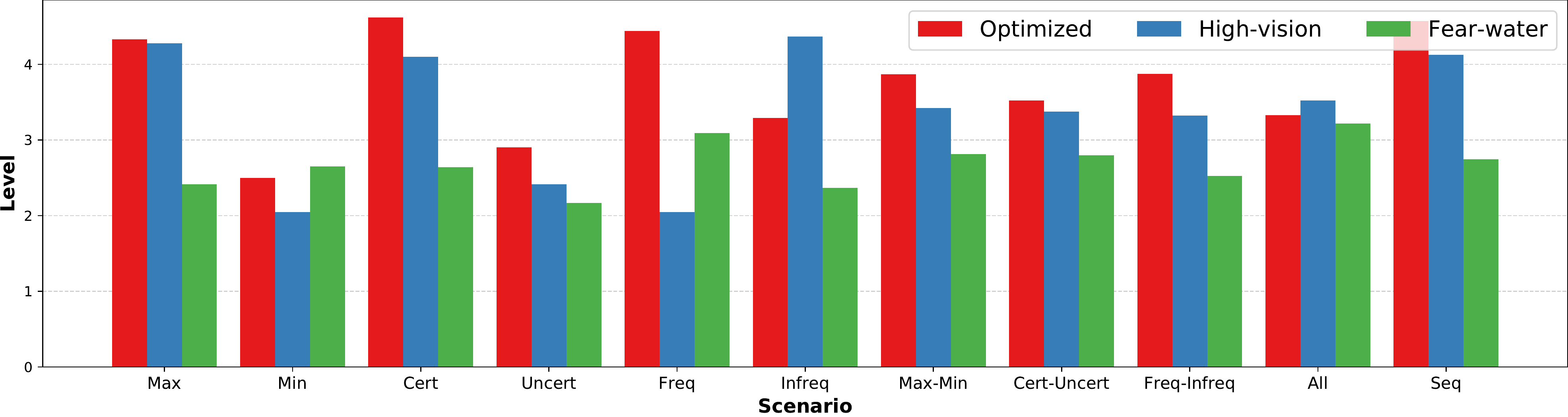}\hspace{5pt}%
            \includegraphics[height=81pt]{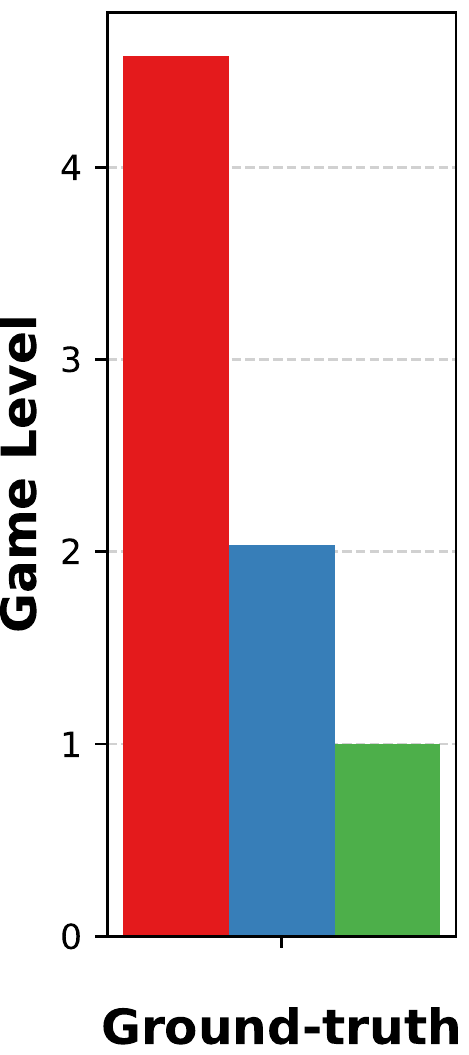}
            \caption{\emph{level} response vs. game level}%
            \label{Fig:LevelCompResults}
            \vspace{10pt}
        \end{subfigure}
        \begin{subfigure}[b]{0.8\textwidth}
            \includegraphics[height=81pt]{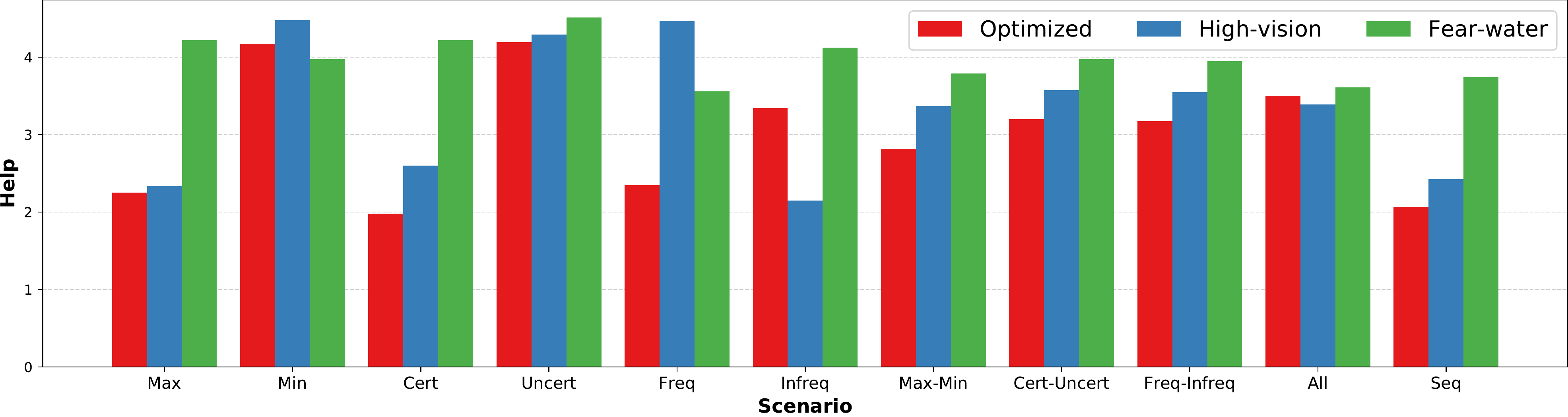}\hspace{5pt}%
            \includegraphics[height=81pt]{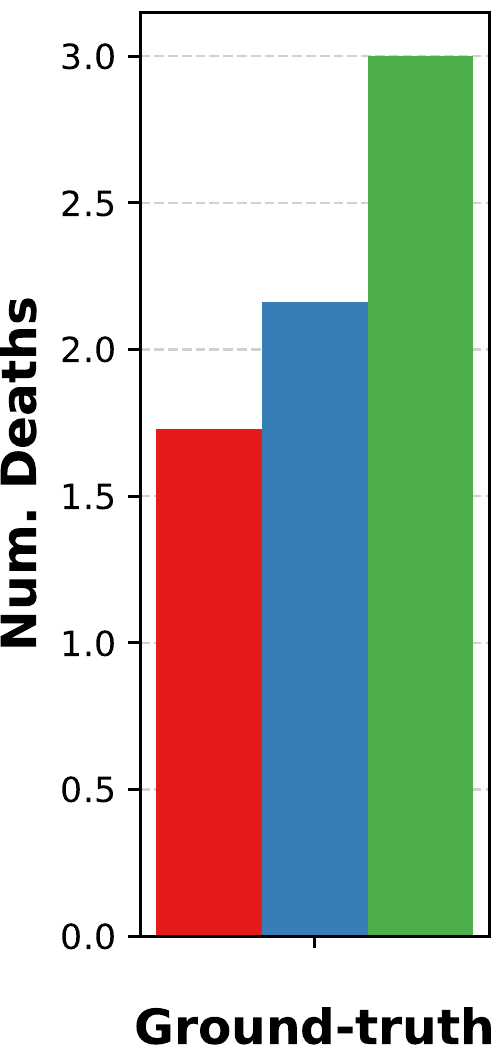}
            \caption{\emph{help} response vs. number of game deaths}%
            \label{Fig:HelpCompResults}
        \end{subfigure}%
    	\caption{Comparison of subjects' mean responses for each agent and summarization technique (Scenario, left) and the agents' game performance across the $2{,}000$ test episodes (Ground-truth, right). We note the different scales between responses and ground-truth data.}%
    	\label{Fig:AptitudeCompResults}
\end{figure*}

The results in the previous section indicate, for some scenarios, a lack of agreement between subjects' responses and the agents' game performance. However, this might be due to our experimental design, where each scenario presented highlights for the three agents simultaneously. Thus, we also analyzed which summarization techniques resulted in a correct perception of the \emph{relative performance} between the agents. 

Figure~\ref{Fig:AptitudeCompResults} illustrates the idea behind our approach. By comparing the shape of the individual scenario plots to the ground truth plot, we see that some techniques appear to support a correct perception of the agents' relative abilities, while others do not. To quantify this difference, we compared the distribution of subjects' responses across the three agents with the distribution of their actual performance. Specifically, for each scenario, we derived Boltzmann probability distributions for the agents from the mean values of the responses and the corresponding performance variables. We then computed the JSD between each response distribution and the ground-truth performance distribution. The results are listed in Table~\ref{Tab:ResultsRelativeAptitudeCorrect}, where the \emph{diverging agents} column shows the agents responsible for the distributions being highly divergent.

\begin{table}[!tb]
    \caption{JSD between the relative response and performance for the aptitude-related questions for each scenario (\lowdiv{blue:} low divergence, \highdiv{red:} high divergence). For high JSD we indicate the agents mostly responsible for the subjects' incorrect perceptions.} %
    \label{Tab:ResultsRelativeAptitudeCorrect}
    \centering
    \begin{tabular}{l | r l | r l}
    \toprule
    \multirow{2}{*}{\textbf{Scenario}} &
    \multicolumn{2}{c|}{\textbf{\emph{Level}}} &
    \multicolumn{2}{c}{\textbf{\emph{Help}}} \\
    & \textbf{JSD} & \textbf{Diverging agents}
    & \textbf{JSD} & \textbf{Diverging agents}\\    
    \midrule
    \textsf{Max} & $0.28$  			&       		& $\lowdiv{0.03}$ 	&  \\
    \textsf{Min} & $\highdiv{0.51}	$ 	& Fear 		& $\highdiv{0.53}$ 	& High, Fear \\
    \textsf{Cert} & $0.09$  			&       		& $\lowdiv{0.02}$ 	&  \\
    \textsf{Uncert} & $\lowdiv{0.04}$  	&       		& $0.21$  			&  \\
    \textsf{Freq} & $\lowdiv{0.00}$ 	&  			& $\highdiv{0.73}$ 	& High, Fear \\
    \textsf{Infreq} & $\highdiv{0.92}$	& Opt., High 	& $\lowdiv{0.03}$ &  \\
    \hdashline
    \textsf{Max-Min} & $0.10$  		&       		& $\lowdiv{0.04}$  	&  \\
    \textsf{Cert-Uncert} & $0.23$ 	&       		& $0.06$ 			&  \\
    \textsf{Freq-Infreq} & $0.06$  	&       		& $0.06$  			&  \\
    \hdashline
    \textsf{All} & $\highdiv{0.52}$ 	& High & $0.24$  	&  \\
    \hdashline
    \textsf{Seq} & $0.12$ 			&       		& $\lowdiv{0.02}$ 	&  \\
    \bottomrule
    \end{tabular}
\end{table}

\subsection{Confidence Analysis}

\begin{figure*}[!tb]
	\centering
	\includegraphics[width=0.4\textwidth]{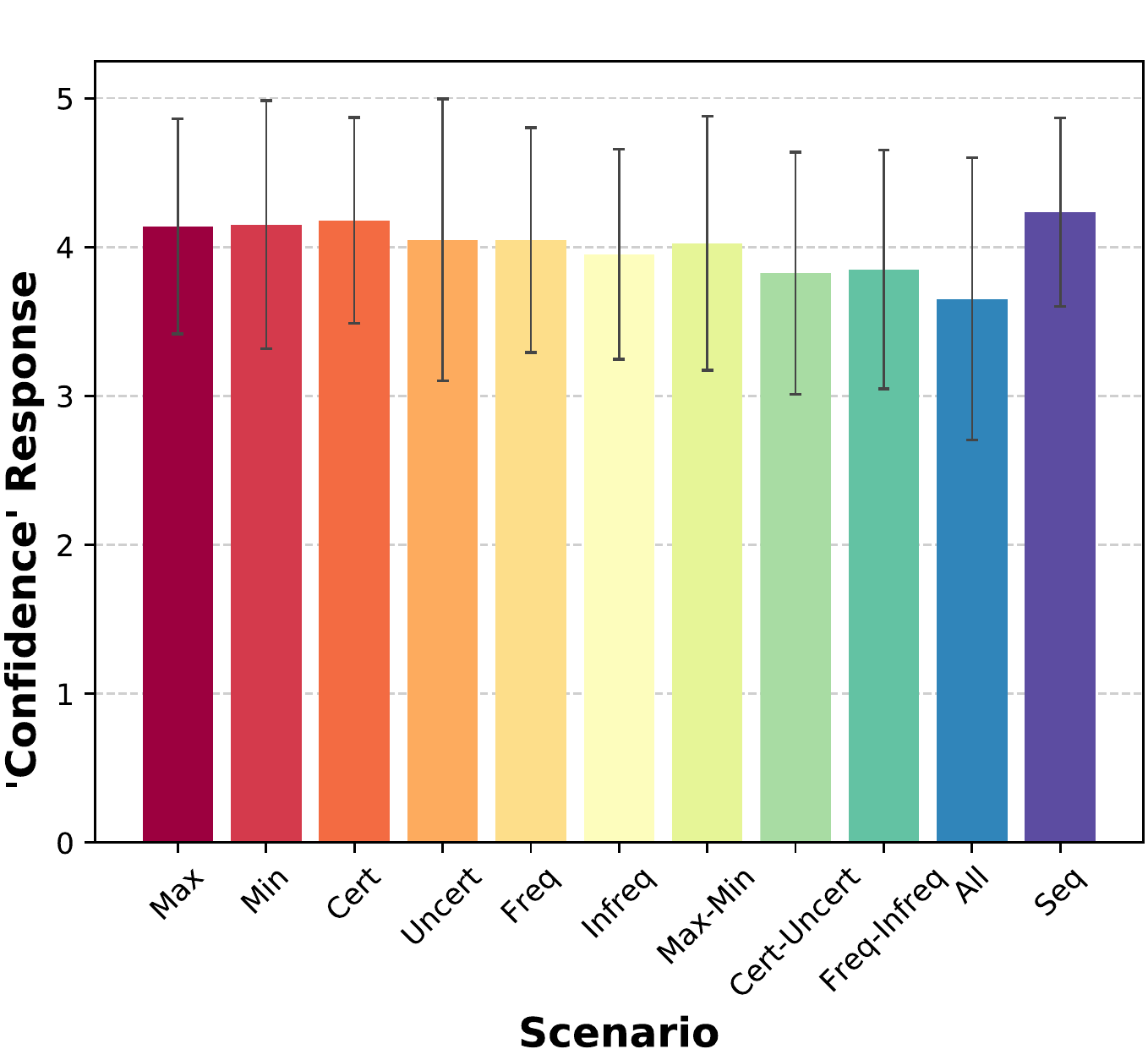}
	\caption{The subjects' mean confidence in their responses in each scenario.}%
	\label{Fig:ConfidenceResults}
\end{figure*}

Finally, we wanted to determine whether some summarization techniques influenced the subjects' confidence response. A Kruskal-Wallis analysis of variance did not reveal a significant effect of the summarization technique (scenario) on the subjects' \emph{confidence} response ($p=0.047$). Fig.~\ref{Fig:ConfidenceResults} shows the mean responses per scenario. We see that, in general, the subjects' confidence in their answers was not greatly affected by exposure to particular scenarios. A Bonferroni correction post-hoc comparison revealed that the largest difference was found between \textsf{Seq} (most confident) and \textsf{All} (least confident) ($p=7.5\times 10^{-4}, \eps^{2}=0.34$).

\subsection{Discussion}

In this section we discuss in depth the results of our experimental study and provide the main insights stemming from using each summarization technique.

\subsubsection{Importance of ``Negative'' Moments}

As detailed in Section~\ref{Sec:RelatedWork}, most previous works within XRL identify key moments of interaction resulting in examples of behavior where the agent excels in the task---this has been shown to be useful \eg to assess relative performance when comparing between different agents. In contrast, our dimensions of analyses result in the presentation of situations where the agent is also less confident, thereby providing opportunities for intervention, \eg for the agent designer to improve the agent's perceptions, improve learning, etc. 

The results of our user study show the importance of showing agent behavior in such situations. Namely, Tables~\ref{Tab:ResultsRegionDiff} and \ref{Tab:ResultsAptitudeDiff} show that the most significant pairwise differences were produced in the \textsf{Min} and \textsf{Uncert} scenarios, denoting the importance of exposing ``hard'' situations for assessing the agents' relative presence in different regions and their overall aptitude in the task. Further, Table~\ref{Tab:ResultsAptitudeCorrect} shows that for the high-vision agent, who could succeed in the task despite perceptual limitations, exposing subjects to these situations allowed for a more accurate assessment of the agent's capabilities in overcoming such flaws.

\subsubsection{Importance of Diversity in Behavior Summaries}

Despite the importance of showing situations where the agents are less confident, alone they do not allow for a correct understanding of the agents' aptitude in the task. In fact, a main conclusion of the study is that no single technique can induce a correct understanding of the underlying capabilities and limitations of \emph{all} agents in \emph{all} situations. In particular, exposing only ``good'' or ``bad'' moments can lead users to an incorrect perception of aptitude. 

For example, Tables~\ref{Tab:ResultsAptitudeCorrect} and \ref{Tab:ResultsRelativeAptitudeCorrect} show that providing highlights of only favorable situations (\textsf{Max} and \textsf{Cert}) does not lead to a completely correct perception of the agents' capabilities. Meanwhile, showing only minima and infrequent moments can lead users to incorrectly perceive relative agent limitations and to believe that all agents fail equally. Two cases worth noting here are the optimized agent, whose performance appears to have been underestimated by subjects in the \textsf{Min} and \textsf{Uncert} scenarios, and the fear-water agent, whose limitations were perceived to be very similar to that of the other two agents, which we know to be incorrect given these agents' true game performance.

Primarily, the problem is that by providing examples only of where an agent excels or where it fails, users may not be exposed to the parts of the task that required more of the agent's attention (time spent), nor where and how the agent needs to be improved. In contrast, our results show that techniques providing examples captured at the two extrema of the dimensions of analysis can provide a more accurate view of the agents' competence in the task. For example, Table~\ref{Tab:ResultsRegionCorrect} shows that the aptitude of the high-vision agent was better depicted by the moments captured in the \textsf{Max-Min} and \textsf{Freq-Infreq} scenarios while Table~\ref{Tab:ResultsAptitudeCorrect} showed similar results for the optimized agent in the \textsf{Freq-Infreq} and \textsf{Cert-Uncert} scenarios. Overall, these techniques provide a balance of ``good'' and ``bad'' moments of performance, which can allow for a better understanding of agents' capabilities and limitations in the task.

\subsubsection{Importance of the Frequency Dimension}

One dimension of analysis providing good overall results was frequency. In particular, the results of the region-related responses in Table~\ref{Tab:ResultsRegionDiff} show that the frequency dimension, by providing a balance between common and rare situations, provides a good indication of the improvements needed in each subtask. Further, Table~\ref{Tab:ResultsRegionCorrect} shows a low divergence when comparing subjects' responses with the agents' true performance in the \textsf{Freq-Infreq} scenario, thus emphasizing the importance of the frequency dimension for correctly assessing where an agent spends most of its time and where it needs more (or less) practice. 

On one hand, the results show the importance of frequent elements for correctly determining where an agent spends most of its time. On the other, infrequent situations are also relevant in that they can help identify when the agent fails. Moreover, frequency seems to be useful for showing an agent's overall behavior characteristics, while the transition-value dimension and other more complex elements relying on the learned value function help reveal its difficulties and capabilities in hard, challenging situations. 

\subsubsection{Importance of the Sequence Dimension}

Another technique worth mentioning is that of most likely sequences to maxima---\ie capturing agent trajectories that go from a difficult situation (local minimum) to a favorable situation (local maximum). An advantage of this technique is that an observer gets to see the ``normal'' behavior of the agent in the task, which is related to its learned strategy. The results in Table~\ref{Tab:ResultsRelativeAptitudeCorrect} show that even with a single trajectory, subjects could correctly perceive the unique characteristics of the agents' behavior and infer their underlying relative aptitude. Although the \textsf{Seq} scenario did not get the lowest divergence scores overall, Fig.~\ref{Fig:ConfidenceResults} shows that subjects were more confident in their responses. Hence, sequences may enable a reasonably accurate assessment of how the agent performs in the task, particularly with a limited explanation budget (time). 

\subsubsection{Richness of an Agent's Behavior}

A related insight is that the summarization techniques, and the associated interestingness elements, are only as good at capturing different characteristics of RL agents as the richness of the agents' underlying capabilities. For example, from Tables~\ref{Tab:ResultsRegionCorrect} and \ref{Tab:ResultsAptitudeCorrect} we see that the behavior exhibited by the fear-water agent due to its ``motivational impairments'' is so limited that no single interestingness element captured qualitatively distinct behaviors. Notably, the summaries gave subjects the impression that the agent needs practice in all the regions equally when in fact it has the most difficulty in jumping onto logs when reaching the river. In contrast, the agents exhibiting more diverse behavior enabled subjects to differentiate the agents' capabilities under different conditions, as evidenced by their responses (see \eg Table~\ref{Tab:ResultsAptitudeDiff}, where a significant effect of the summarization technique was found).

In general, agents that have predictable, monotonous behavior may not generate the necessary numerical nuances targeted by our introspection framework. For example, if the values associated with the states are all high, one cannot distinguish between minima and maxima. Likewise, if the observed behavior is highly deterministic, then uncertain elements cannot be determined. 

\subsubsection{Problems with Combining All Elements}

Our results also captured an interesting result regarding the \textsf{All} scenario, where summaries contain one highlight for each interestingness element. One might think this technique would be optimal, given that it highlights all the different aspects of the agent's interaction. However, in our study it led to the lowest confidence in responses (Fig.~\ref{Fig:ConfidenceResults}) and poor perceptions overall, \eg  Tables~\ref{Tab:ResultsRegionCorrect}, \ref{Tab:ResultsAptitudeCorrect} and \ref{Tab:ResultsRelativeAptitudeCorrect} show that this technique resulted in overall high divergence between subjects' responses and true agent performance. We believe that the diversity of moments highlighted by each element may have confused the subjects, making it difficult for them to determine an agent's overall aptitude. Also, half the highlights show infrequent situations or where agents are uncertain of what to do, leading subjects to underestimate their performance.

\section{Future Work and Conclusions}%
\label{Sec:Conclusions}

\subsection{Future Work}

A problem with visual summaries based on highlights is that, by definition, they do not capture all aspects of an agent's interaction with the environment. Because of that, users may extrapolate the behavior of an agent beyond what they see, potentially leading them to incorrectly perceive its overall performance. Further, this can be exacerbated when user expectations are not met. For example, in our study subjects may have formed beliefs about how ``normal'' players behave in the Frogger game prior to seeing the videos, which may have led to incorrect perceptions of aptitude in some scenarios. Also, by design, our experiment involved visualizing agents simultaneously, which could have led to the creation of false expectations based on the best and worst performance observed among all agents.

One possible solution would be to augment highlights with more quantitative measures about the agents' overall performance in the task. This may help users form a better assessment of the agents' capabilities and limitations, and of when and how they should intervene. For example, observing a high-performing agent failing to perform a particular subtask can direct a user to correct specific aspects of the agent's design. Such information could also help users understand that not all agent limitations are the same and that different agents may require distinct adjustments.

Another solution to mitigate problems with unmet expectations would be to provide users with short descriptions about what the highlights are conveying about an agent's behavior. This would be particularly useful in our framework given the variety of interestingness elements and the qualitatively distinct aspects of an agent's experience that each captures. This way, users could better understand agents' limitations and set more realistic expectations about their behavior.

Our introspection framework was designed to be used throughout the different stages of an agent's lifecycle and for different explanation modes. For example, it can be used \emph{during learning}, to track the agent's learning progress and acquired preferences or \emph{after learning}, to summarize the most relevant aspects of the interaction. It can be used \emph{passively}, to let a user ask the agent about its current goals or to justify its behavior in a given situation or \emph{proactively}, to let the agent consult the user in situations where its decision-making is more uncertain or unpredictable. We are currently developing a platform where the competency of RL agents, including competency in learning, can be assessed for different tasks and in novel situations.

We are also currently extending the introspection framework to support analyzing high-dimensional domains that require the use of deep-RL techniques by defining interpretations of the concepts behind the different interestingness elements that can leverage deep network representations. Some elements allow for a direct translation to deep-RL architectures, \eg estimating execution certainty from the output distribution of a policy network or transition value from situations where executing different actions leads to strictly increasing or decreasing value. Others may require more specialized structures, \eg the transition certainty of a situation can be estimated through generative models \citep[\eg][]{Ha2018,Hafner2019} that predict the likelihood of next states, while a model for characterizing the familiarity of inputs \citep[\eg][]{Akcay2019} could be used to determine the frequency of observations.

\subsection{Summary and Main Conclusions}

In this paper we introduced a novel framework for introspective XRL that extracts different \emph{interestingness elements}, each capturing a particular aspect of the agent's experience with an environment. We showed how to generate visual summaries from these elements in the form of video-clips highlighting diverse moments of the agent's experience while using its learned policy. To assess the usefulness of the elements in helping humans understand RL agents' competency, we performed a user study using the game of Frogger. First, we created agents with different perceptual capabilities and motivations in the task to simulate potential problems in the agents' design. We trained the agents using RL, resulting in different game behaviors and levels of performance. We then performed an online survey where we presented subjects with videos of the agents' behavior generated using different summarization techniques, each using a distinct combination of the interestingness elements. 

Our results show that no single summarization technique---and hence no single interestingness element---provides a complete understanding of every agent in all possible situations of a task. Ultimately, a combination of elements enables the most accurate understanding of an agent's aptitude in a task. Different combinations may be required for agents with distinct capabilities and levels of performance. 
%
The results highlighted in particular the element of frequency, which provides a good understanding of an agent's overall behavior characteristics. As for the transition-value dimension, it can denote difficulties and capabilities of agents in hard, challenging situations. In addition, the element producing the most likely sequences to desirable situations can provide a short but accurate understanding of an agent's learned strategy. Another insight stemming from the results is that combining many types of elements in a single video summary can confound users. 

A few general notes about the proposed framework are worth mentioning. Our framework for explainable reinforcement learning extracts a diverse set of interestingness elements capturing significant moments of agent behavior. The introspection framework, comprising the dimensions of analysis that extract the interestingness elements, is \emph{domain-independent}: the data that is collected and analyzed is agnostic to the specific learning scenario. In addition, it is \emph{algorithm-independent} in that it can be used in conjunction with standard RL methods without having to modify the learning mechanism itself. 

We note also that the dimensions of analysis proposed here provide mathematical interpretations of different \emph{concepts} for analyzing an RL agent's interaction with the environment. Namely, each concept provides different insights over the agent's behavior that are useful for judging its performance and to understand how it can be improved. We do not claim that the list presented in this paper is complete or definitive---other aspects of an RL agent's interaction can be defined, and different interpretations, possibly applicable to different RL algorithms, can be proposed for the same concepts. Similarly, if one RL algorithm does not learn the interaction data required by some element, then only a subset of the interestingness dimensions will be available for analysis.

Moreover, the analyses in our framework do not make direct assertions about potential problems with the agent's training or with the resulting policy. Neither do they prescribe any specific intervention by a human user to improve the performance of the agent in the task. Rather, they help indicate important aspects of the agent's behavior. The purpose of the user study was precisely to help make the connection between the different elements and the perception of the agent's characteristics by human observers.

Regarding state-space dimensionality, we chose to develop our initial framework for simpler tasks because, given the inherent complexity of sequential decision-making, it is important to first determine how to generate the critical \emph{elements} for explainable RL. Thus, the concepts behind our elements---and their effects in enabling users to understand the aptitude of agents---are more important than the particular evaluation domain or the RL architecture used to train the agents. Furthermore, we note that the subjects of our study were unaware that the players were in fact autonomous agents, let alone that RL was used to derive their behavior. As such, our evaluation of the usefulness of the explanation concepts is agnostic to the underlying representations and RL algorithm used.

The results of this paper show that the \emph{diversity} of aspects captured by the proposed interestingness elements can help humans correctly identify an RL agent's aptitude in a task. Further, not only do they help reveal the underlying characteristics of agents, but as a consequence, we believe they can be good indicators of which adjustments might be required to improve their performance.

\footnotesize
\bibliographystyle{abbrvnat}
\bibliography{19-aijxai-xrl}

\clearpage


\normalsize
\appendix
\section{Additional Introspection Analyses and Interestingness Elements}%
\label{Sec:AppendixOtherElements}

This appendix details the additional interestingness elements captured by our framework that were \emph{not} included and evaluated in the user study. The complete framework operates at three levels, with the first analyzing characteristics of the task that the agent has to solve, the second behavior of the agent while interacting with the environment, and the third performs a meta-analysis combining information gathered at the lower levels. Given the number of elements generated by our framework, in the study presented in the main document subjects were exposed to only a subset of the elements. In particular, we selected elements from dimensions analyzing an agent's interaction with the environment, thereby excluding level 0 of analyses. Moreover, we focused on elements that help characterize the agents' overall behavior.

The complete framework relies on the agent collecting the following additional \emph{interaction data}:
\begin{itemize}
	\item $\ts(z)$ is the last (most recent) time-step of the agent's history of interaction in which $z$ was observed. Similarly, $\tsa(z,a)$ denotes the last time-step in which it executed action $a$ after observing $z$;
	\item $\hat{\Rwd}(z,a)$ is the agent's \emph{estimate of the reward} received for performing action $a$ after observing $z$. It can be estimated by maintaining a running average of the rewards received, \ie $\hat{\Rwd}(z,a)=\sum_{t\in T_{z,a}} r / \csa(z,a)$, where $T_{z,a}$ contains the time-steps where action $a$ was executed after observing $z$;
	\item $\widehat{\Delta Q}(z,a)$: the expected \emph{prediction (Bellman) error} associated with $Q(z,a)$. Given an observed transition $(z,a,r,z')$, the prediction error corresponds to $\Delta Q(z,a)=r+\gamma\max_{b\in\A}Q(z',b)-Q(z,a)$. As with $\hat{\Rwd}(z,a)$, the agent can maintain a running average of the prediction errors by using $\widehat{\Delta Q}(z,a)=\sum_{t_{z,a}} \Delta Q(z,a) / \csa(z,a)$.
\end{itemize}

In addition, as is the case with many RL scenarios, we assume that, at each time-step $t$, the agent observes its environment through a finite set of \emph{features} $Z_{t}^{i}=z^{i},i=1,\ldots,\numfeats$, each taking values in some feature space $\Z^{i}$. The observation space thus corresponds to the cartesian product $\Z=\Z_1\times\ldots\times\Z_\numfeats$. When this is the case, the structure exhibited by such \emph{factored MDPs} can also be exploited to derive interesting aspects related to specific observation elements.

Table~\ref{Table:Elements} lists all the analyses implemented so far, how they are grouped at different levels, and a short description of the elements that each generates.

\begin{sidewaystable}[htbp]
	\centering
	\small
	\renewcommand{\arraystretch}{1.5}
	\setlength\tabcolsep{0pt}
	\caption{Overview of the dimensions of analysis at each level of the introspection framework and the generated interestingness elements. Dimensions and elements included for evaluation in the user study are marked in bold, and were detaile din the main document.}%
	\label{Table:Elements}
	\begin{tabular}{ L{75pt}@{\hspace{6pt}} L{80pt}@{\hspace{6pt}} L{65pt}@{\hspace{6pt}} L{310pt} }
	\toprule
	\textbf{Level} &
	\textbf{Purpose} & 
	\textbf{Dimension} & 
	\textbf{Generated Interestingness Elements} \\ 
	\hline
	\multirow{2}{*}{0. Task} & 
	\multirow{2}{80pt}{\setstretch{1.1}\textit{analyze task characteristics}} &
	Transition Certainty & 
	(un)certain transitions: \emph{hard/easy to predict with regards to next-step transitions}\\ 
	&  & 
	Reward & 
	reward outliers: \emph{where the agent receives a relatively high/low reward}\\ 
	\hdashline
	\multirow{6}{*}{1. Interaction} & 
	\multirow{6}{80pt}{\setstretch{1.1}\textit{analyze history of interaction with environment}} & 
	\textbf{Frequency} & 
	observation coverage and dispersion: \emph{characterize the agent's exploration}\newline 
	\textbf{(in)frequent situations:} \emph{situations the agent finds very (un)common}\newline
	strongly-/weakly-associated features: \emph{patterns in the observation features}\\
	&  & 
	\textbf{Execution Certainty} &
	\textbf{(un)certain executions:} \emph{hard/easy to predict with regards to action execution}\\
	&  & 
	Recency & 
	ancient situations: \emph{experienced by the agent a long time ago}\\
    	&  & 
	Value & 
	value outliers: \emph{where the agent expects a relatively high/low value}\newline
	mean prediction error: \emph{how hard is the task for the agent}\newline
	prediction outliers: \emph{situations that are ``hard to learn''}\\
	\hdashline
	\multirow{3}{*}{2. Meta-analysis} & 
	\multirow{3}{80pt}{\setstretch{1.1}\textit{combine elements from different analyses}} & 
	\textbf{Transition-Value} & 
	\textbf{observation minima and maxima:} \emph{learned goals and situations to avoid}\\
	&  & 
	\textbf{Sequence} & 
	\textbf{most likely sequences to maxima:} \emph{summarizes the agent's learned strategies}\\
	&  & 
	Contradiction & 
	contradictory values and goals: \emph{where action selection contradicts internal/external expectations}\\
	\bottomrule
	\end{tabular}
\end{sidewaystable}


\subsection{Level 0: Analyzing the Environment}

This level of introspection analyzes characteristics of the task that the agent has to solve. The focus is therefore on the estimated reward and transition probability functions, \ie $\hat{\Rwd}$ and $\hat{\P}$, respectively.

\subsubsection{Transition Certainty Analysis}

The estimated transition probability function $\hat{\P}(z'\mid z,a)$ can be used to expose relevant aspects of the environment dynamics, in the perspective of the agent. Namely, it can be used to identify certain and uncertain transitions. Given an observation $z$ and an action $a$, the transition certainty associated with $(z,a)$ is measured according to how concentrated the observations $z'\in\Z$ following $(z,a)$ are. For that purpose, we use the \emph{evenness} measure defined in Eq.~1 to calculate the dispersion of the distribution over possible next observations according to the data stored in $\hat{\P}$. Namely, for a given observation-action pair $(z,a)$, $\evenness_{za}=\evenness(\hat{\P}(\cdot|z,a))$ provides the evenness associated with transitioning from $(z,a)$. The following interestingness elements are generated by this dimension of analysis:

\textbf{(Un)Certain transitions}: denote situations in which the next state is hard/easy to be predicted by the agent. For observations, we use $\evenness_{za}$ to identify the certainty associated with a given $(z,a)$ pair. For features, the certainty of a feature $z^{i}$ is calculated by averaging the evenness $\evenness_{za}$ for all observations $z\in\Z$ in which $z^{i}$ is active.

The rationale behind the transition dimension is that transitions leading to many different states have a high dispersion and are considered \emph{uncertain}. Likewise, transitions leading to a few states have a low dispersion can be considered \emph{certain}. This analysis thus highlights the (un)certain elements of the agent's transitions, actions and of its observation features. Uncertain elements are especially important as people tend to resort to ``abnormal'' situations for the explanation of behavior \citep{Miller2019}. This information can also be used by the agent in a more proactive manner while interacting with the environment. For example, the agent can express its confidence in the result of its actions when necessary, or request the help of a human user when it faces a very uncertain situation. 

\subsubsection{Reward Analysis}

The idea behind this analysis is to identify uncommon situations regarding the reward received by the agent during its interaction with the environment. As $\hat{\Rwd}$ corresponds to a model of the rewards received by the agent, parts of the true reward function may have not been captured as the model will reflect the agent’s behavior in the environment. Notwithstanding, the purpose of our framework is to analyze a \emph{particular history} of interaction rather than the fidelity of the models or the optimality of the learned behavior. The following interestingness elements are generated:

\textbf{Reward outliers:} correspond to observation-action pairs in which, on average among all other states and actions, the agent received significantly more or less reward. Reward outliers correspond to: 

\begin{equation}\label{Eq:Outliers}
	\forall_{(z,a)\in\Z\times\A} \colon \abs{\hat{\Rwd}(z,a)-\rwdavg} > \lambda_{\sigma} \sigma_{\rwdavg},
\end{equation}
where $\rwdavg$ is the overall average reward collected by the agent, $\sigma_{\rwdavg}$ is the standard deviation of $\rwdavg$ and $\lambda_{\sigma}$ is an arbitrary threshold to determine outliers. When considering observations as a whole, this information may be used to identify situations in which the agent is likely to receive relatively low or high rewards. When considering individual features, it can help capture significant individual contributions of features to the agent's reward that are otherwise ``diluted'' amongst all features if considering only observations.

\subsection{Level 1: Analyzing the Interaction with the Environment}

The purpose of this level is to help characterize the environment's dynamics, as captured by an agent, and extract important aspects of the agent's behavior and history of interaction with it.

\subsubsection{Frequency Analysis}

This analysis identifies interestingness elements that can be found given information stored in the frequency tables $\cs(z)$, $\cs(z,a)$ and $\cs(z,a,z')$. In addition to the (in)frequent situations, detailed in the main document, the following elements are extracted:

\textbf{Observation coverage:} corresponds to how much of the observation-space---regarding all possible combinations between the observation features---were actually observed by the agent. Formally, it corresponds to: $\sum_{z\in\Z}{\cs(z)}/\card{\Z}$. This is an overall performance element that can provide an indication of how much of the state-space was covered by the agent’s behavior, which is an important quality of its exploration strategy.

\textbf{Observation dispersion:} corresponds to how even the distribution of visits to the observation space was. In particular, it analyzes the histogram of observations using the evenness metric in Eq.~\ref{Eq:Evenness}. It can be used to infer how unbalanced the visits to states were, which in turn may denote how interesting the dynamics of the environment are, \eg denoting situations that are physically impossible to occur, or how exploratory the agent was during the interaction with it. 

\textbf{Strongly-/weakly-associated features:} are sets of observation features (feature-sets) that frequently/rarely co-occur. To determine such elements, we resort to frequent pattern-mining (FPM) \citep{Agrawal1994}, a data-mining technique to find patterns of items in a set of transactions. We first transform each observation $z\in\Z$ into a transaction corresponding to the set of features that are active in that observation, \ie $(z^{1}, z^{2}, \ldots, z^{\numfeats})$. Each observation-transaction is then repeatedly added to a data-base for a number of times according to its frequency, as given by $\cs(z)$. After that, we create a frequent-pattern tree (FP-tree) \citep{Han2004} that facilitates the systematic discovery of frequent combinations between the items in the data-base. We use the \emph{Jaccard index} \citep{Jaccard1912}, a metric that can be used to measure the association strength of item-sets \citep{Sequeira2010,Sequeira2013}, and the FP-Growth algorithm \citep{Han2004} to retrieve all observation feature-sets that have a Jaccard index above or below a given threshold.%
\footnote{Further technical details can be found in \citep{Sequeira2019}.}
This element may be used to denote both patterns in the agent’s perceptions or regularities in its environment, and also rare on inexistent combinations of features. In turn, these aspects may be important to explain the agent's physical interaction with the environment and expose its perceptual limitations to an external observer. 

\textbf{Feature-rules:} these are rules in the form $\z_{a} \Rightarrow \z_{c}$, where $\z_{a}$ is the rules antecedent and $\z_{c}$ the consequent. The idea is to determine sets of features--the antecedent---that frequently appear conditioned on the appearance of other sets of features--the consequent. We use the interest or \emph{lift} statistical measure \citep{Agrawal1994} to determine the \emph{confidence} of every rule given the strongly-associated feature-sets. This element can be used to determine causal explanations in the environment as observed by the agent.

\subsubsection{Recency Analysis}

This analysis uses the data stored in the recency tables $\ts(z)$ and $\tsa(z,a)$ to produce the following element:

\textbf{Ancient situations:} these correspond to situations that were encountered by the agent in the beginning of its interaction with the environment but that have not been visited recently. This is achieved by filtering observations and observation-action pairs whose last time-step, according to the information stored in $\ts(z)$ and $\tsa(z,a)$, respectively, is below a given threshold. These elements can be useful to identify rare situations encountered by the agent, or situations that the agent tends to avoid according to its action-selection policy.

\subsubsection{Value Analysis}

The value functions $Q(z,a)$ and $V(z)$ learned by the agent are a very important source of interestingness. Specifically, they can be used to identify situations that are valuable for the agent, thereby denoting its goals and sub-goals when interacting with the environment. In addition, the mean prediction errors stored in $\widehat{\Delta Q}(z,a)$ can be used to identify situations in which the agent experienced learning difficulties. This analysis uses these sources of information to generate the following interestingness elements:

\textbf{Value outliers:} correspond to the observation- and feature-action pairs that are significantly more or less valued than other observations. We use the same outlier-detection method used outlined in Eq.~\ref{Eq:Outliers} but using the action-value function $Q(z,a)$.
 This element denotes desirable situations with regards to the agent’s goals---high-value pairs denote situations conducive for the agent to attain its goals while low-valued situations might prevent the agent of fulfilling its task. If the agent's observations are composed of features, then this element can be further used to denote significant individual contributions of feature values to achieving the agent’s goals, \eg to explain which objects in the environment are more valuable to the agent or to identify which elements the agent tries to avoid.

\textbf{Mean prediction error:} this is the mean prediction error amongst all states and actions, corresponding to: $\sum_{z\in\Z}{\sum_{a\in\A}{\widehat{\Delta Q}(z,a)}}/\card{\Z}\card{\A}$. This element is another overall performance metric, and can be used to evaluate the accuracy of the agent’s world model, \ie how well can the agent predict the consequences and future value of its actions in most of the situations it encounters. In addition, by tracking this element while the agent is learning we may verify its learning progress---if the average prediction error is decreasing over time, it means that the agent is learning the consequences of its actions. Similarly, if the value is not decreasing or is actually increasing this may mean that the agent is not learning the policy properly or simply that the environment is very dynamic and unpredictable---even if only in the agent’s perspective, \ie caused by its perceptual limitations.

\textbf{Prediction outliers:} correspond to the observation-action pairs that have a significantly higher or lower mean prediction error associated to them. This element may denote situations that are very hard (or easy) for the agent to learn. For example, in situations where the next observation and reward received are very stochastic, the agent's future might be very unpredictable and uncertain. Exposing an end-user to such situations may help her decide that the agent is not ready for deployment and needs more training in some regions of the state space. It is also a good opportunity for the user to provide corrective feedback.

\subsection{Level 2: Meta-Analysis}

This level refers to analyses combining information from the different interaction data and the analyses at previous levels, resulting in the identification of more complex aspects of the interaction.

\subsubsection{Transition-Value Analysis}

This analysis combines information from the agent's estimated $V(z)$ function and the transition function $\hat{\P}(z'\mid z,a)$. The goal is to analyze how the value attributed to some observation changes with regards to possible observations taken at the next time-step. In addition to the maxima and minima elements, detailed in the main document, the following elements are produced:

\textbf{Variance outliers:} Let $\bar{v}_{za}=\sum_{z'\in\trans_{za}}{\abs{V(z)-V(z')}}/\card{\trans_{za}}$ be the mean absolute difference of values to the immediate observations $z'$ taken after executing $a$ when in $z$, where $\trans_{za} \doteq \set{\forall_{z'\in\Z} \colon \hat{\P}(z'\mid z,a)>0}$ is the set of observed transitions starting from observation $z$ and executing action $a$. Then, let $\sigma_{\bar{v}_{za}}^{2}$ denote the variance associate with the mean $\bar{v}_{za}$ for observation $z$ and action $a$. For each observation $z\in\Z$, we then calculate $\sum_{a\in\A}{\sigma_{\bar{v}_{za}}^{2} \csa(z,a)/\cs(z)}$, \ie the mean difference variance among all actions, where each action is weighted according to the relative number of times it was executed. Finally, we take the standard deviation of each mean to select the observation variance outliers, \ie observations where the variance of the difference in value to possible next observations is significantly higher or lower than in other observations. This interestingness element is important in that it can be used to identify highly-unpredictable and especially risky situations, \ie in which executing actions might lead to either lower- or higher-valued next states.

\subsubsection{Contradiction Analysis}

This analysis combines information from the value, reward and frequency functions, the value analysis and provided domain knowledge. The goal is to identify \emph{unexpected} situations, where the agent was expected to behave in a certain manner, but the collected data informs us otherwise. Hence, we can automatically determine the foils for behavior in specific situations. 

\textbf{Contradictory-values:} correspond to observations in which the actions' value distribution proportionally diverges from that of their rewards. Specifically, for each observation $z\in\Z$ we first derive probability distributions over actions in $\A$ by normalizing the values and rewards associated with $z$, according to the data stored in $Q(z,\cdot)$ and $\hat{\Rwd}(z,\cdot)$, respectively. We then compute the Jensen-Shannon divergence (JSD) \citep{Fuglede2004} that measures how (dis)similar two discrete probability distributions are with regards to the relative proportion that is attributed to each element. If the value attributed to actions is proportionally very different from the reward that the agent expects to receive by executing the same actions, the JSD will be close to $1$. On contrary, low JSD values (close to $0$) denote a low divergence and hence similar, or aligned, distributions. 

This distinction is relevant because in typical RL scenarios, the reward is something \emph{externally} defined by the agent's designer according to the task, while the value is something \emph{internally} learned by the agent. Therefore, by using this technique and resorting to the information stored in $Q$ and $\hat{\Rwd}$, we can identify situations with a \emph{value-reward} JSD higher than a given threshold. In such situations, the agent may select actions contradicting what an external observer would expect, \eg choosing a low-reward but high-value action. Further, we analyze the individual components of the JSD to identify which indexes are responsible for the non-alignment or dissimilarity between the distributions. This allows us to automatically detect the contradictory situations and provide justifications to users regarding the agent's unexpected behavior, \eg by showing that the selected action leads to a certain subgoal and is thus a better option compared to the expected action.

\textbf{Contradictory-goals:} to identify these elements, we assume that the system is provided with domain-knowledge regarding goal states. This corresponds to the agent designer's expectations over its learned behavior, \ie situations that would normally be considered as highly-desirable for the agent to perform the task by an external observer. Based on this information we determine which observations that were found by our framework to be sub-goals for the agent---identified as local maxima by the transition-value analysis---are not in the known list of goals. The goal is to avert user surprise, which can lead to the erosion of trust in the system \citep{Gervasio2018}. Namely, from the contradiction list an end-user can resort to the explanation techniques to gain insights over the agent's performance, \eg visualize its behavior in contradictory situations, query about the existence of likely paths from those situations to known goals, etc.

\section{Agent Experiments}%
\label{Sec:AppendixAgentExperiments}

This appendix provides additional details regarding the agent experiments performed in Frogger. In particular, the results of the performance in the task and of the interestingness elements are detailed for each of the three agents.



\subsection{Agents Performance Results}

\begin{figure}[!tb]
	\centering
        \begin{subfigure}[b]{0.4\columnwidth}
		\includegraphics[width=\textwidth]{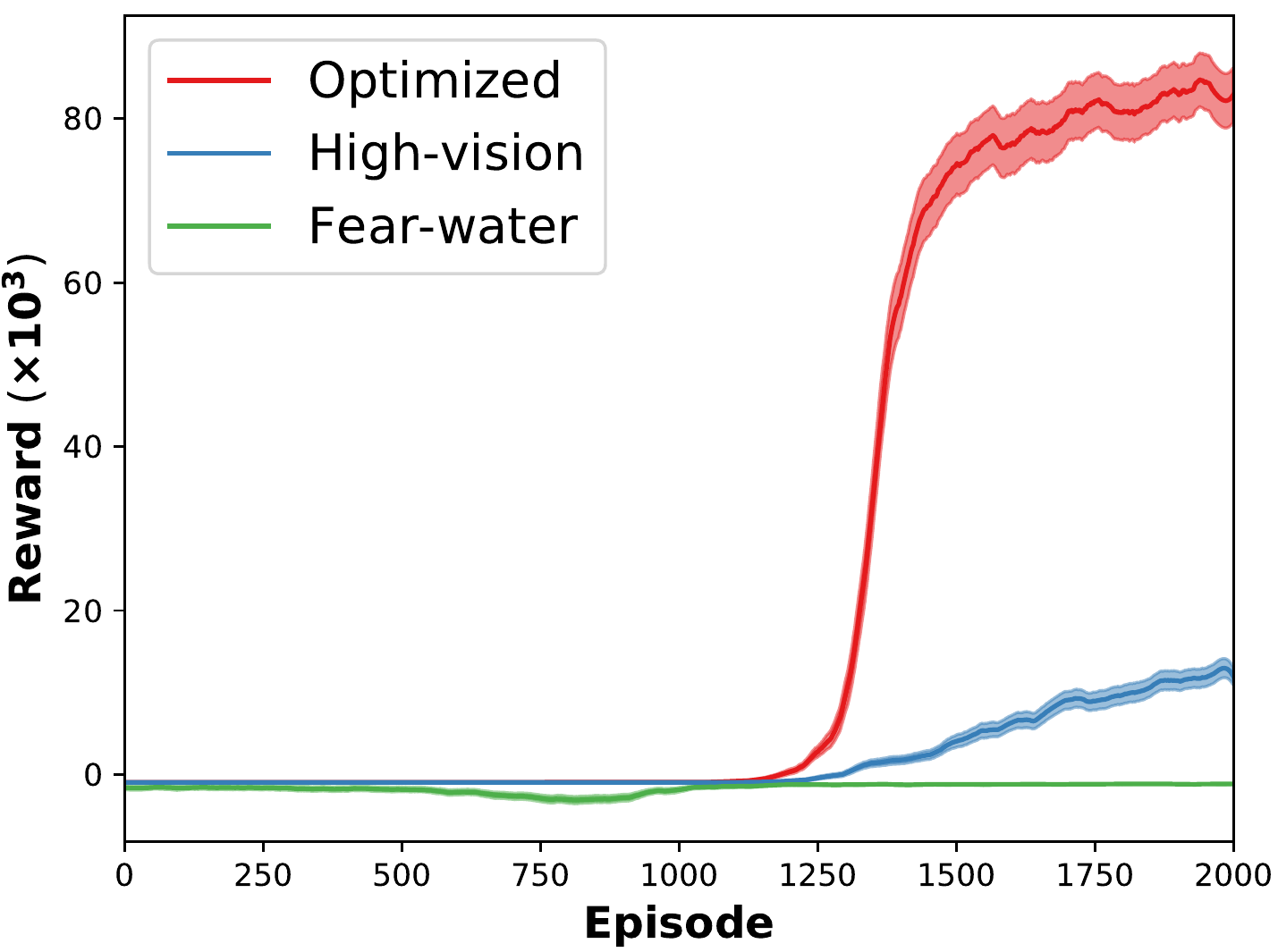}
		\caption{Total reward per episode}%
		\label{Fig:EvoReward}
        \end{subfigure}%
        \hspace{10pt}
	\begin{subfigure}[b]{0.4\columnwidth}
		\includegraphics[width=\textwidth]{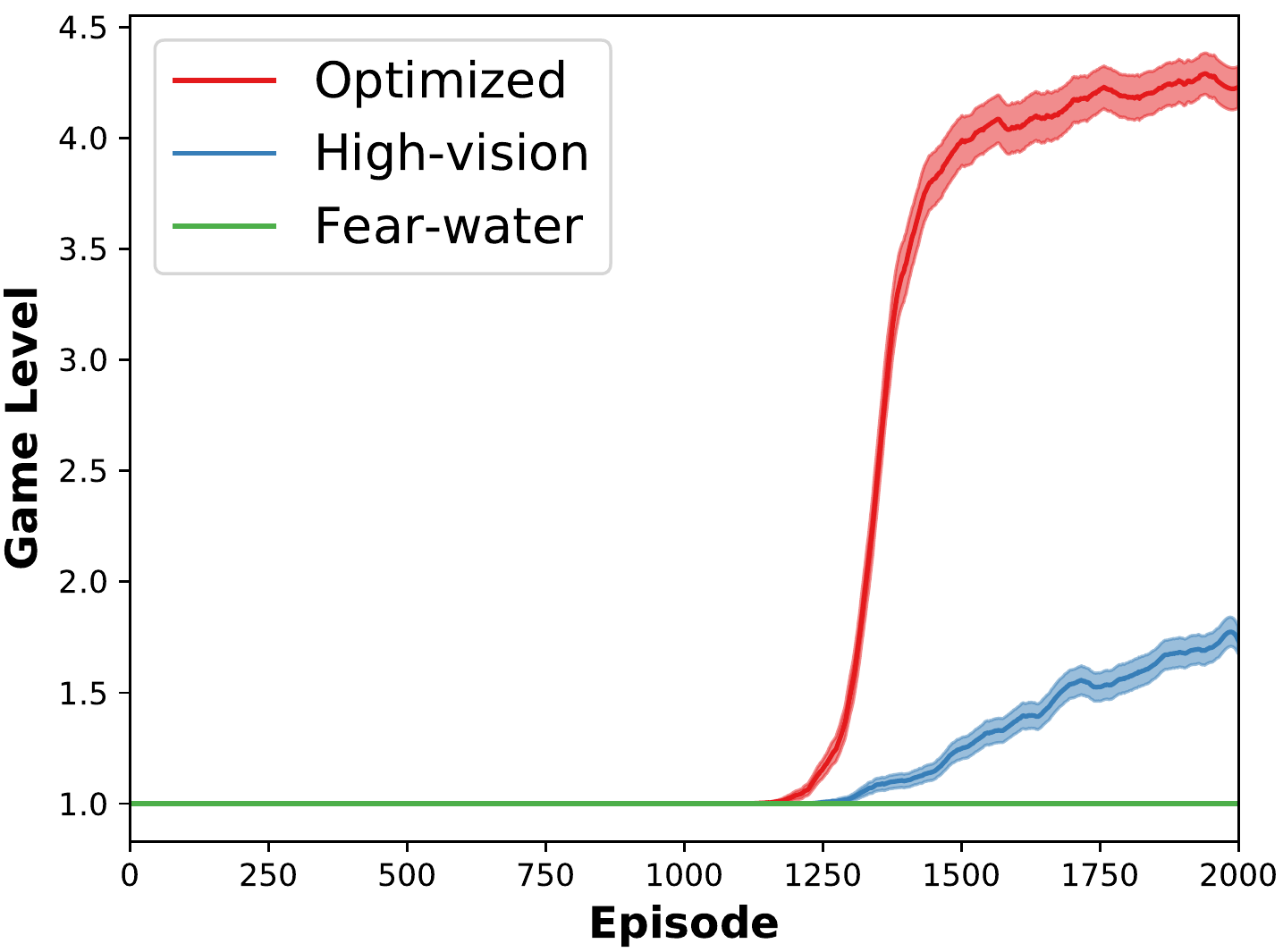}
		\caption{Game level}%
		\label{Fig:EvoGameLevel}
        \end{subfigure}
        \begin{subfigure}[b]{0.4\columnwidth}
		\includegraphics[width=\textwidth]{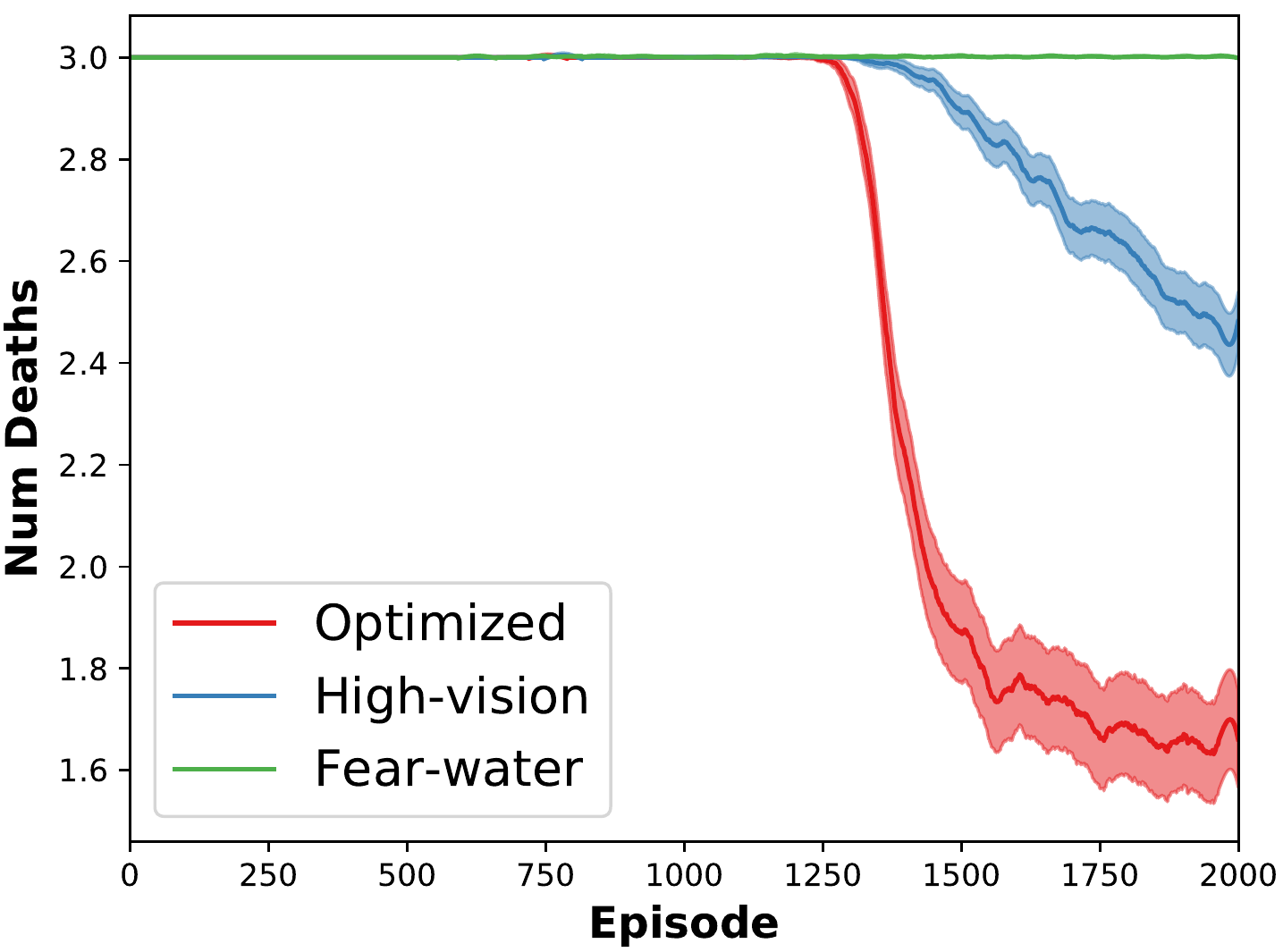}
		\caption{Num. deaths}%
		\label{Fig:EvoNumDeaths}
        \end{subfigure}%
        \hspace{10pt}
        \begin{subfigure}[b]{0.4\columnwidth}
		\includegraphics[width=\textwidth]{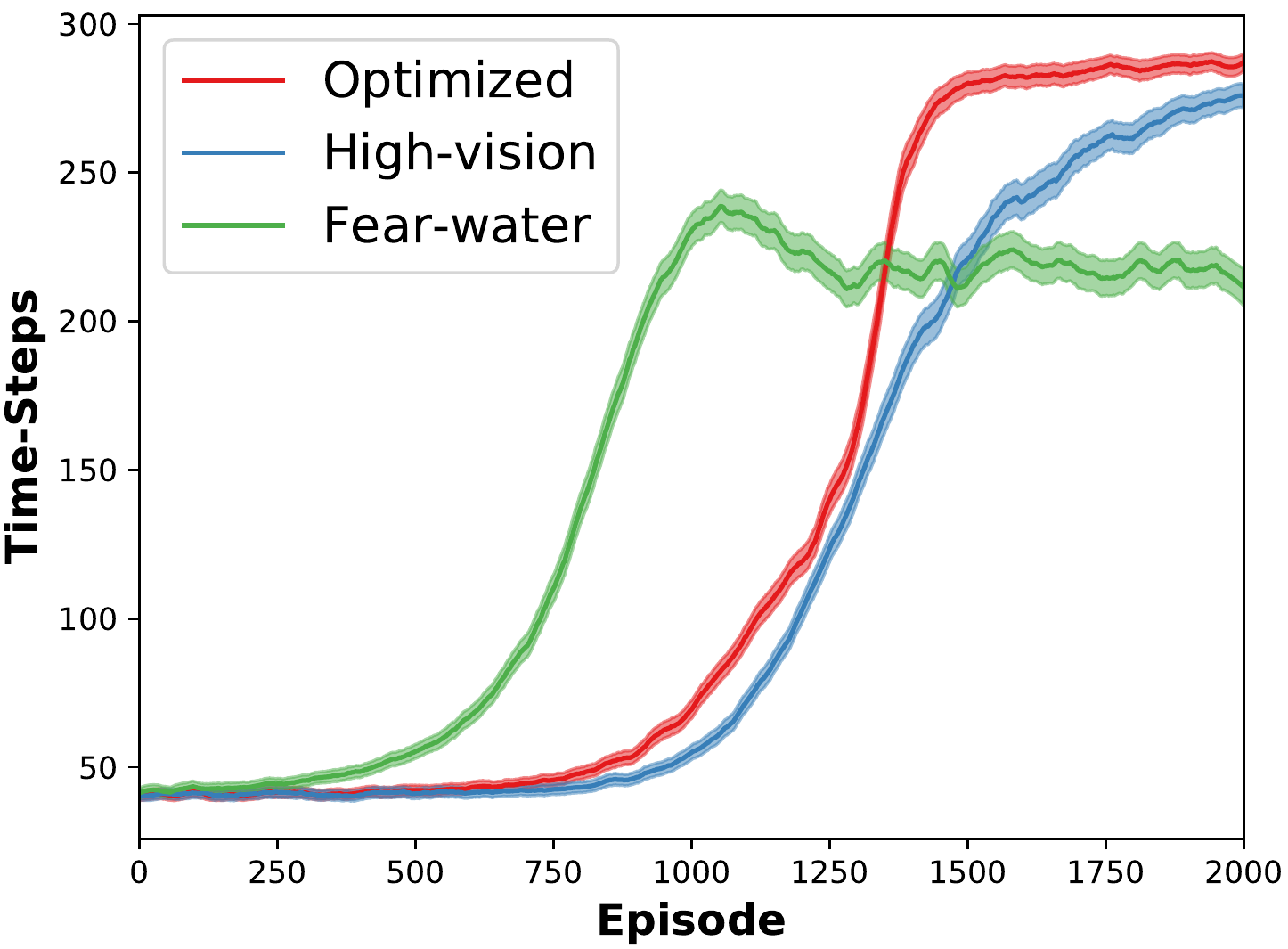}
		\caption{Time-steps per episode}%
		\label{Fig:EvoTimeSteps}
        \end{subfigure}
        \begin{subfigure}[b]{0.4\columnwidth}
		\includegraphics[width=\textwidth]{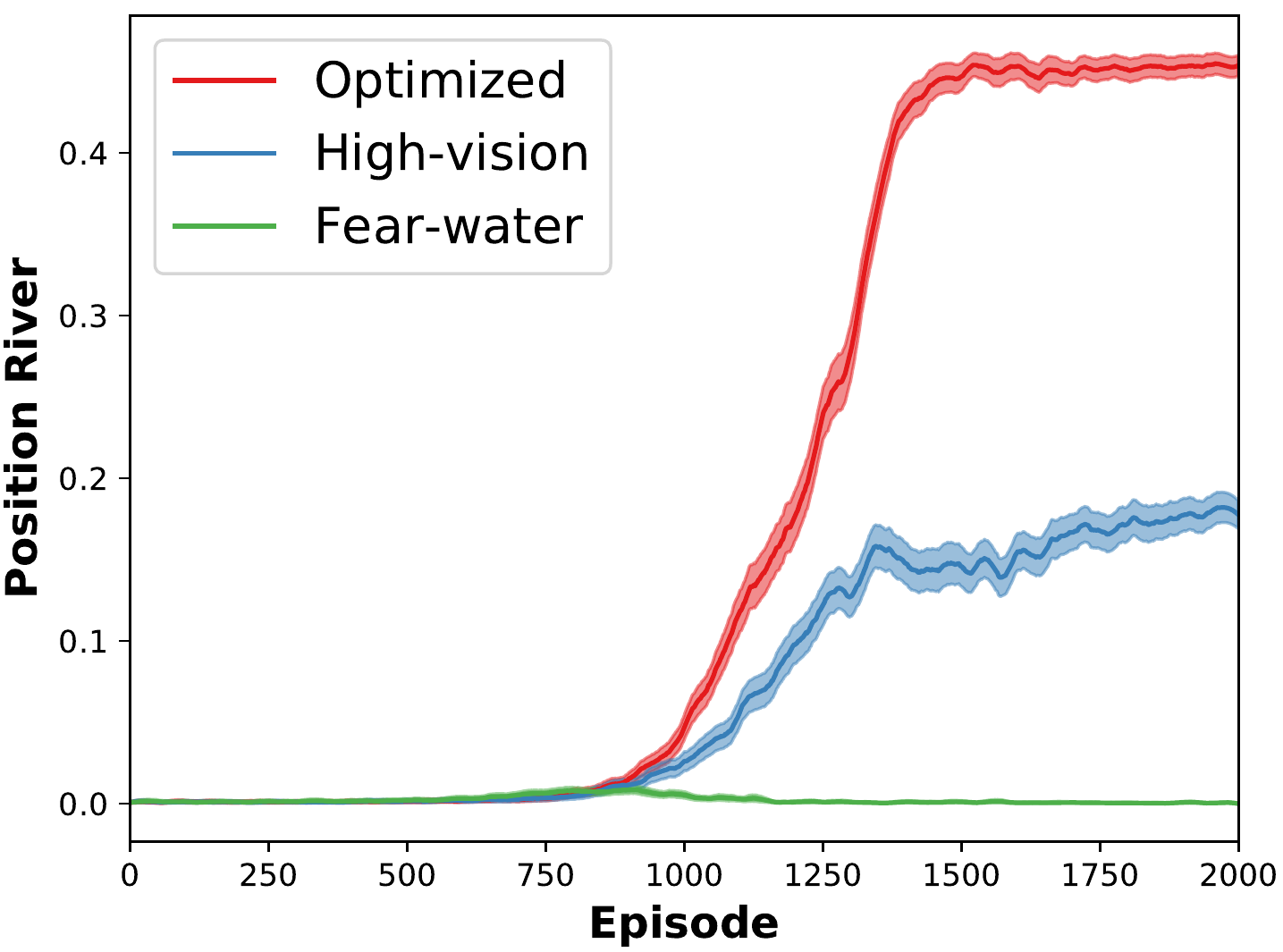}
		\caption{Agent in the river}%
		\label{Fig:EvoPosRiver}
        \end{subfigure}%
        \hspace{10pt}
        \begin{subfigure}[b]{0.4\columnwidth}
		\includegraphics[width=\textwidth]{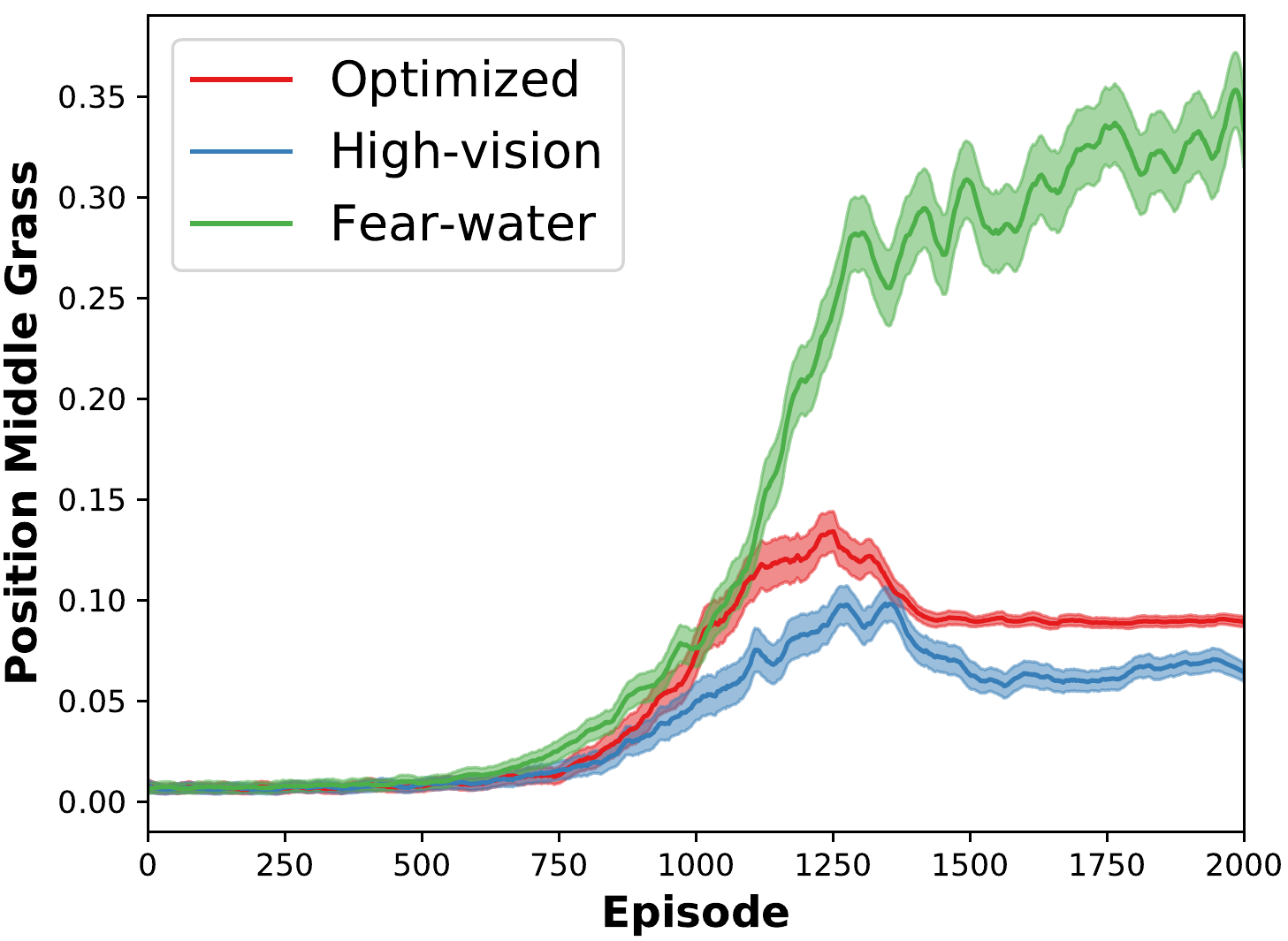}
		\caption{Agent in the middle grass row}%
		\label{Fig:EvoPosMiddle}
        \end{subfigure}
        \begin{subfigure}[b]{0.4\columnwidth}
		\includegraphics[width=\textwidth]{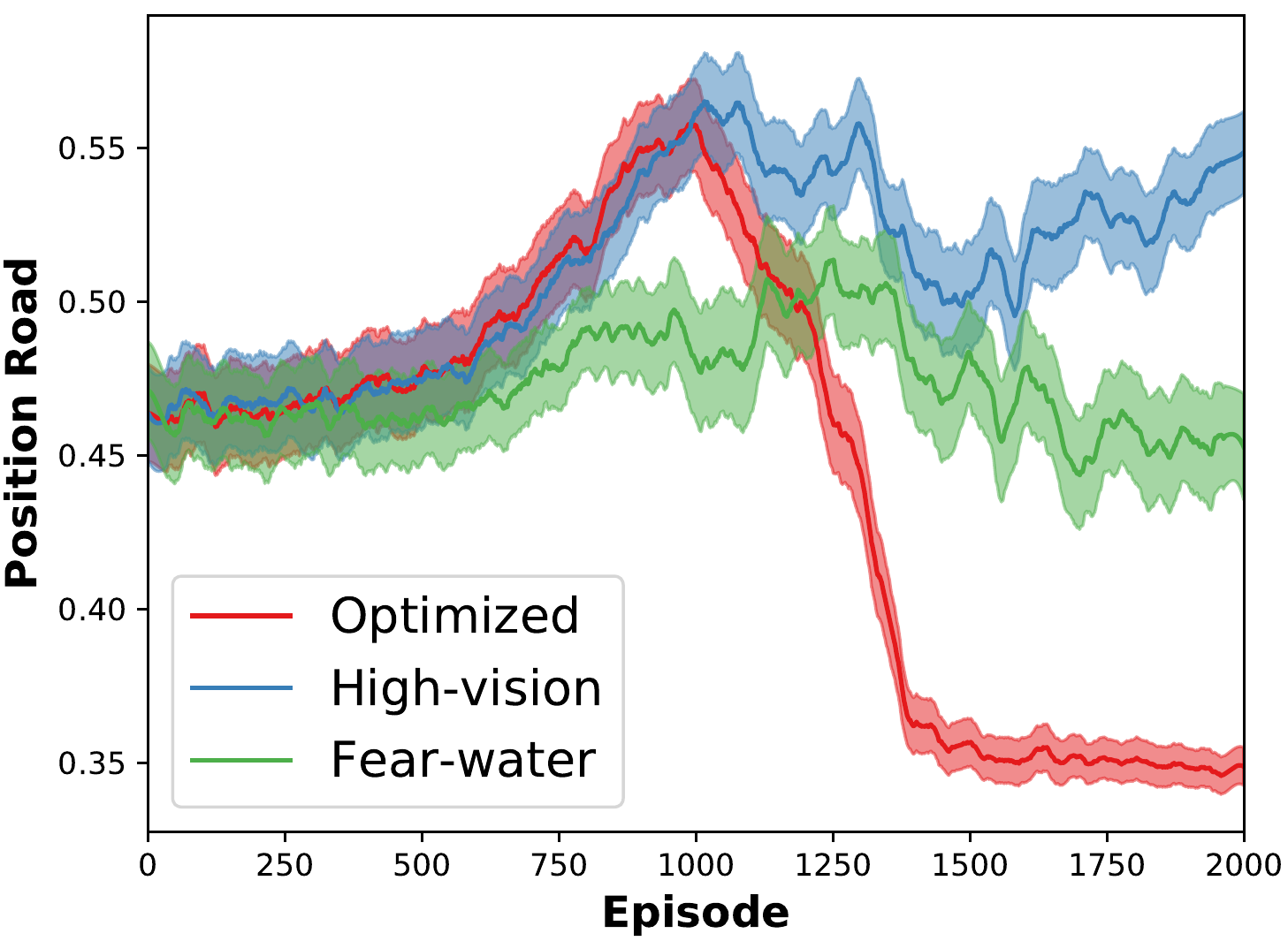}
		\caption{Agent in the road}%
		\label{Fig:EvoPosRoad}
        \end{subfigure}%
        \hspace{10pt}
        \begin{subfigure}[b]{0.4\columnwidth}
		\includegraphics[width=\textwidth]{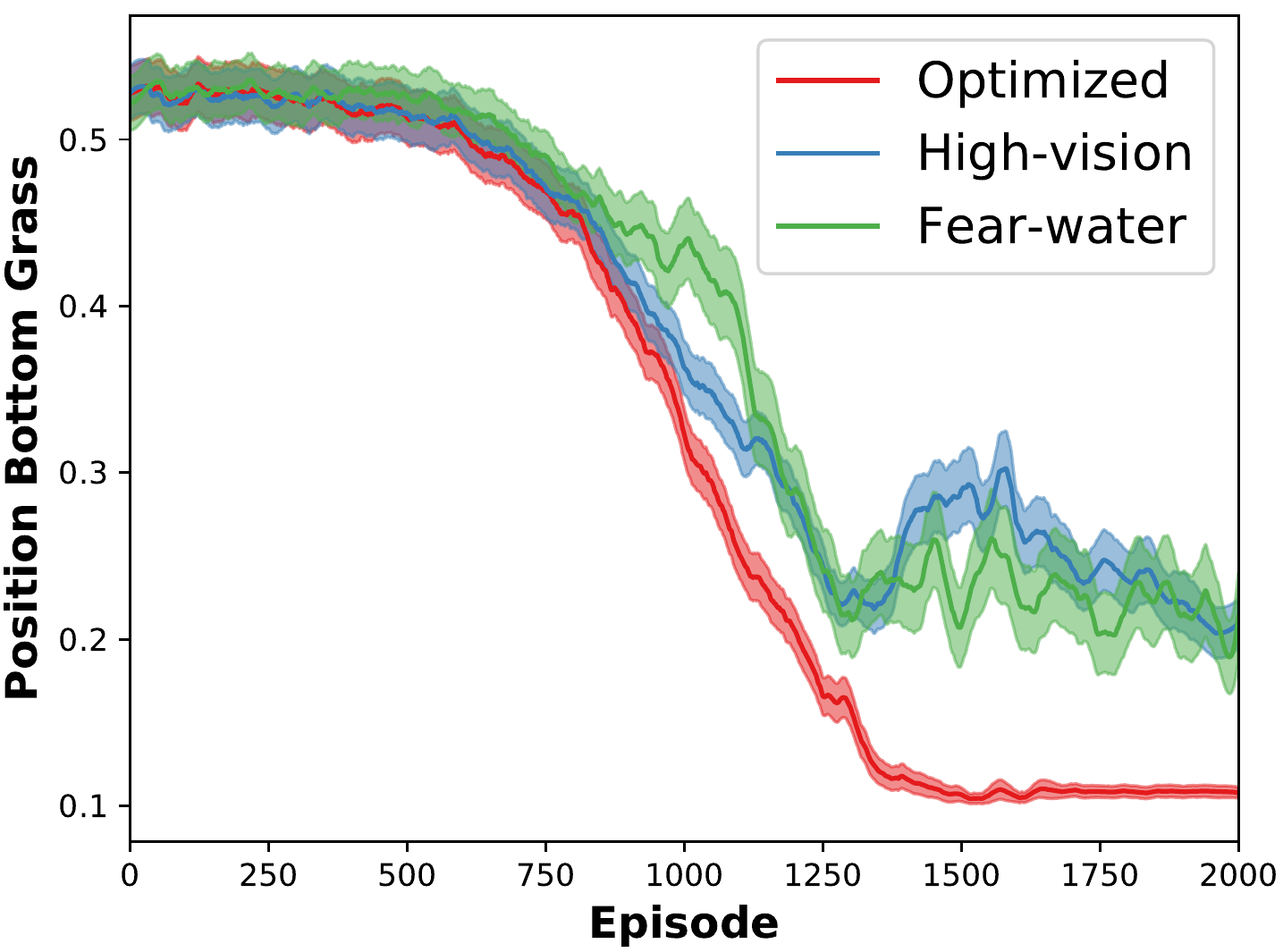}
		\caption{Ag. in the bottom grass row}%
		\label{Fig:EvoPosBottom}
        \end{subfigure}
    	\caption{Mean evolution of the agents' training procedure. Plots correspond to averages over $100$ training trials of $2,000$ episodes each. Shaded areas correspond to standard errors. Figs.~\protect\subref{Fig:EvoPosRiver}--\protect\subref{Fig:EvoPosBottom} represent the mean percentage of time spent by the agent in a specific region.}%
    	\label{Fig:TrainResults}
\end{figure}

Fig.~\ref{Fig:TrainResults} illustrates the learning performance of each agent averaged over $100$ trials. As we can see, the optimized agent learns about reaching the lilypads around five eights of the training procedure and quickly learns a policy allowing it to reach high game levels and significantly decrease the number of lost lives. The high-vision agent follows a similar but less pronounced learning curve---it can survive for as long as the optimized agent but, because it takes more time crossing the road (cf. Fig.~\ref{Fig:EvoPosRoad}), it does not reach such high levels. Contrary to these agents, the fear-water agent has a significantly poorer learning performance. As seen in Fig.~\ref{Fig:EvoTimeSteps}, it starts by surviving longer than the other two agents but when observing Figs.~\ref{Fig:EvoPosMiddle} and \ref{Fig:EvoPosRoad} we see that it spends most of its time on the road and in the middle grass region, eventually being hit by cars.

\subsection{Introspection Results}

\begin{table}[!tb]
	\caption{Statistics about the interestingness elements found by our framework for each agent in the Frogger task. Dimensions and elements included for evaluation in the user study are marked in bold. See the text for details on each criterion.}%
	\label{Tab:Introspection}
	\def\lendim{55pt}
	\def\lenelem{70pt}
	\def\lencrit{60pt}
	\def\lenagent{44pt}
	\def\spacelen{4pt}
	\centering
	\small
	\renewcommand{\arraystretch}{1.3}
	\setlength\tabcolsep{0pt}
	\begin{tabular}{ 
	L{\lendim}@{\hspace{8pt}} 
	L{\lenelem}@{\hspace{8pt}} 
	L{\lencrit}@{\hspace{\spacelen}}
	R{\lenagent}@{\hspace{\spacelen}} 
	R{\lenagent}@{\hspace{\spacelen}} 
	R{\lenagent} }
	\toprule
	\multirow{2}{*}{\textbf{Dimension}} & 
	\multirow{2}{*}{\textbf{Element}} & 
	\multirow{2}{*}{\textbf{Criterion}} & 
	\multicolumn{3}{c}{\textbf{Agent}}  \\
	&  &  & \textbf{Optim.} & \textbf{High-vis.} & \textbf{Fear-wat.} \\
	\midrule
	\multirow{2}{\lendim}{\emph{Transition Certainty}}
	& Certain trans. 		& $ \evenness_{za}\leq 0.03$ 		
	& $25$	& $26$ 	& $3$ \\ 
	& Uncertain trans. 		& $ \evenness_{za}\geq 0.9$
	& $9$	& $20$ 	& $5$ \\ 
	\emph{Reward} 
	& Reward outliers 		& $ \lambda_{\sigma} \geq 2.5$
	& $9$ 	& $7$ 	& $2$ \\ 
	\hdashline
	\multirow{6}{\lendim}{\textbf{\emph{Frequency}}} 
	& Obs. dispersion 		& --- 				
	& $0.78$	& $0.78$ 	& $0.73$ \\
	& Obs. coverage (\# unique obs.) 	& --- 	
	& $0.09\ (348)$ & $0.09\ (390)$ & $0.06\ (194)$ \\
	& \textbf{Frequent sit.} 	& $\cs(z)\geq 15\times 10^{3}$ 
	& $18$ & $19$ 	& $14$ \\
	& \textbf{Infrequent sit.} 	& $\cs(z)\leq 150$ 
	& $8$	& $20$ 	& $35$ \\ 
	& Assoc. feat.-sets 		& Jacc. $\geq 0.4$
	& $5$ 	& $6$ 	& $4$ \\
	& Feature-rules 		& lift $\geq 0.8$
	& $14$ 	& $3$ 	& $3$ \\
	\multirow{2}{\lendim}{\textbf{\emph{Execution Certainty}}} 
	& \textbf{Certain exec.} 	& $\evenness_{z}\leq 0.1$ 
	& $33$ 	& $10$ 	& $2$ \\
	& \textbf{Uncert. exec.}	& $\evenness_{z}\geq 0.85$ 
	& $8$ 	& $13$ 	& $14$ \\
	\emph{Recency} 
	& Ancient sit. 			& $\ts(z)\leq 0.4\ts_{\text{max}}$
	& $2$ 	& $6$ 	& $16$ \\
	\multirow{3}{\lendim}{\emph{Value}}
	& Value outliers 		& $ \lambda_{\sigma}\geq 2$ 
	& $8$ 	& $9$ 	& $2$ \\ 
	& Mean pred. error 		& ---	
	& $270\pm 277$ & $196\pm 193$ & $30\pm 96$ \\
	& Pred. outliers 		& $ \lambda_{\sigma}\geq 2$
	& $12$ 	& $27$ 	& $6$ \\ 
	\hdashline
	\multirow{2}{\lendim}{\textbf{\emph{Transition- Value}}} 
	& \textbf{Maxima} 		& --- 	
	& $7$	& $2$ 	& $3$ \\
	& \textbf{Minima} 		& --- 	
	& $4$ 	& $3$ 	& $2$ \\
	\textbf{\emph{Sequence}} 
	& \textbf{Certain seq.} 	& --- 
	& $28$ 	& $8$ 	& $6$ \\ 
	\multirow{2}{\lendim}{\emph{Contradiction}}
	& Contrad. values 		& --- 	
	& $8$ 	& $7$ 	& $1$ \\ 
	& Contrad. goals 		& --- 	
	& $0$ 	& $0$ 	& $3$ \\ 
 	\bottomrule
	\end{tabular}
\end{table}

A quantitative summary of the generated elements for each agent is listed in Table.~\ref{Tab:Introspection}. Overall, we can see that the optimized agent acquired a better understanding of the task by means of the RL training, which resulted mainly from an adequate parameterization. A first indication of that is the significantly higher number of certain sequences to sub-goals found. Additionally, this agent has more maxima and minima situations, which reflects a more solid knowledge in the situations it wants to achieve and avoid, respectively. Qualitatively, the introspection analysis for this agent determined as maxima situations where the frog is on a log facing the lilypads ($\phi_{N}=lilypad$) and as minima the re-start states, \ie after dying or putting a frog on a lilypad, and the bottom corners of the environment. As for execution certainty, this agent has less uncertain transitions and action executions when compared to the other agents. In particular, most of the situations occurring in the river, \ie involving jumping on logs, were considered certain while the most uncertain situations occur on the road and involve the frog being trapped by cars near the borders. As for the frequent observations, they occur both in the road---when seeing $empty$ or seeing $car$---and in the river when seeing $log$ in any direction. 

The results for the high-vision agent are very similar to those of the optimized agent with regards to maxima, minima, and certain-execution situations. However, this agent has significantly more uncertain situations than the other agents. Namely, these occur when executing action $N$ when $\phi_{N}=empty$ on the road near the left border---although the agent is good at seeing cars from afar, when it's near the border it cannot see cars entering the lane above, thus making it one of its most unpredictable situations.%
\footnote{The majority of car lanes in Frogger flow west to east.}
As for its most frequent situations, they occur while the agent is in the road and involve observing the presence of cars, which is expected given that it spends most of its time in that region and has a very high vision range. Also, because it tries to avoid the left side of the road, this agent tends to cross it on the right side and jumps on logs mostly in the right side of the river---as a result, the less frequent situations occur when the frog is on a log near the left border. 

The optimized and high-vision agents, having performed better in the task, covered more of the state space and their visits were more disperse comparing to agent fear-water. Interestingly, their mean prediction error, especially that of the optimized agent, is very high. This means that they were exposed to extreme situations during learning, where choosing the correct action makes a difference between receiving a very high or low reward.  

The results for the fear-water agent are quite different, as expected. Because the agent was not optimistically initialized, the world from its perspective is less interesting. First, it has fewer prediction error and value outliers. Also, due to its ``fear'' of dying in the river, its normal behavior is to cross the road on the right until reaching the up-most car lane, at which point it tries to avoid cars by going to the left. Its learned goal (absolute maximum) occurs when observing $z=\vect{water, car, empty, bounds}$. As for the absolute minimum it corresponds to being surrounded by cars. 
In addition, its uncertainty over the environment is different from that of the other agents---while it has less uncertain transitions, it has significantly more uncertain executions which results from a less certain learned policy. In particular, this agent is only certain about when to avoid cars on the road, but very uncertain when on the logs. Similarly, the frequent observations occur in the road and middle grass row. The fact that it has more infrequent and ancient situations is also related to the fact that it quickly learned to avoid jumping on logs to avoid the high penalty of falling on the river. 

Finally, with regards to contradictory goals, in Frogger these correspond to all maxima observations where $\phi_{N}\neq lilypad$. As expected, no unexpected goals were found for both the optimized and high-vision agents, while the goals of the fear-water agent are all unexpected, thereby denoting a learned policy not contingent with the intended task. 

\section{Survey Materials}%
\label{Sec:AppendixSurveyMaterials}

This appendix details the materials used in the user survey. Namely, we provide the instructions provided to subjects, the Frogger quiz and all questions and options used in the questionnaire.

\subsection{Frogger Instructions}

\begin{figure}[!tb]
	\centering
	\includegraphics[width=1.\columnwidth]{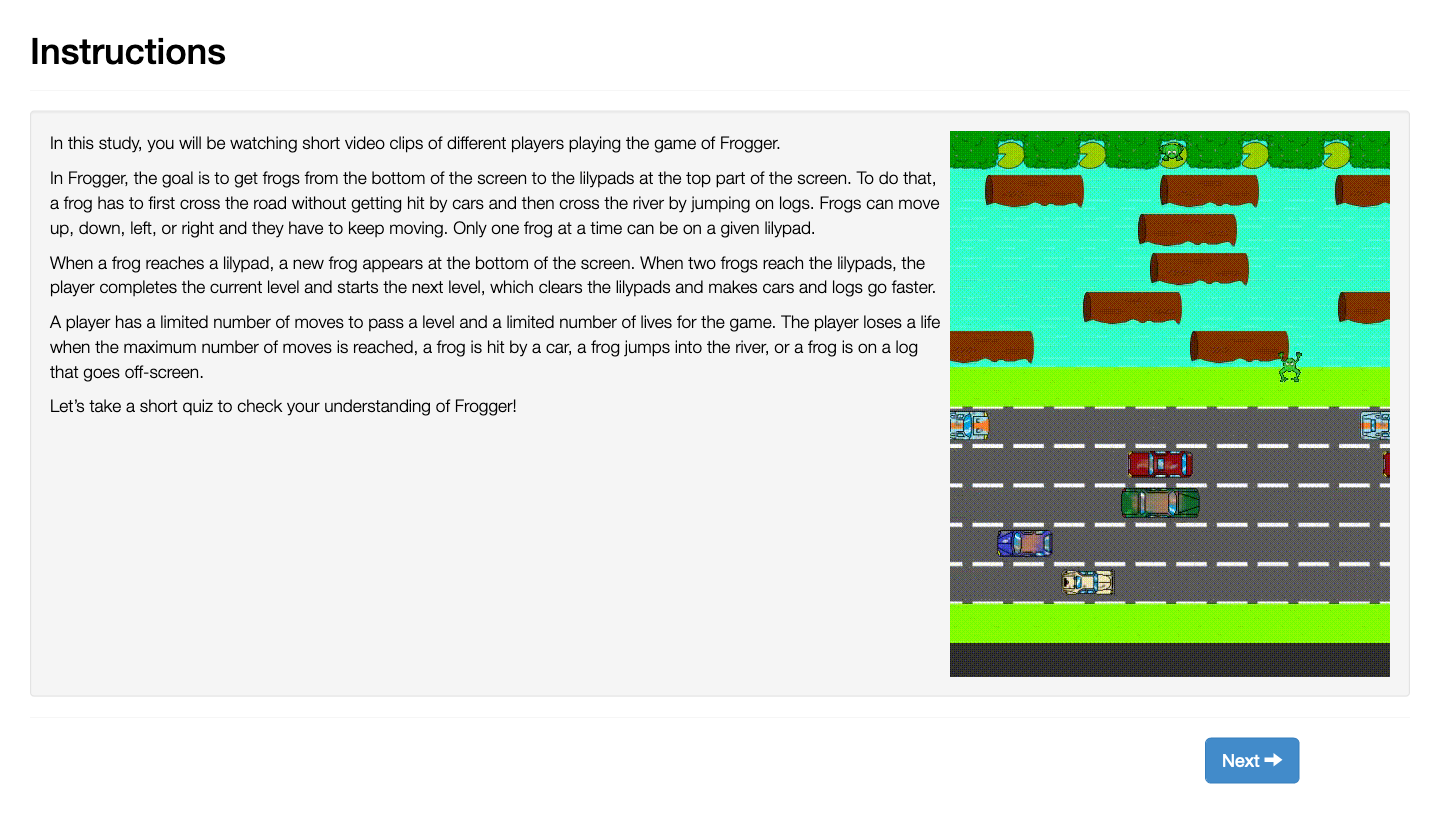}
	\caption{Frogger instructions presented to subjects of the user study.} 
	\label{Fig:FroggerInstructions}
\end{figure}

Subjects were first presented the screen in Fig.~\ref{Fig:FroggerInstructions} which contained instructions on the elements and dynamics of the game of Frogger (left). On the right, there was a looping video showing a frog being controlled and avoiding cars and jumping to logs until reaching a lilypad. For ease of reading, we transcribe the provided instructions below:

\begin{itquote}
In this study, you will be watching short video clips of different players playing the game of Frogger.

In Frogger, the goal is to get frogs from the bottom of the screen to the lilypads at the top part of the screen. To do that, a frog has to first cross the road without getting hit by cars and then cross the river by jumping on logs. Frogs can move up, down, left, or right and they have to keep moving. Only one frog at a time can be on a given lilypad.

When a frog reaches a lilypad, a new frog appears at the bottom of the screen. When two frogs reach the lilypads, the player completes the current level and starts the next level, which clears the lilypads and makes cars and logs go faster.

A player has a limited number of moves to pass a level and a limited number of lives for the game. The player loses a life when the maximum number of moves is reached, a frog is hit by a car, a frog jumps into the river, or a frog is on a log that goes off-screen.

Let’s take a short quiz to check your understanding of Frogger!
\end{itquote}

\subsection{Frogger Quiz}

\begin{figure}[!tb]
	\centering
	\includegraphics[width=1.\columnwidth]{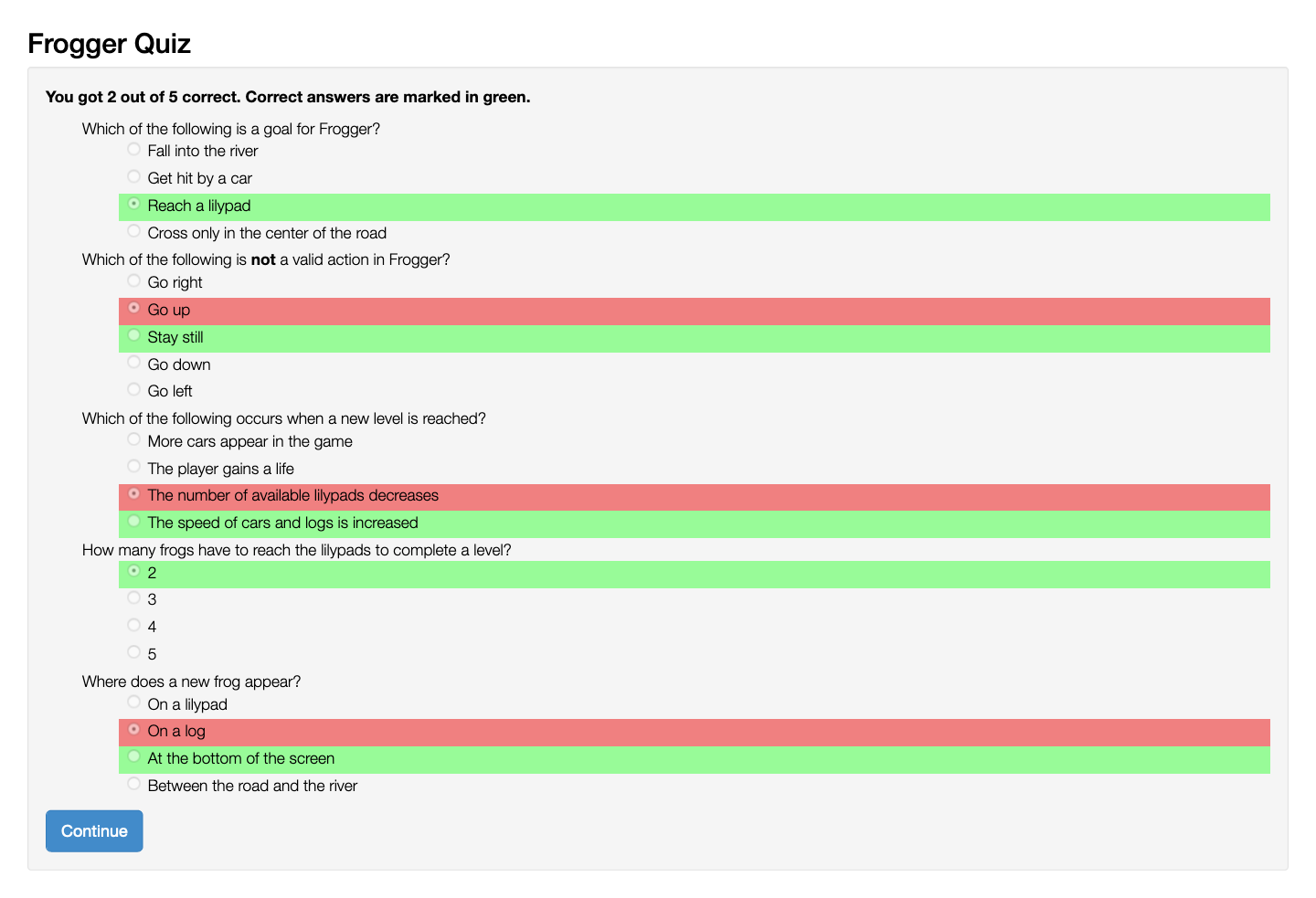}
	\caption{Frogger quiz presented to subjects of the user study with correct answers marked in a green color and incorrect answers marked with a red color.} 
	\label{Fig:FroggerQuiz}
\end{figure}

Participants were then presented a quiz about the game of Frogger as show in Fig.~\ref{Fig:FroggerQuiz}. The goal was to assess their knowledge on the task. Incorrect answers were marked with a red color, while correct answers appeared with a green background. The questions presented to the participants and corresponding options are transcribed below:

\begin{itquote}
Please answer the following questions about the game of Frogger.

Which of the following is a goal for Frogger?
\begin{itemize}
    \item Fall into the river
    \item Get hit by a car
    \item Reach a lilypad
    \item Cross only in the center of the road
\end{itemize}

Which of the following is not a valid action in Frogger?
\begin{itemize}
    \item Go right
    \item Go up
    \item Stay still
    \item Go down
    \item Go left
\end{itemize}

Which of the following occurs when a new level is reached?
\begin{itemize}
    \item More cars appear in the game
    \item The player gains a life
    \item The number of available lilypads decreases
    \item The speed of cars and logs is increased
\end{itemize}
    
How many frogs have to reach the lilypads to complete a level?
\begin{itemize}
    \item 2
    \item 3
    \item 4
    \item 5
\end{itemize}

Where does a new frog appear?
\begin{itemize}
    \item On a lilypad
    \item On a log
    \item At the bottom of the screen
    \item Between the road and the river
\end{itemize}    
\end{itquote}

As depicted in Fig.~\ref{Fig:FroggerQuiz}, after answering all questions and pressing the ``Continue'' button, subjects were presented with the following message on the top of the screen: \blockquote{\emph{You got $x$ out of 5 correct. Correct answers are marked in green.}} Subjects had to select the correct answers to continue with the study. As explained in the main paper, we recorded the number of incorrect answers and used it to filter out the responses of certain participants.

\subsection{Questionnaire Instructions}

\begin{figure}[!tb]
	\centering
	\includegraphics[width=1.\columnwidth]{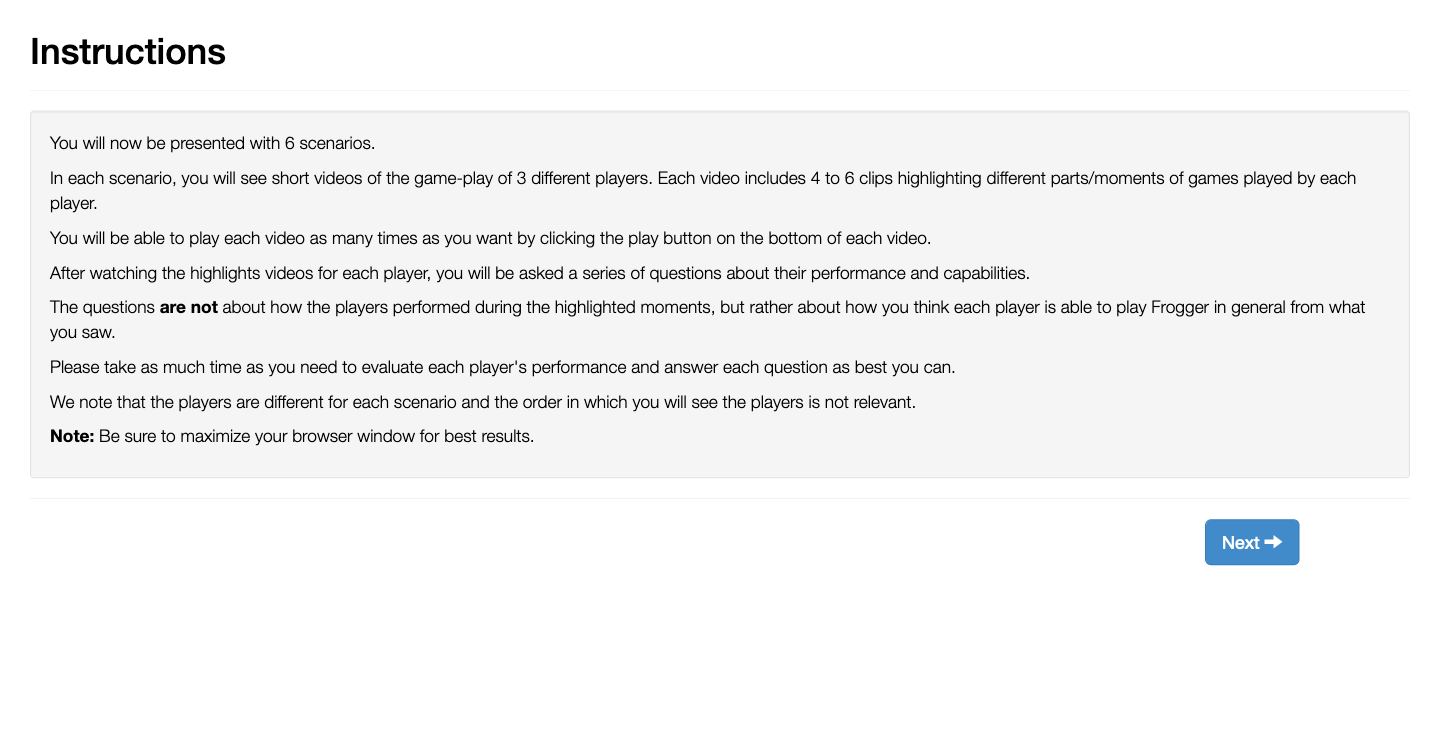}
	\caption{Frogger quiz presented to subjects of the user study with correct answers marked in a green color and incorrect answers marked with a red color.} 
	\label{Fig:FroggerQuestionnnaireInstructions}
\end{figure}

The screen shown in Fig.~\ref{Fig:FroggerQuestionnnaireInstructions} was then presented to participants. It provided detailed instructions on the questionnaire regarding the 6 scenarios presented thereafter. In particular, the instructions were as follows:

\begin{itquote}
You will now be presented with 6 scenarios.

In each scenario, you will see short videos of the game-play of 3 different players. Each video includes 4 to 6 clips highlighting different parts/moments of games played by each player.

You will be able to play each video as many times as you want by clicking the play button on the bottom of each video.

After watching the highlights videos for each player, you will be asked a series of questions about their performance and capabilities.

The questions \textbf{are not} about how the players performed during the highlighted moments, but rather about how you think each player is able to play Frogger in general from what you saw.

Please take as much time as you need to evaluate each player's performance and answer each question as best you can.

We note that the players are different for each scenario and the order in which you will see the players is not relevant.

\textbf{Note:} Be sure to maximize your browser window for best results.
\end{itquote}

\subsection{Questionnaire}

\begin{figure}[!tb]
	\centering
	\includegraphics[width=1.\columnwidth]{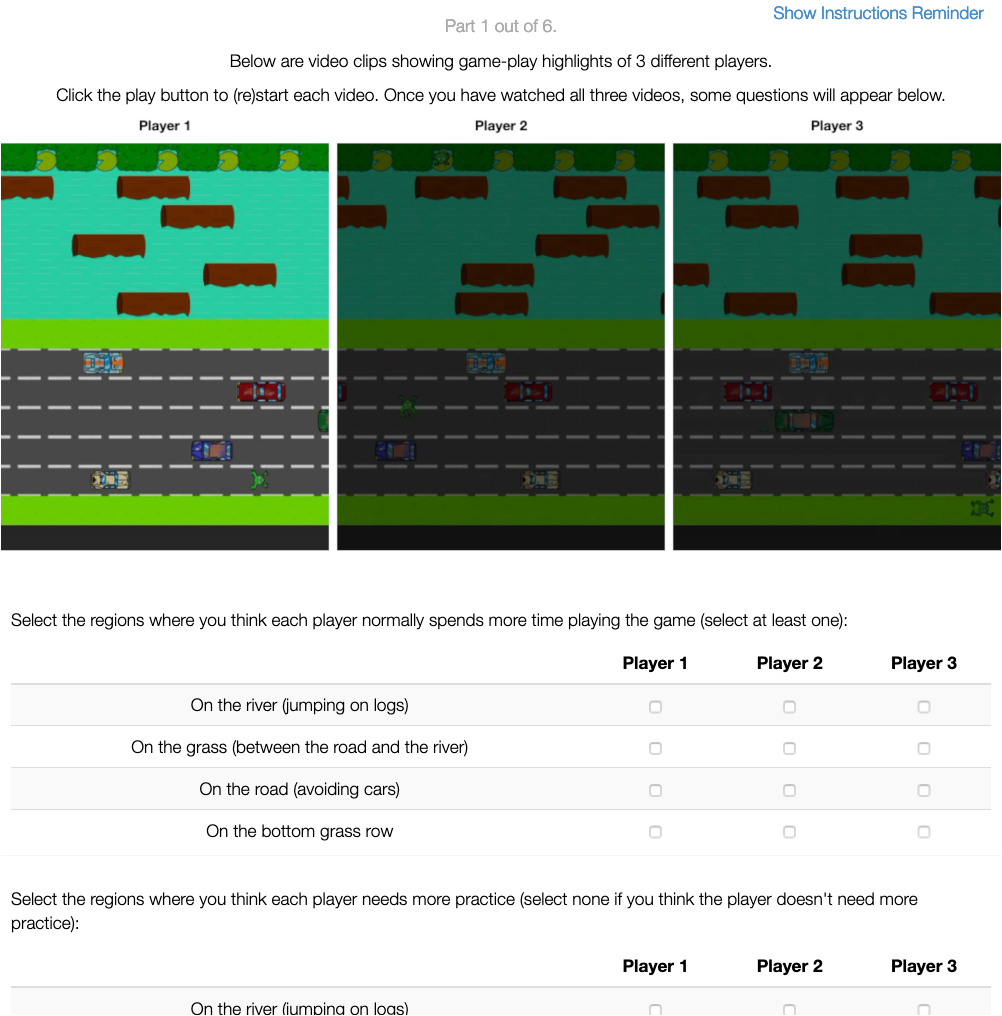}
	\caption{Questionnaire used in the user study. Copied from the main paper for convenience.} 
	\label{Fig:FroggerQuestionnaire}
\end{figure}

After reading the instructions, subjects were presented with the questionnaire itself, part of which can be seen in Fig.~\ref{Fig:FroggerQuestionnaire}. Participants responded to the same questionnaire for $6$ different scenarios as detailed in the main paper. As seen in Fig.~\ref{Fig:FroggerQuestionnaire}, a label indicating the progress appeared at the top of the page stating \blockquote{\emph{Part $n$ of 6.}} Also, by clicking on the \blockquote{\emph{Show Instructions Reminder}} link at the top-right corner of the page, the instructions text as detailed in the previous subsection appeared on the screen. 

Below we provide a transcription of each question as it appeared in the questionnaire, as well as the options that participants could choose from. Subjects were first presented with two \emph{region-related} questions, namely:

\begin{itquote}
Select the regions where you think each player normally spends more time playing the game (select at least one): 

\begin{itemize}
    \item On the river (jumping on logs)			
    \item On the grass (between the road and the river)			
    \item On the road (avoiding cars)			
    \item On the bottom grass row
\end{itemize}

Select the regions where you think each player needs more practice (select none if you think the player doesn't need more practice):

\begin{itemize}
    \item On the river (jumping on logs)			
    \item On the grass (between the road and the river)			
    \item On the road (avoiding cars)			
    \item On the bottom grass row
\end{itemize}

\end{itquote}

For the above set of questions, participants had to choose at least one region for each player. Columns were labeled \blockquote{\emph{Player 1}}, \blockquote{\emph{Player 2}} and \blockquote{\emph{Player 3}}. Then, we presented two \emph{aptitude-related} questions:

\begin{itquote}

Based on what you have seen of each player, rate the following statements with regards to how much you agree with them. 

``The player is capable of reaching advanced levels.''

\begin{itemize}
    \item Player 1					
    \item Player 2					
    \item Player 3
\end{itemize}

``The player needs help to be better at the game.''

\begin{itemize}
    \item Player 1					
    \item Player 2					
    \item Player 3
\end{itemize}

\end{itquote}

To answer the above questions, subjects had to rate their agreement to each statement by choosing one option for each player according to a 5-point Likert scale. The options were (left-to-right): \blockquote{\emph{Strongly Disagree}}, \blockquote{\emph{Disagree}}, \blockquote{\emph{Neither Agree nor Disagree}}, \blockquote{\emph{Agree}} and \blockquote{\emph{Strongly Agree}}.

Finally, the last question assessed the participants' confidence in responding to the questionnaire. Specifically, they had to answer the following question:

\begin{itquote}
How confident were you in answering the responses for this scenario?
\end{itquote}

Again, a 5-point Likert scale was used with the following options (left-to-right): \blockquote{\emph{Not Confident}}, \blockquote{\emph{Slightly Confident}}, \blockquote{\emph{Somewhat Confident}}, \blockquote{\emph{Confident}} and \blockquote{\emph{Very Confident}}.

\section{Survey Responses}%
\label{Sec:AppendixUserResponses}

\begin{figure}[!tb]
	\centering
        \begin{subfigure}[b]{0.33\columnwidth}
		\includegraphics[width=\textwidth]{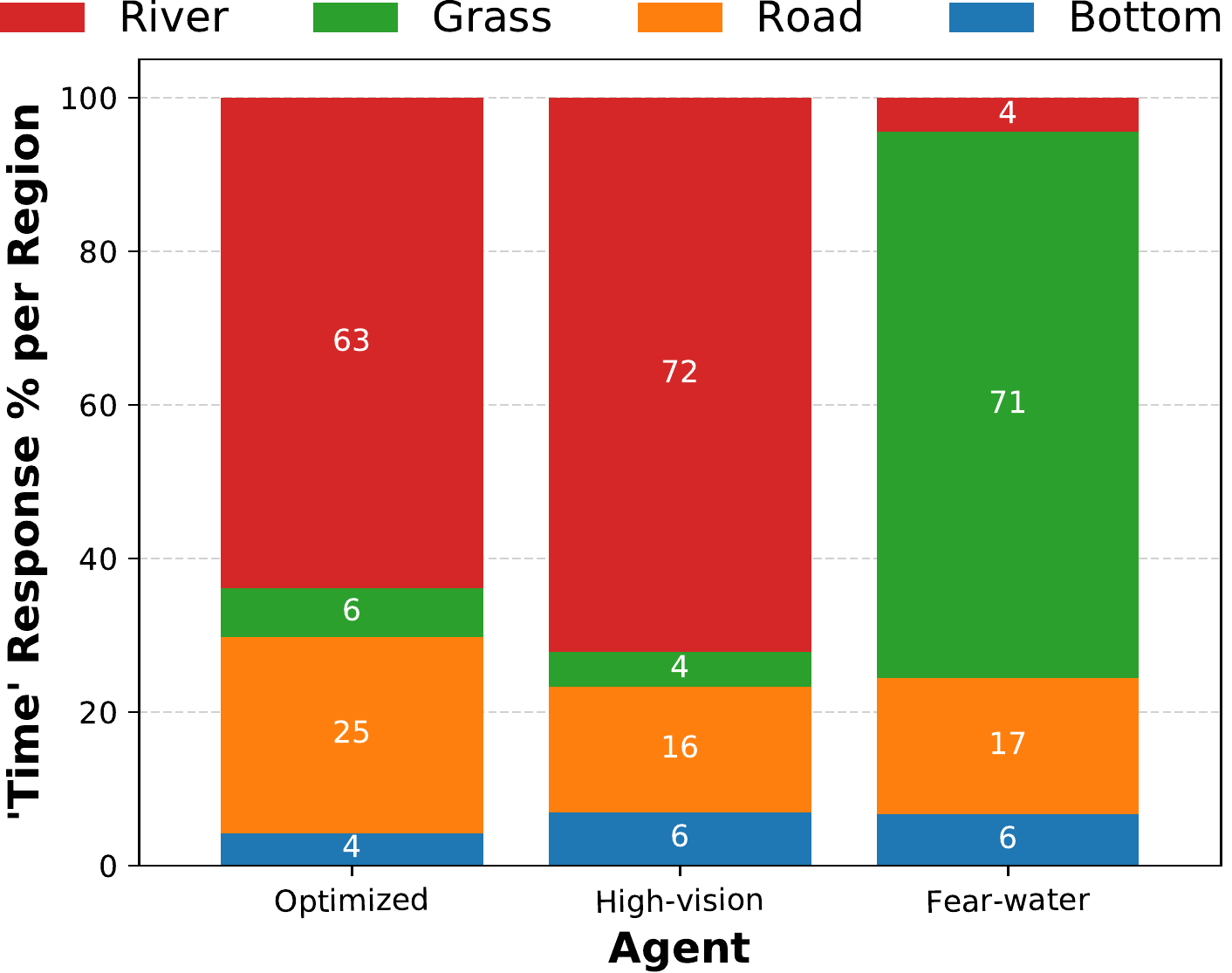}
		\caption{Scenario 1}
        \end{subfigure}%
	\begin{subfigure}[b]{0.33\columnwidth}
		\includegraphics[width=\textwidth]{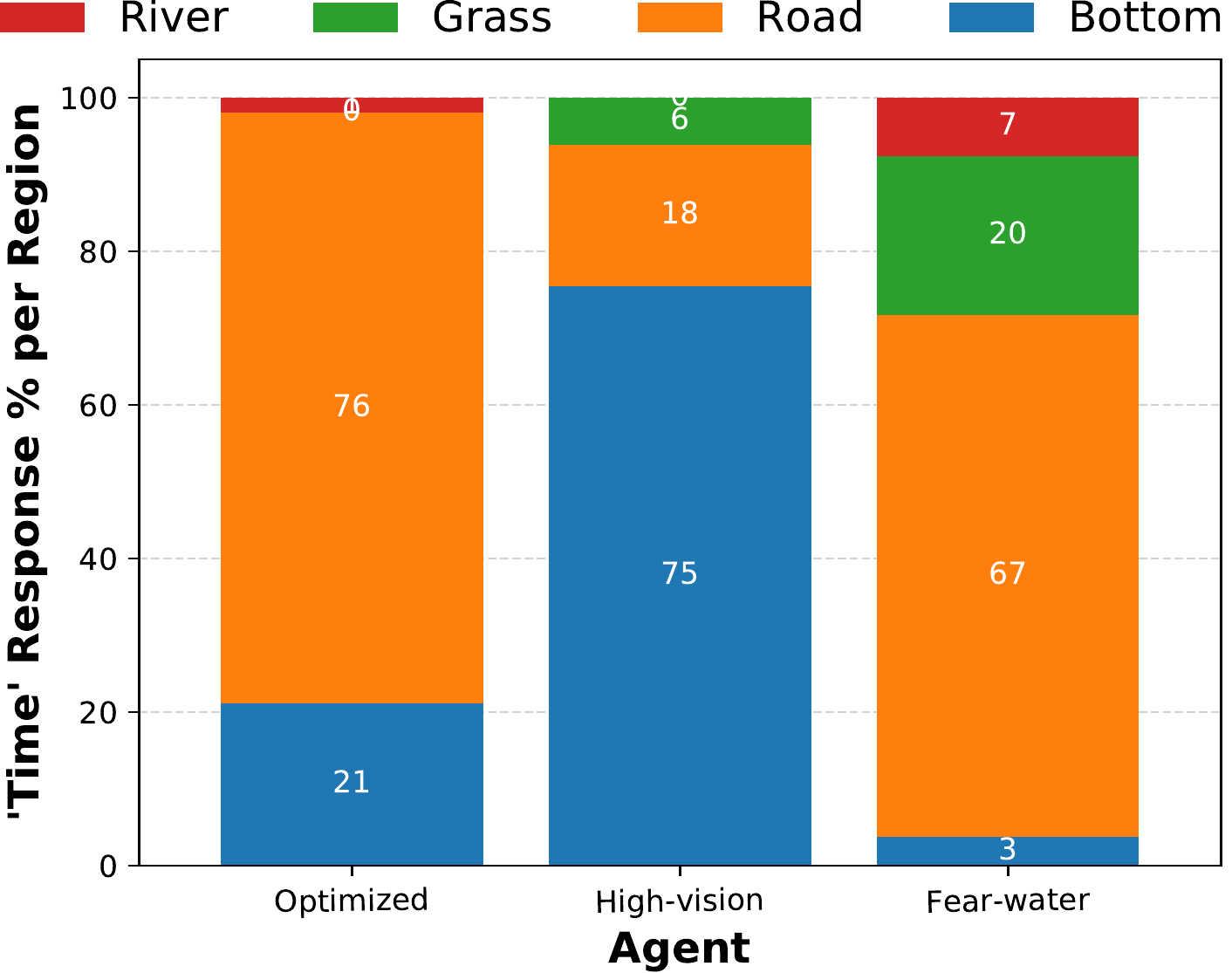}
		\caption{Scenario 2}
        \end{subfigure}%
        \begin{subfigure}[b]{0.33\columnwidth}
		\includegraphics[width=\textwidth]{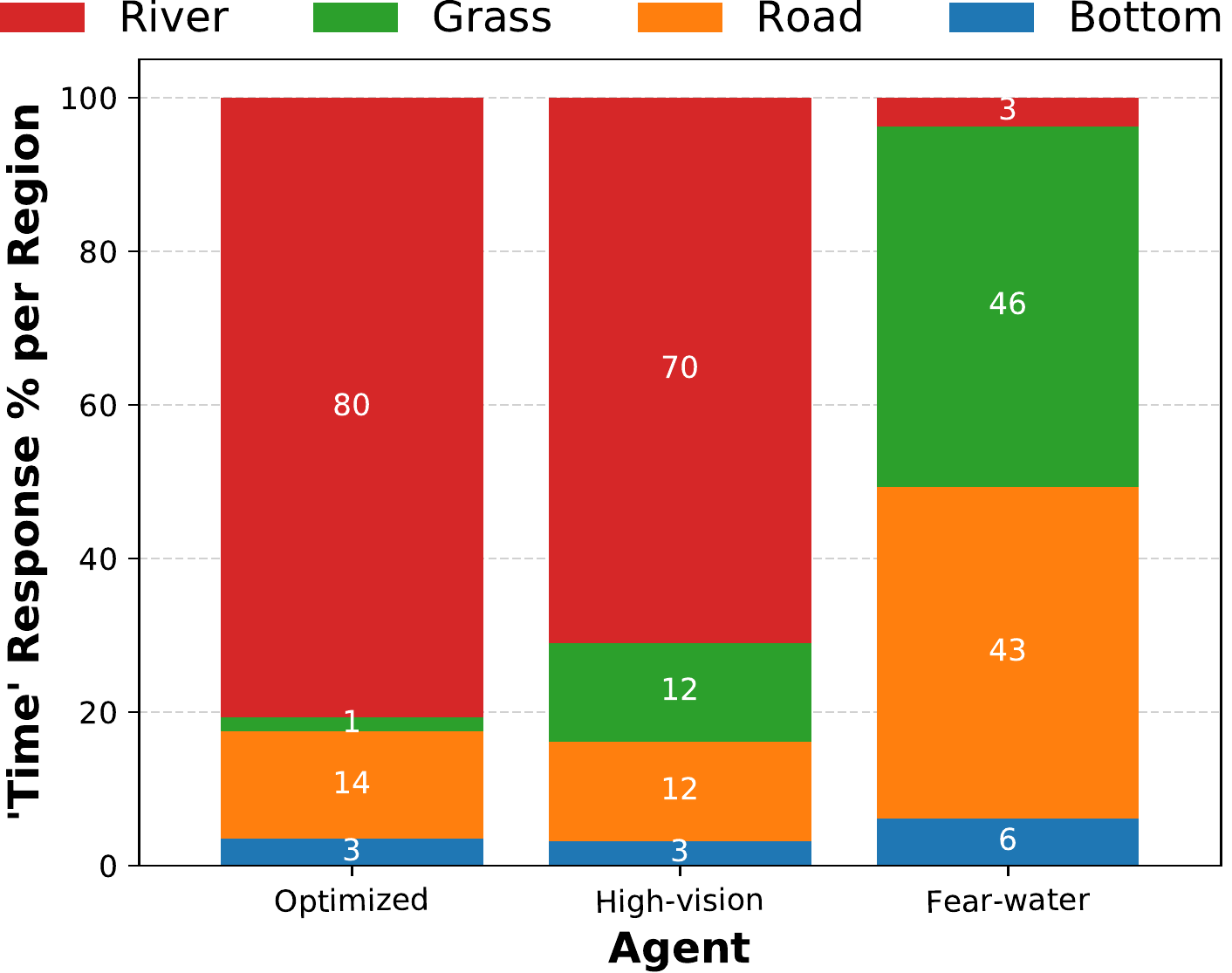}
		\caption{Scenario 3}
        \end{subfigure}
        \begin{subfigure}[b]{0.33\columnwidth}
		\includegraphics[width=\textwidth]{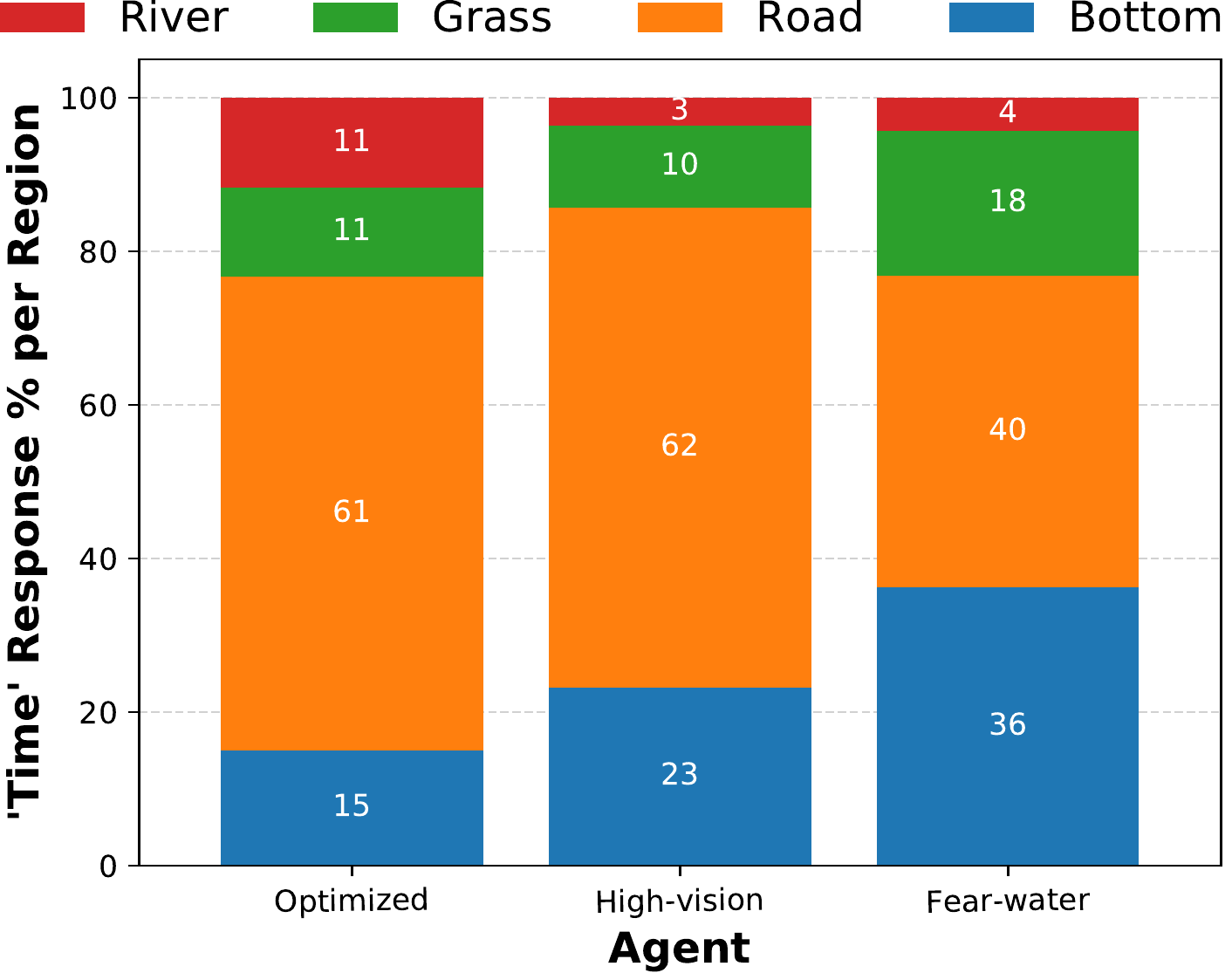}
		\caption{Scenario 4}
        \end{subfigure}%
	\begin{subfigure}[b]{0.33\columnwidth}
		\includegraphics[width=\textwidth]{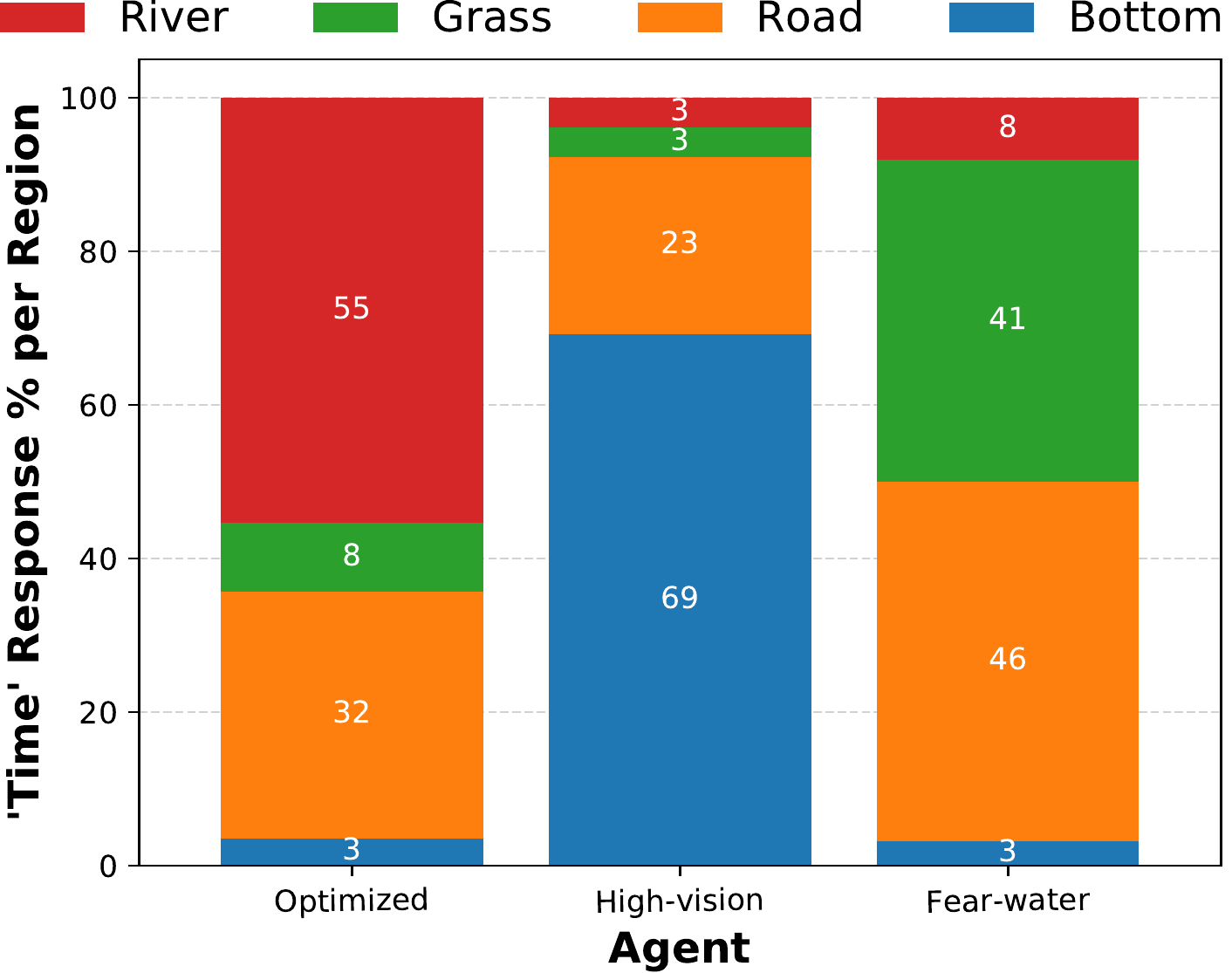}
		\caption{Scenario 5}
        \end{subfigure}%
        \begin{subfigure}[b]{0.33\columnwidth}
		\includegraphics[width=\textwidth]{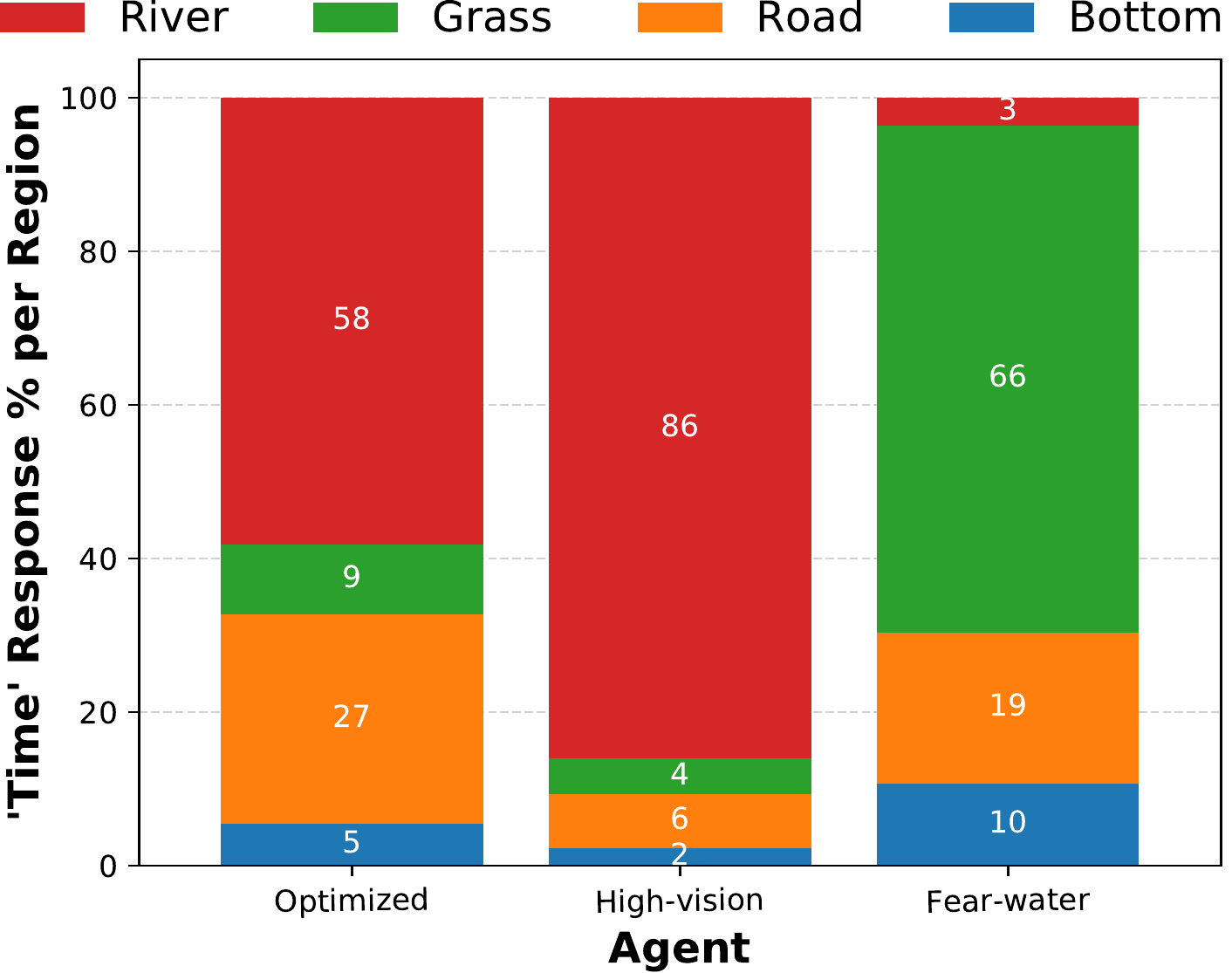}
		\caption{Scenario 6}
        \end{subfigure}
        \begin{subfigure}[b]{0.33\columnwidth}
		\includegraphics[width=\textwidth]{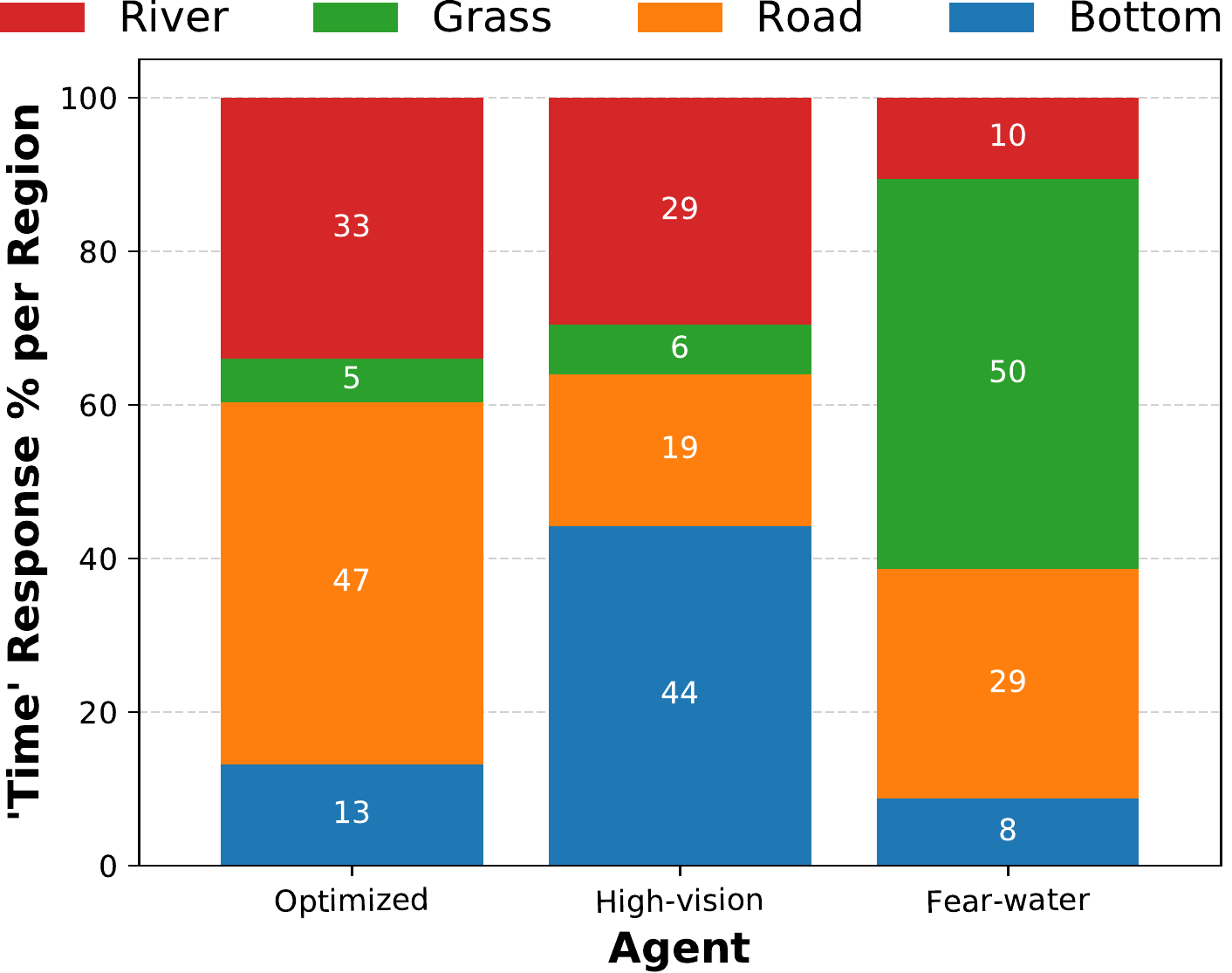}
		\caption{Scenario 7}
        \end{subfigure}%
	\begin{subfigure}[b]{0.33\columnwidth}
		\includegraphics[width=\textwidth]{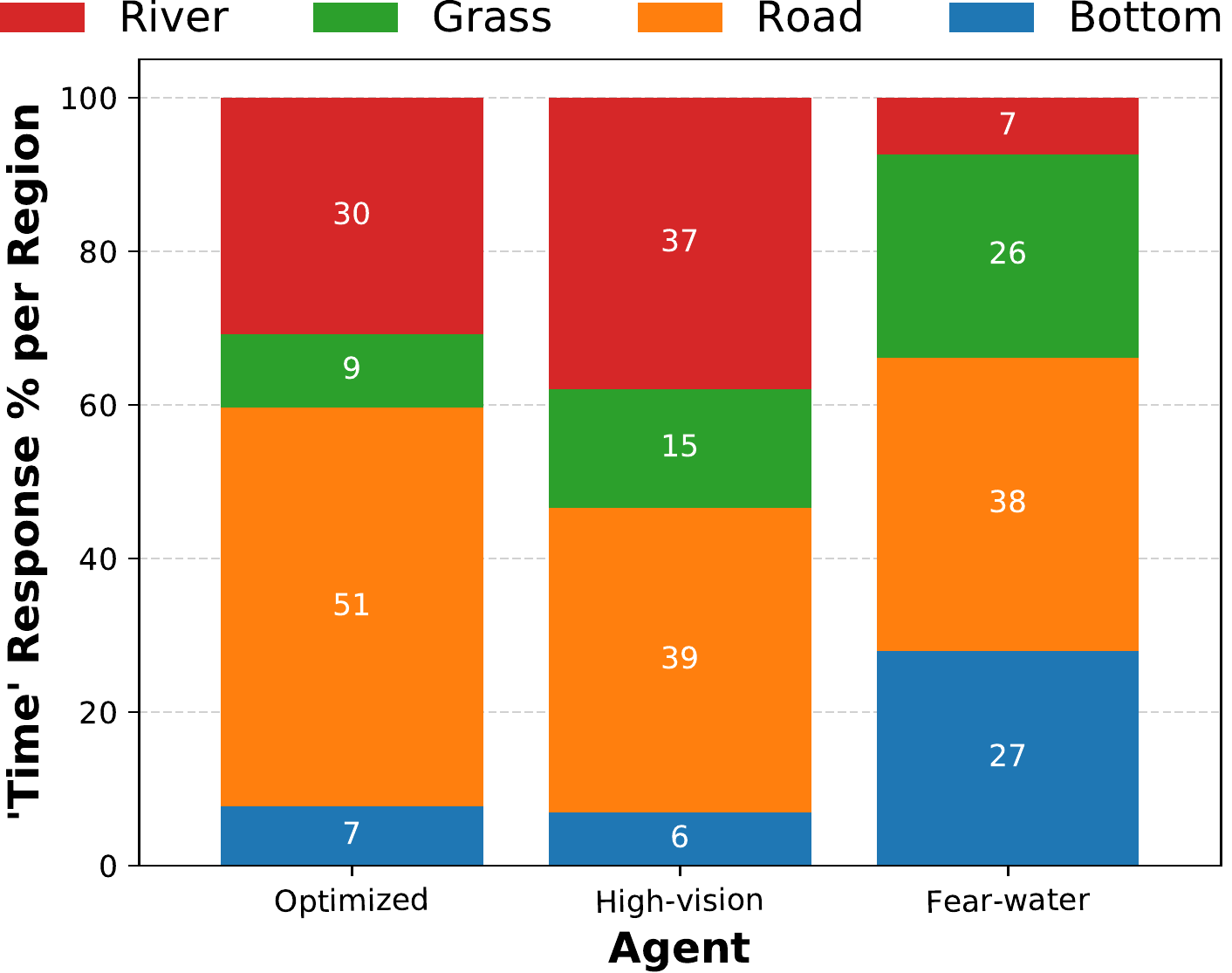}
		\caption{Scenario 8}
        \end{subfigure}%
        \begin{subfigure}[b]{0.33\columnwidth}
		\includegraphics[width=\textwidth]{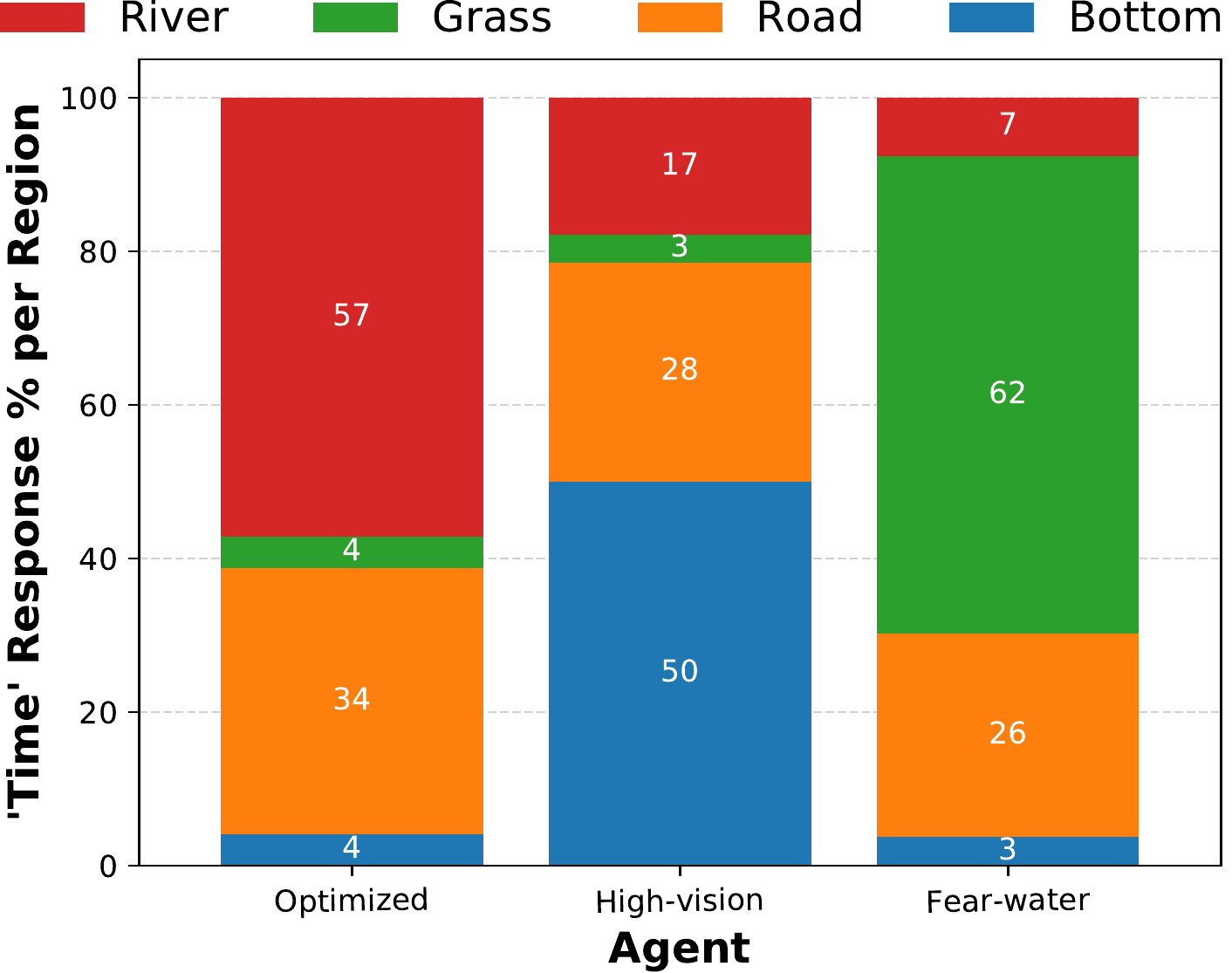}
		\caption{Scenario 9}
        \end{subfigure}
        \begin{subfigure}[b]{0.33\columnwidth}
		\includegraphics[width=\textwidth]{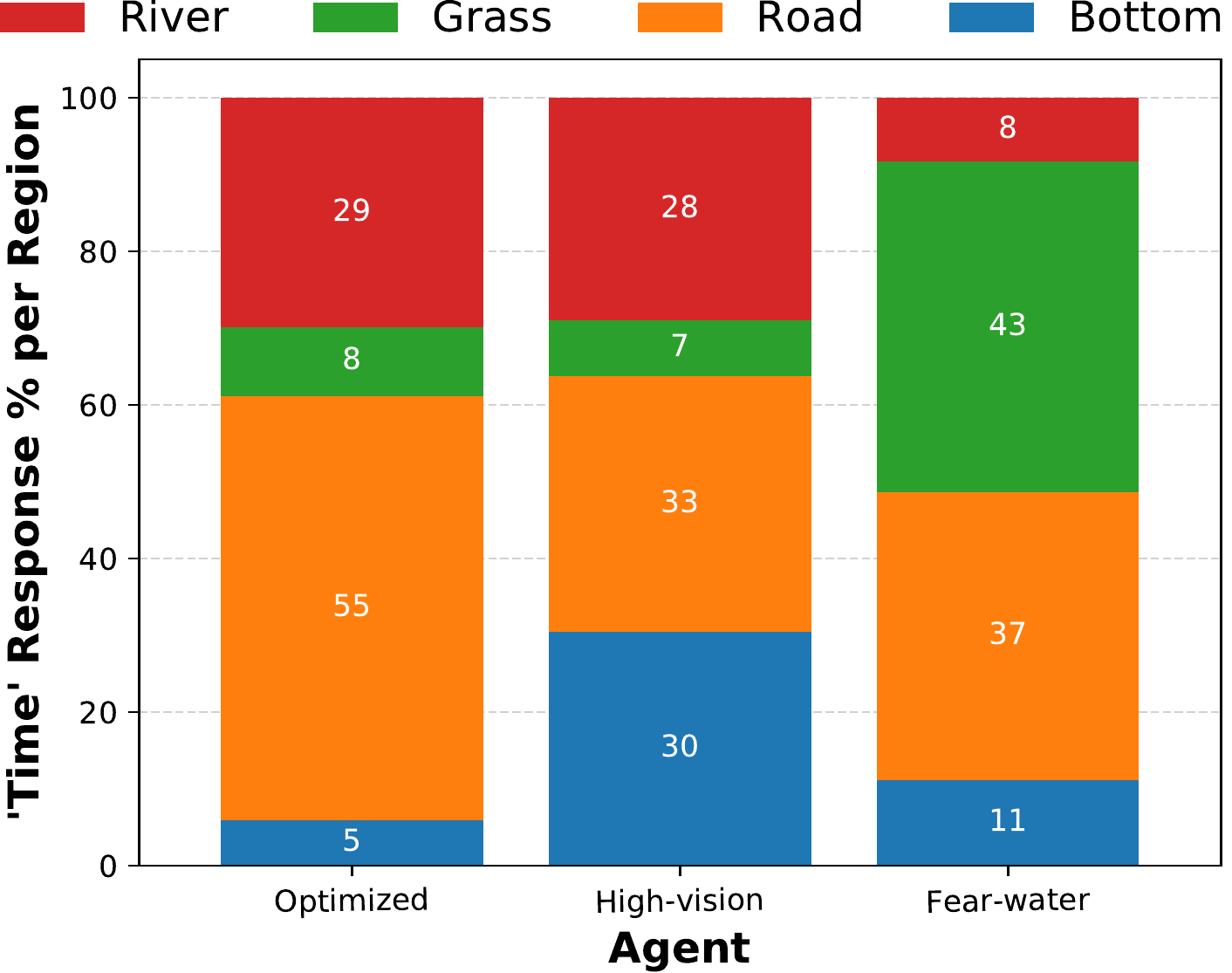}
		\caption{Scenario 10}
        \end{subfigure}%
        \hspace{10pt}
	\begin{subfigure}[b]{0.33\columnwidth}
		\includegraphics[width=\textwidth]{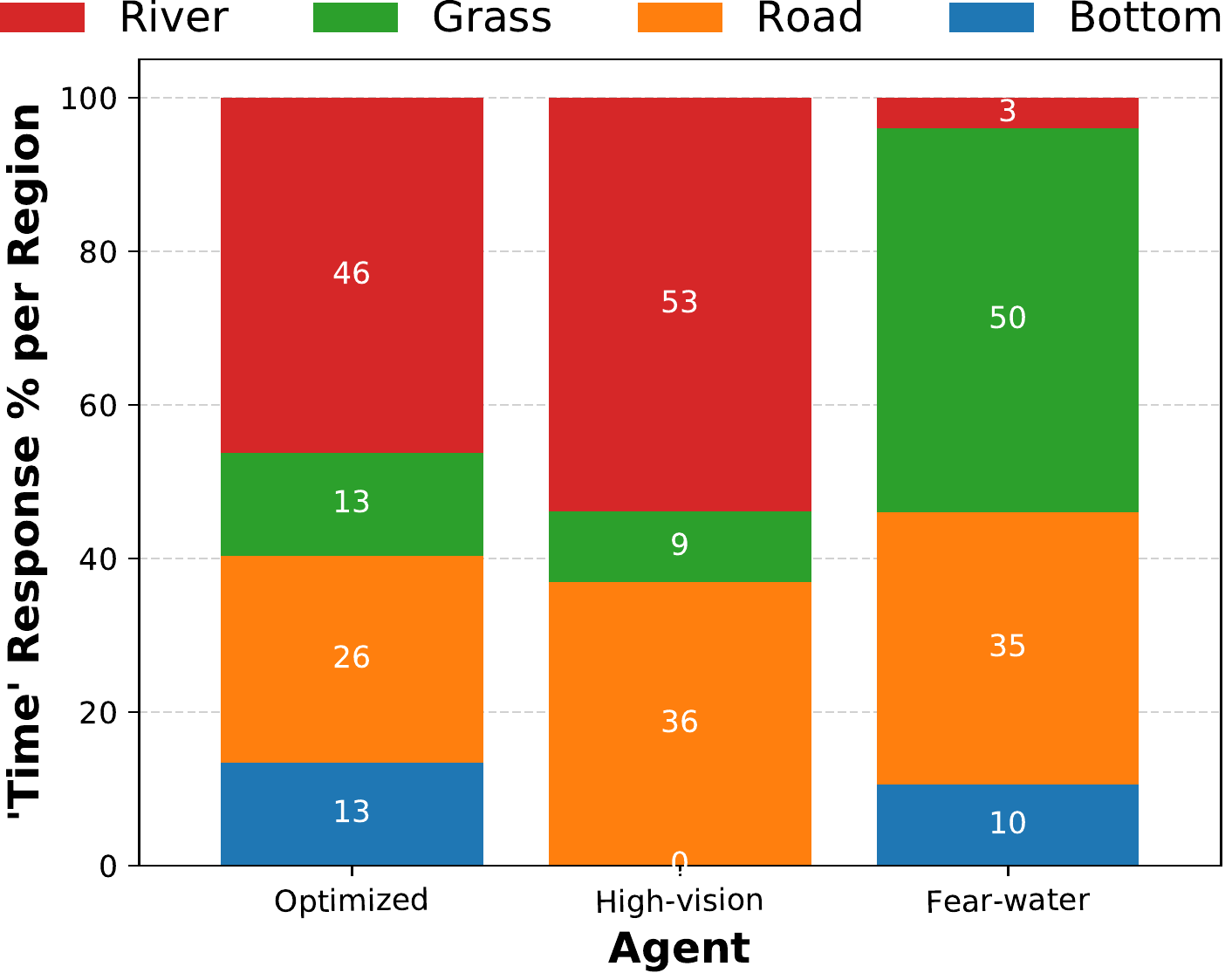}
		\caption{Scenario 11}
        \end{subfigure}
    	\caption{Subjects' responses to the \emph{time} question, for each scenario. Plots correspond to the percentage of positive responses, \ie relative frequency that subjects selected some region for an agent in each scenario.}%
    	\label{Fig:TimeResponses}
\end{figure}

\begin{figure}[!tb]
	\centering
        \begin{subfigure}[b]{0.33\columnwidth}
		\includegraphics[width=\textwidth]{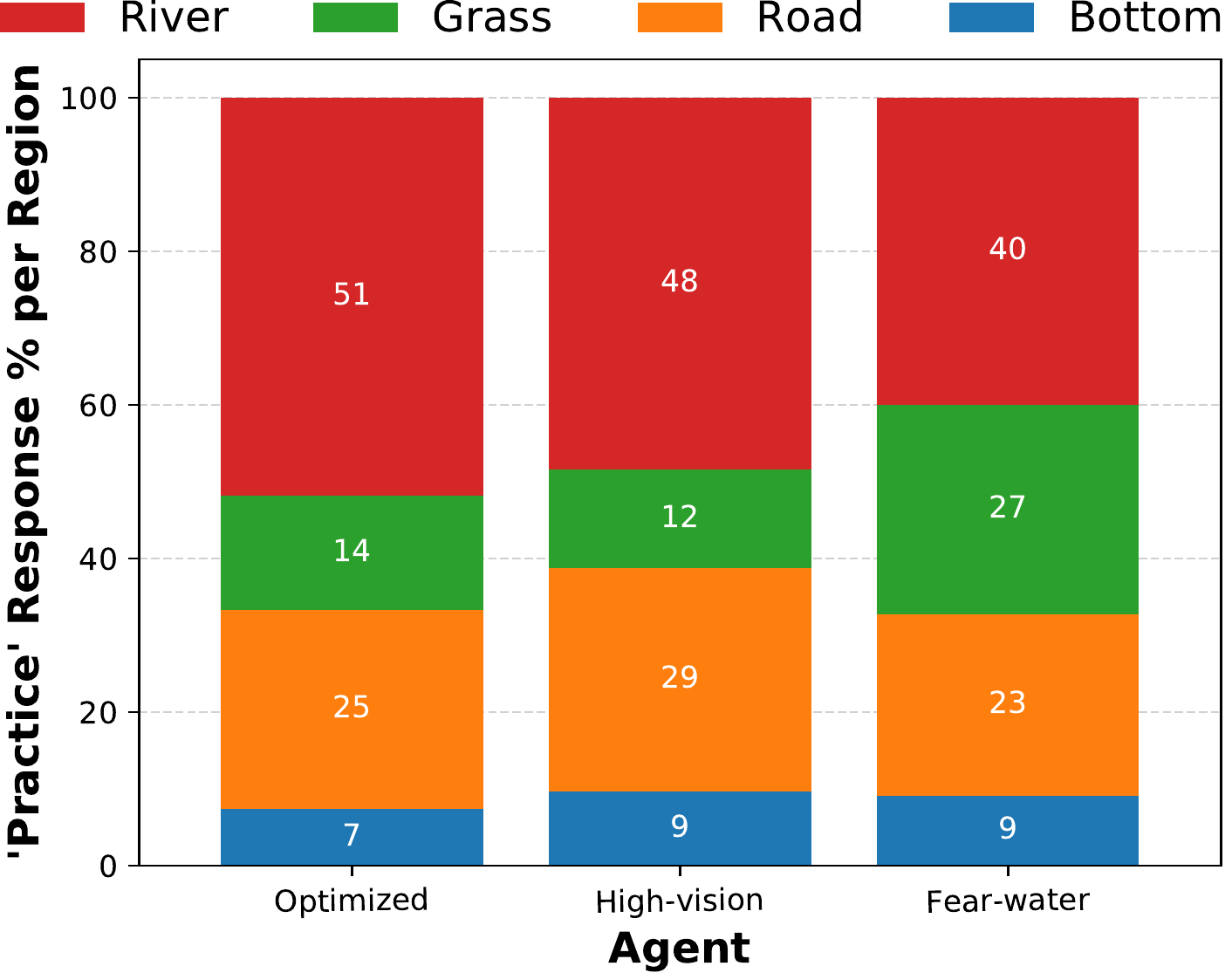}
		\caption{Scenario 1}
        \end{subfigure}%
	\begin{subfigure}[b]{0.33\columnwidth}
		\includegraphics[width=\textwidth]{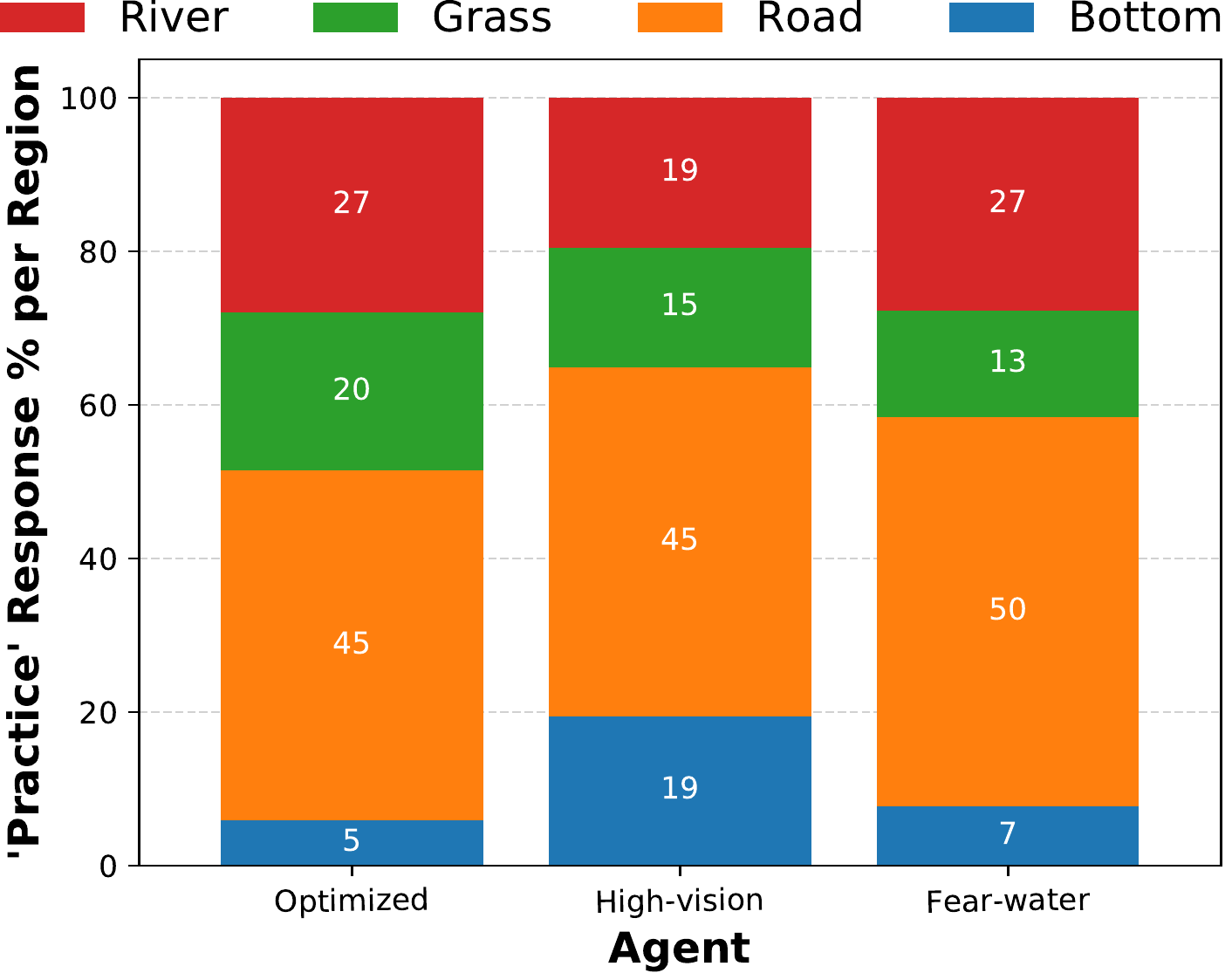}
		\caption{Scenario 2}
        \end{subfigure}%
        \begin{subfigure}[b]{0.33\columnwidth}
		\includegraphics[width=\textwidth]{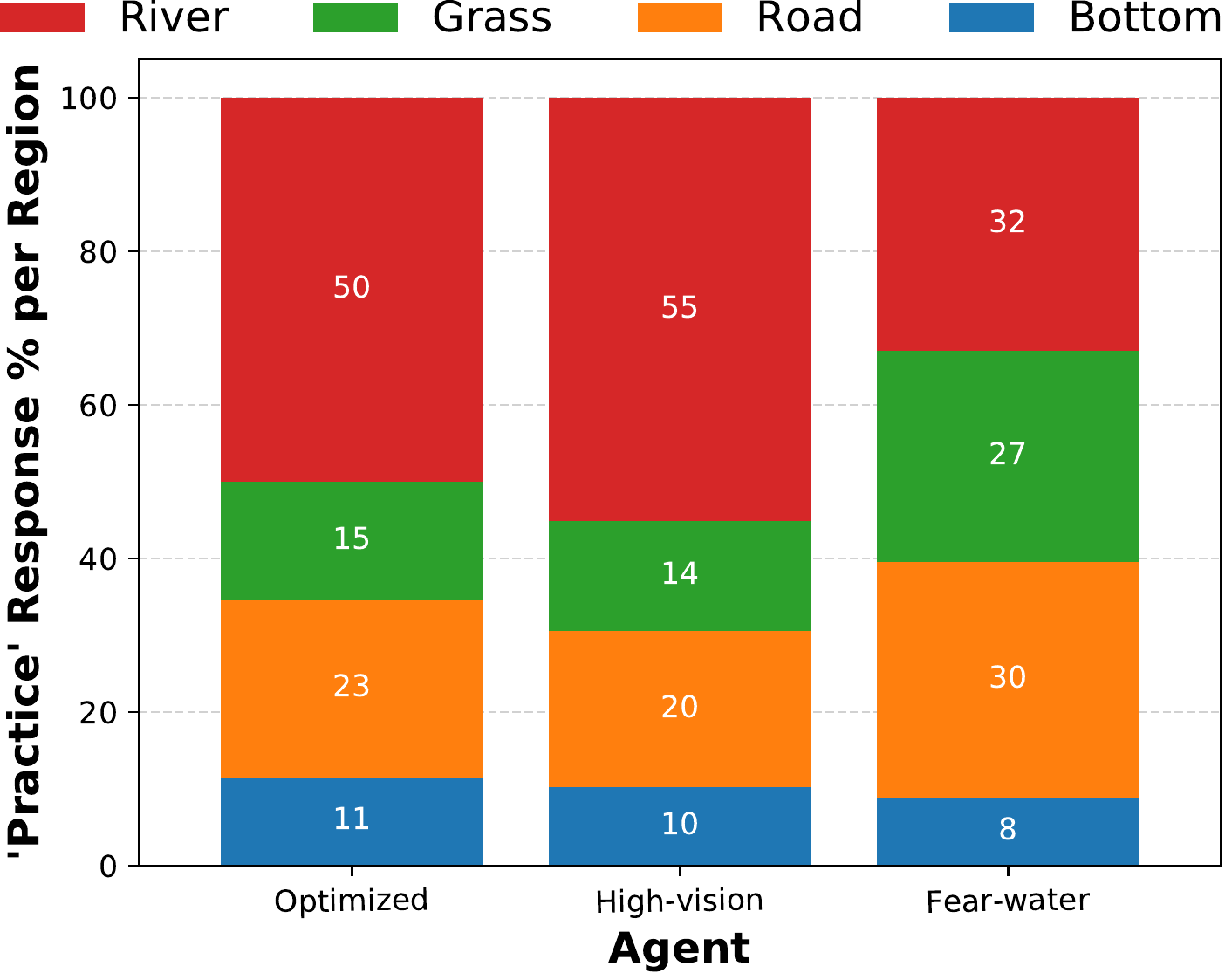}
		\caption{Scenario 3}
        \end{subfigure}
        \begin{subfigure}[b]{0.33\columnwidth}
		\includegraphics[width=\textwidth]{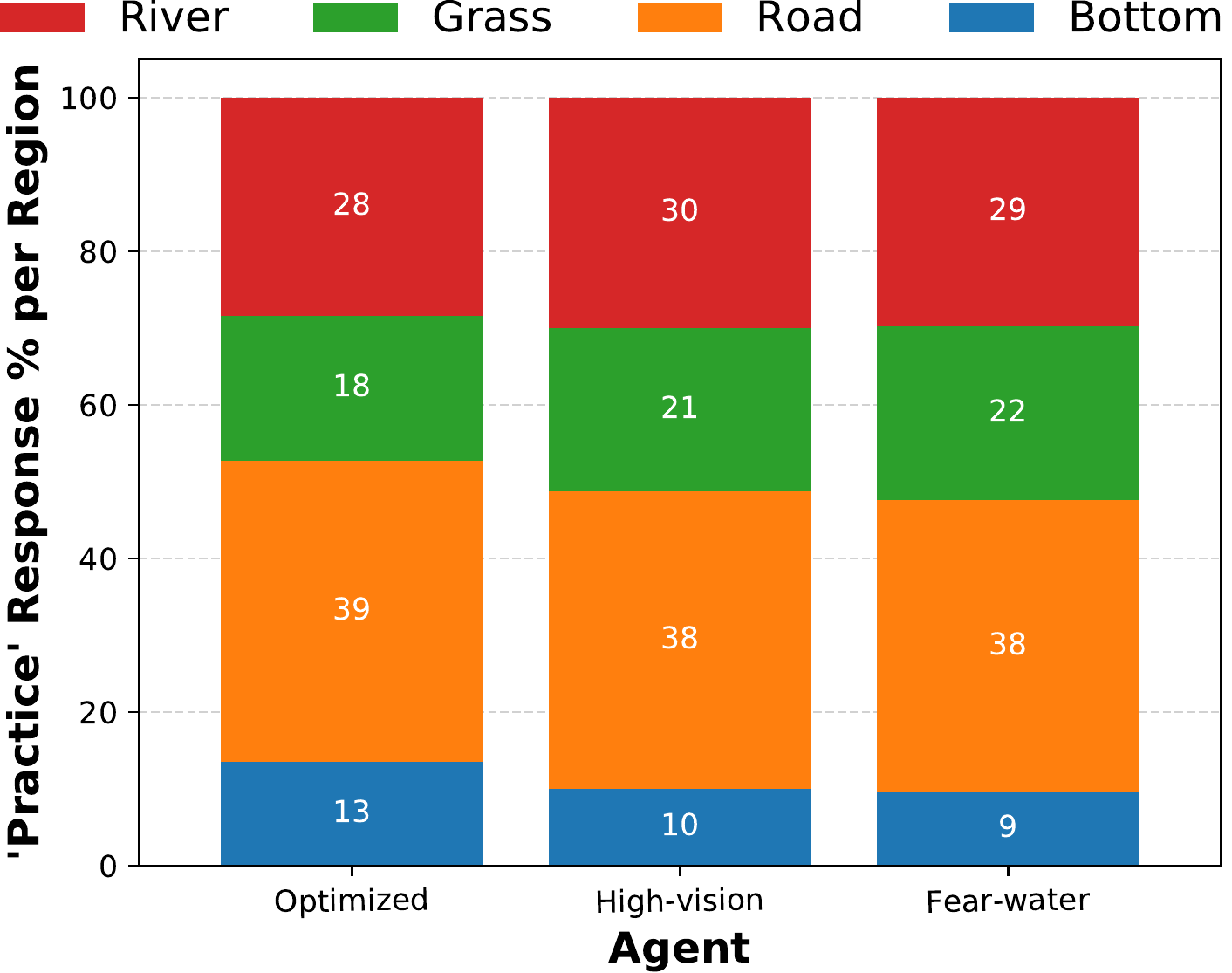}
		\caption{Scenario 4}
        \end{subfigure}%
	\begin{subfigure}[b]{0.33\columnwidth}
		\includegraphics[width=\textwidth]{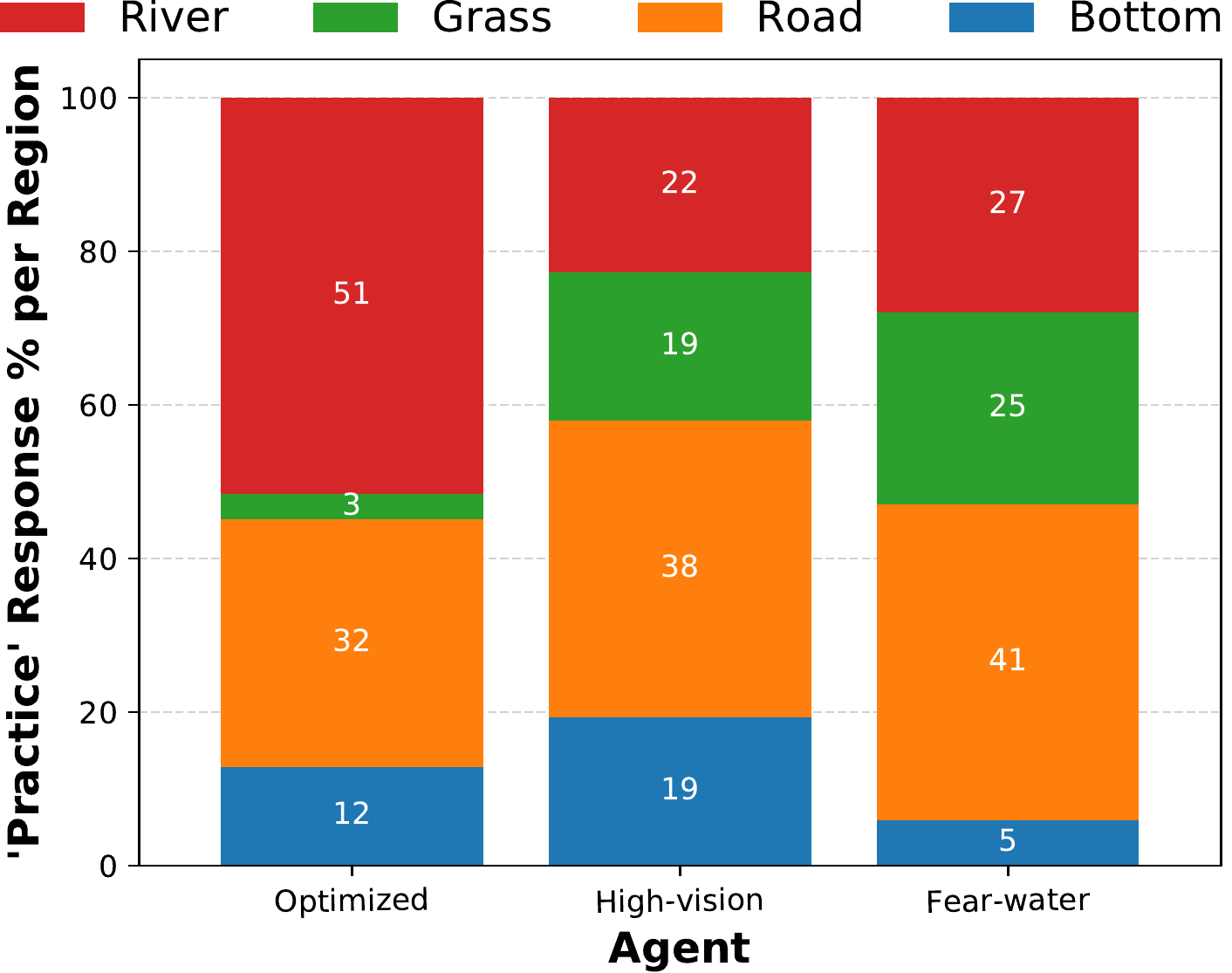}
		\caption{Scenario 5}
        \end{subfigure}%
        \begin{subfigure}[b]{0.33\columnwidth}
		\includegraphics[width=\textwidth]{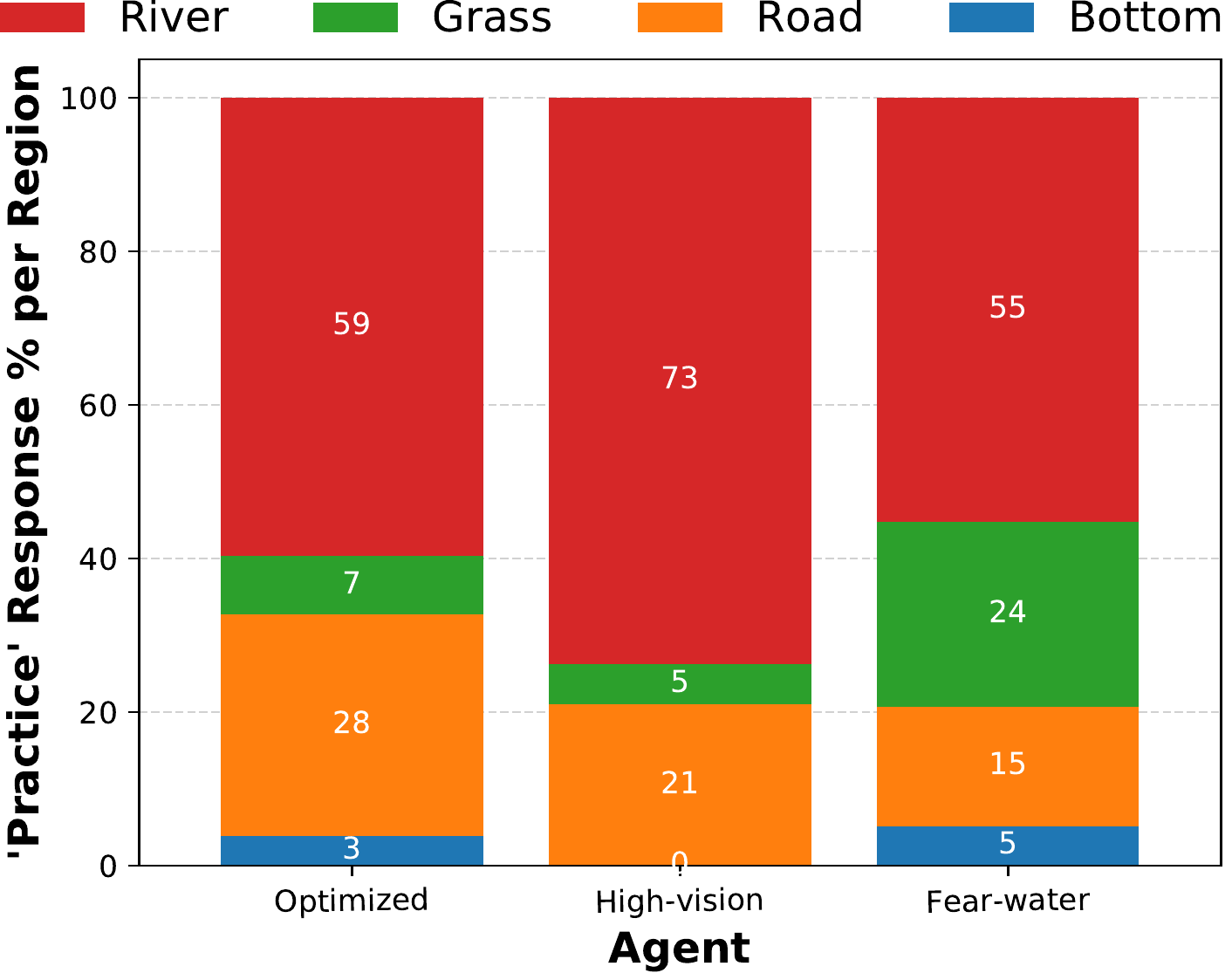}
		\caption{Scenario 6}
        \end{subfigure}
        \begin{subfigure}[b]{0.33\columnwidth}
		\includegraphics[width=\textwidth]{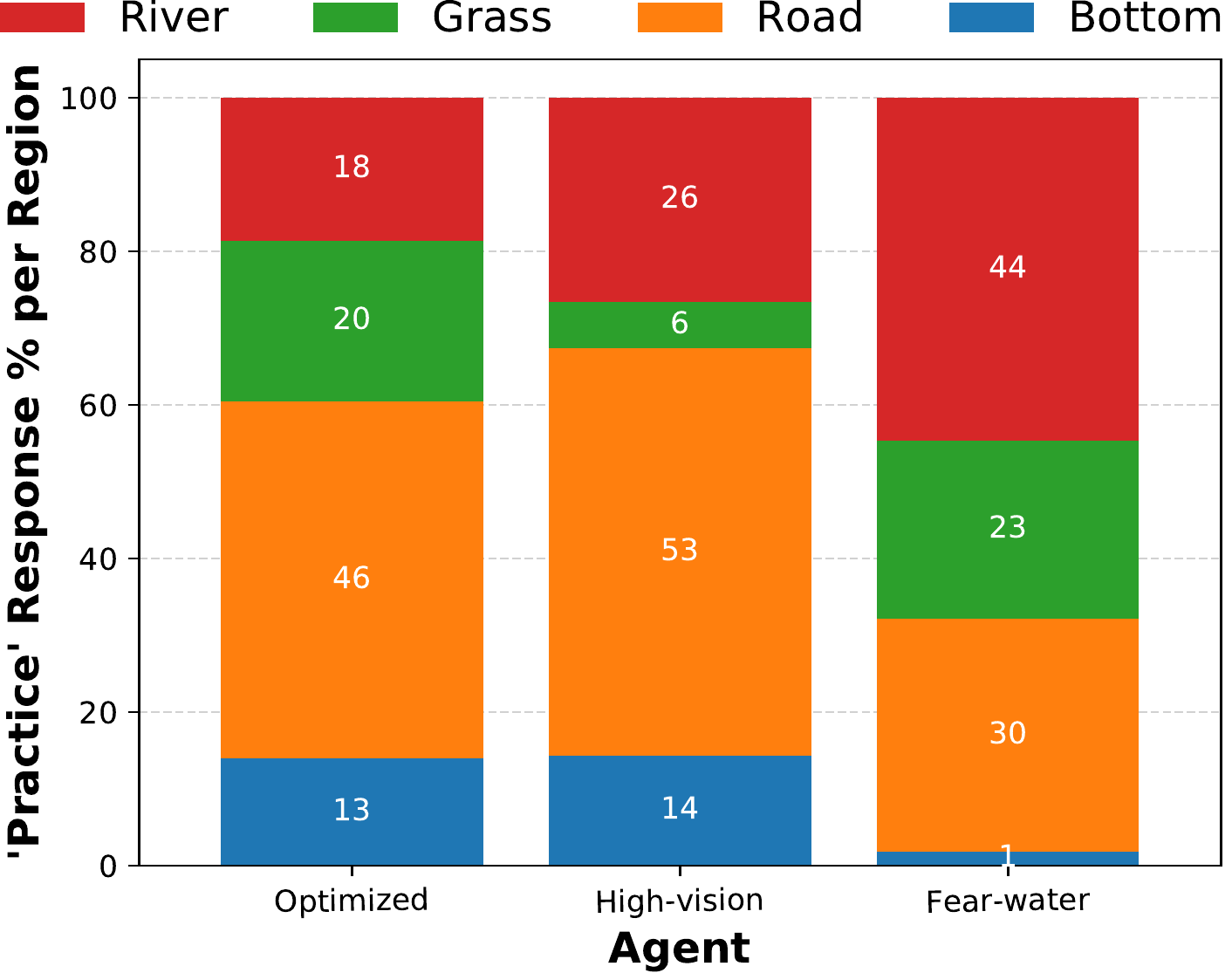}
		\caption{Scenario 7}
        \end{subfigure}%
	\begin{subfigure}[b]{0.33\columnwidth}
		\includegraphics[width=\textwidth]{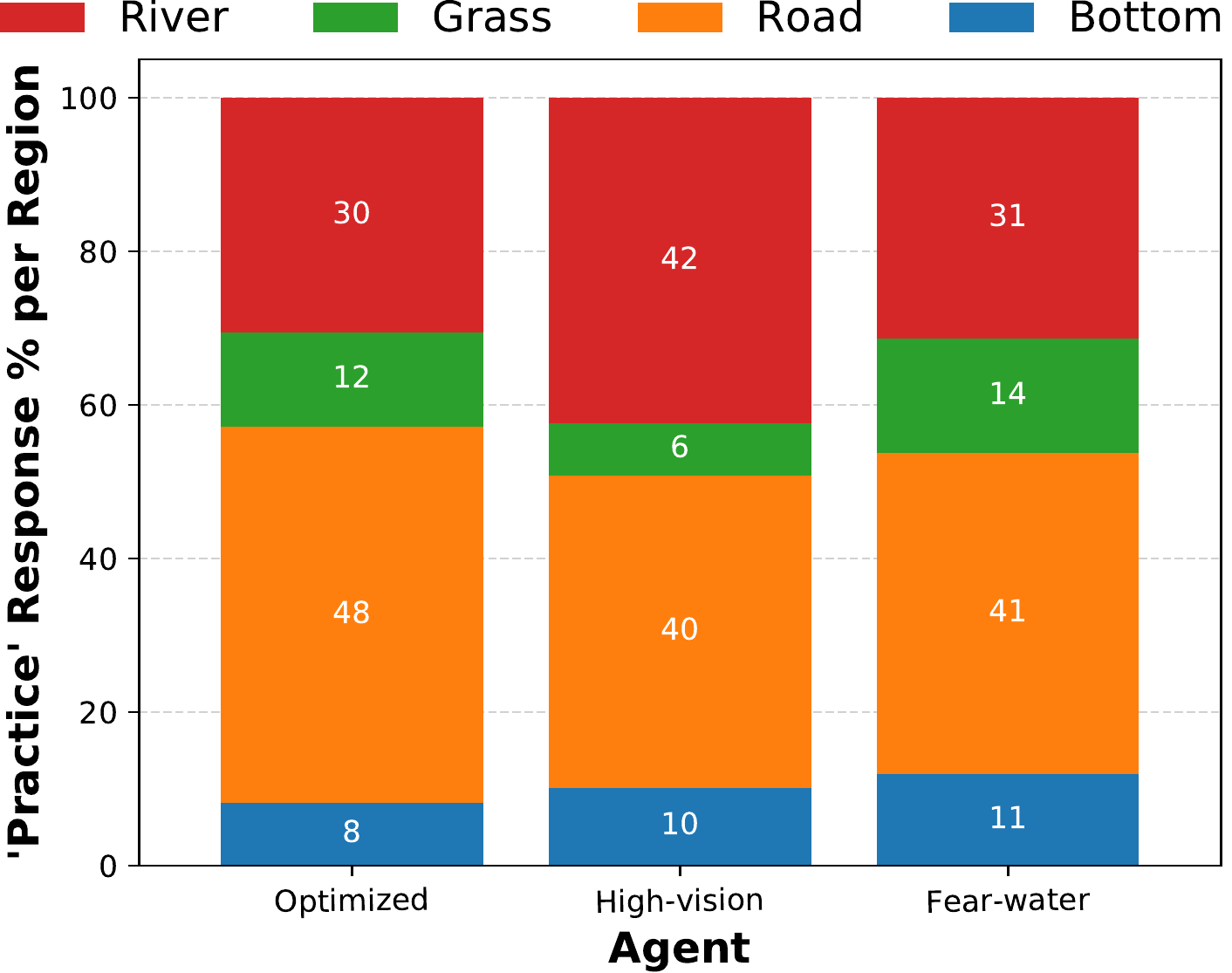}
		\caption{Scenario 8}
        \end{subfigure}%
        \begin{subfigure}[b]{0.33\columnwidth}
		\includegraphics[width=\textwidth]{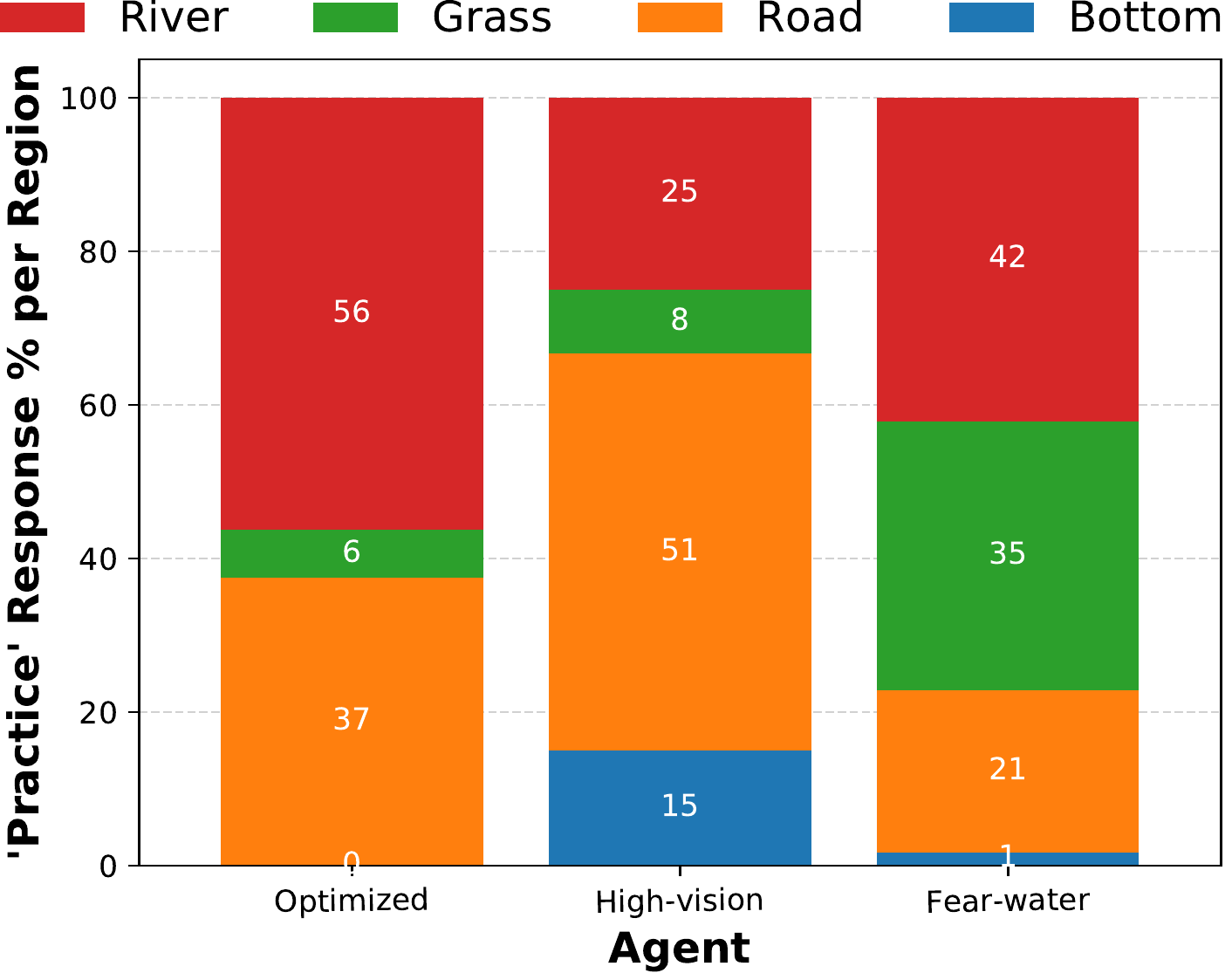}
		\caption{Scenario 9}
        \end{subfigure}
        \begin{subfigure}[b]{0.33\columnwidth}
		\includegraphics[width=\textwidth]{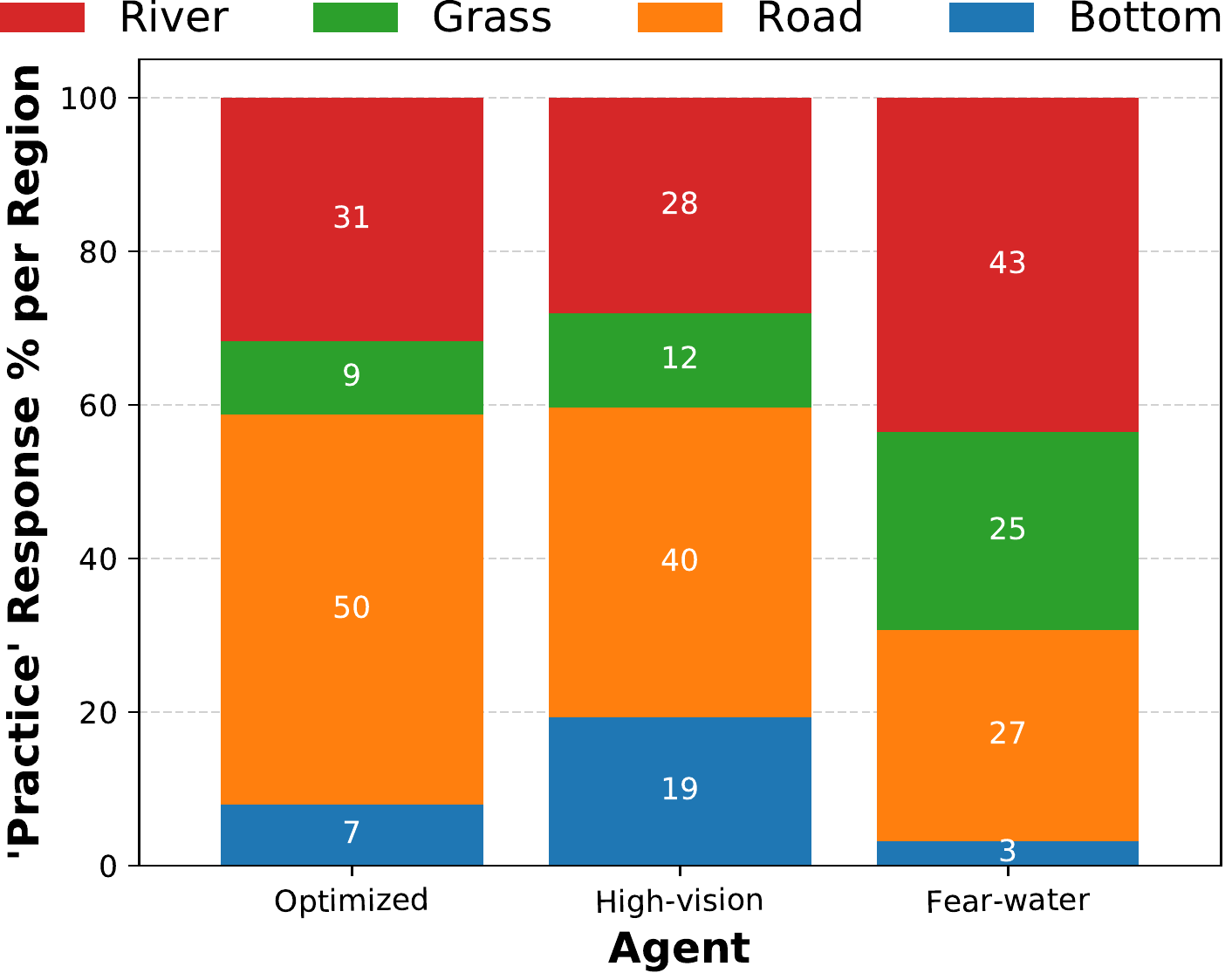}
		\caption{Scenario 10}
        \end{subfigure}%
        \hspace{10pt}
	\begin{subfigure}[b]{0.33\columnwidth}
		\includegraphics[width=\textwidth]{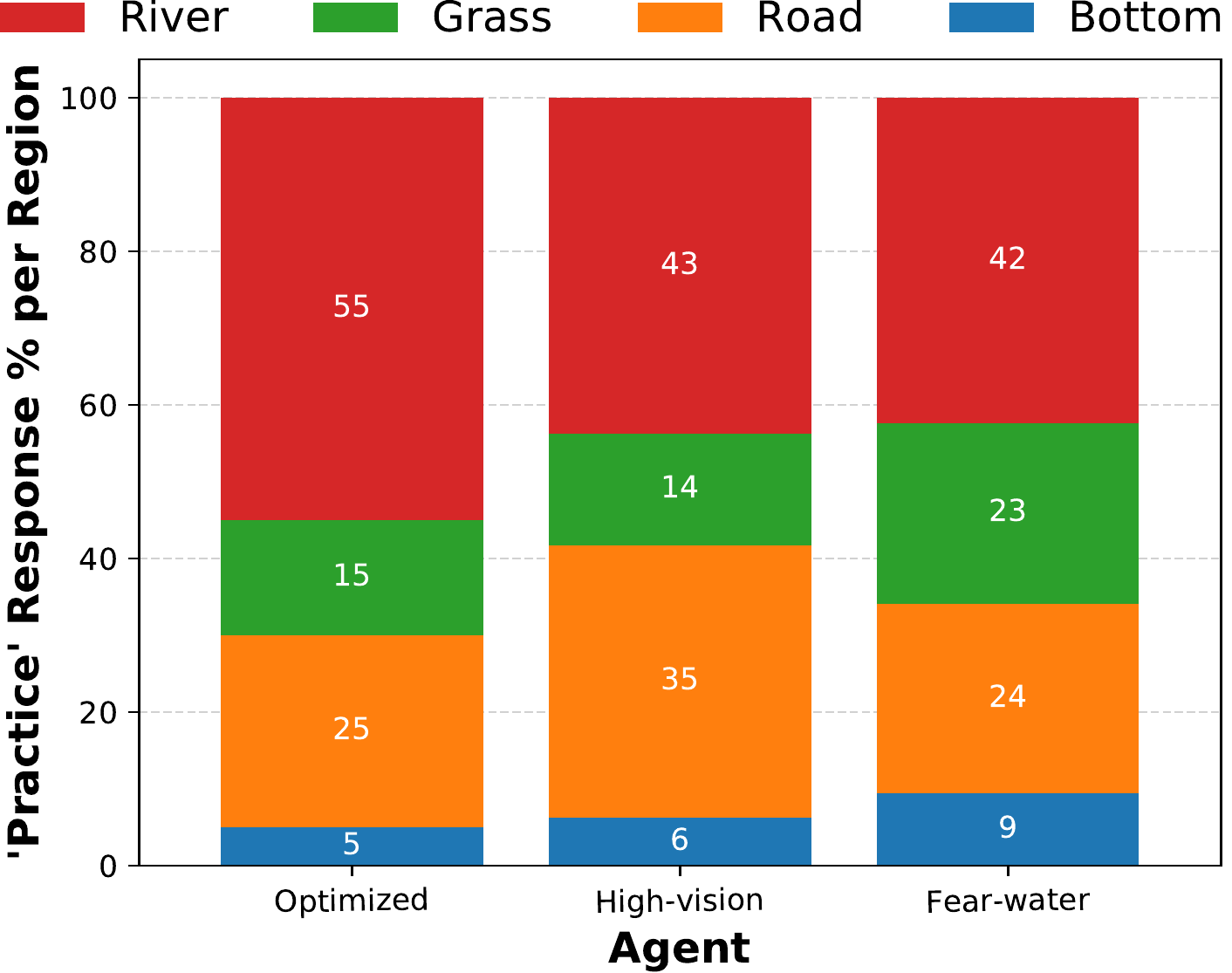}
		\caption{Scenario 11}
        \end{subfigure}
    	\caption{Subjects' responses to the \emph{practice} question, for each scenario. Plots correspond to the percentage of positive responses, \ie relative frequency that subjects selected some region for an agent in each scenario.}%
    	\label{Fig:PracticeResponses}
\end{figure}

This appendix provides in Figs.~\ref{Fig:TimeResponses} and \ref{Fig:PracticeResponses} the percentage of subject responses relative to each region of the Frogger environment and each agent for the region-related survey questions, corresponding to variables \emph{time} and \emph{practice}. 

\end{document}